\documentclass[11pt]{article}
\usepackage{jair}
\usepackage{theapa}

\jairheading{27}{2006}{617-674}{07/06}{12/06}
\ShortHeadings{Uncertainty in soft temporal constraint problems 
}
{Rossi, Venable,\& Yorke-Smith}
\firstpageno{617}

\usepackage{graphics}
\input{epsf}

\usepackage{times}
\usepackage{helvet}
\usepackage{courier}

\usepackage{theorem}
\usepackage{amsmath}
\usepackage{amssymb}
\usepackage{amsfonts}
\usepackage{graphicx}
\usepackage{wasysym}
\usepackage{multirow}
\usepackage{array}
\usepackage{subfigure}
\usepackage{xspace}
\usepackage{url}
\usepackage[all]{xypic}

\def\defemb#1#2{\expandafter\def\csname #1\endcsname 
  {\relax\ifmmode #2\else\hbox{$#2$}\fi}} 

\defemb{cA}{{\cal A}} 
\defemb{cB}{{\cal B}} 
\defemb{cC}{{\cal C}} 
\defemb{cD}{{\cal D}} 
\defemb{cE}{{\cal E}} 
\defemb{cG}{{\cal G}} 
\defemb{cI}{{\cal I}} 
\defemb{cK}{{\cal K}} 
\defemb{cN}{{\cal N}} 
\defemb{cM}{{\cal M}} 
\defemb{cP}{{\cal P}} 
\defemb{cR}{{\cal R}} 
\defemb{cS}{{\cal S}} 
\defemb{cT}{{\cal T}} 
\defemb{cV}{{\cal V}} 
\defemb{cO}{{\cal O}} 

\defemb{bF}{{\bf F}} 
\defemb{bp}{{\bf p}} 
\defemb{bU}{{\bf U}} 
\defemb{bu}{{\bf u}} 
\defemb{bV}{{\bf V}} 
\defemb{bW}{{\bf W}} 
\defemb{bw}{{\bf w}} 
\defemb{bX}{{\bf X}} 
\defemb{bx}{{\bf x}} 
\defemb{bY}{{\bf Y}} 
\defemb{by}{{\bf y}} 
\defemb{bZ}{{\bf Z}} 
\defemb{bz}{{\bf z}}

\def\0{{\bf 0}}
\def\1{{\bf 1}}

\newcommand{\semiring}{\langle A, +, \times, \0, \1 \rangle}

\newtheorem{theorem}{Theorem} 
 
\newtheorem{corollary}{Corollary}

{\theorembodyfont{\normalfont} 
  \newtheorem{definition}{Definition} 
  \newtheorem{example}{Example}} 

\newcommand{\proof}[1]{\noindent{\bf Proof:} #1}

\newtheorem{theorem2}{Theorem}
\newtheorem{lemma2}{Lemma}
\newtheorem{corollary2}{Corollary}

\newcommand{\defn}[1]{\emph{#1}}

\newcommand{\eg}{e.g.\@\xspace}

\newcommand{\ie}{i.e.\@\xspace}

\newcommand{\bigoh}[1]{\ensuremath{\mathcal{O}(#1)}}

\newcommand{\wrt}{with respect to \@\xspace}

\newcommand{\comment}[1]{}

\newcommand{\solver}[1]{\mbox{\textsf{#1}}\xspace}
\newcommand{\solverns}[1]{\mbox{\textsf{#1}}} 

\newcommand{\pc}{\solver{PC}}
\newcommand{\pca}{\solverns{PC}}
\newcommand{\weakcon}{\solver{WeaklyControllable}}
\newcommand{\stppchop}{\solver{Chop-solver}}
\newcommand{\scstpu}{\solver{StronglyControllable}}
\newcommand{\scstpua}{\solverns{StronglyControllable}}

\newcommand{\DC}{\solver{DynamicallyControllable}}
\newcommand{\DCA}{\solverns{DynamicallyControllable}}
\newcommand{\bestsc}{\solver{Best-SC}}
\newcommand{\event}[1]{\mathsf{#1}}
\newcommand{\bestdc}{\solver{Best-DC}}
\newcommand{\pref}[1]{\mathrm{pref}(#1)}
\newcommand{\receives}{\leftarrow}

\begin{document}


\title{Uncertainty in Soft Temporal Constraint Problems:
A General Framework and Controllability Algorithms for
The Fuzzy Case}

\author{\name Francesca Rossi \email frossi@math.unipd.it\\
       \name Kristen Brent Venable \email kvenable@math.unipd.it\\
       \addr University of Padova, 
       Department of Pure and Applied Mathematics,\\
       Via Trieste, 63  35121 PADOVA ITALY
       \AND
       \name Neil Yorke-Smith \email nysmith@ai.sri.com\\
       \addr SRI International,\\
       333 Ravenswood Ave,
       Menlo Park, CA 94025 USA}


\maketitle

\begin{abstract}
In real-life temporal scenarios, uncertainty and preferences are often
  essential and coexisting aspects.  
We present a formalism where quantitative
  temporal constraints with both preferences and uncertainty can be
  defined.  We show how three classical notions of controllability (that
  is, strong, weak, and dynamic), which have been developed for uncertain
  temporal problems, can be generalized to handle preferences as well.
  After defining this general framework, we focus on
  problems where preferences follow the fuzzy approach, and with properties 
that assure tractability.
  For such problems, we
  propose algorithms 
to check the presence of the controllability
  properties. In
  particular, we show that in such a setting
  dealing simultaneously with preferences and
  uncertainty does not increase the complexity of controllability testing.
We also develop a dynamic execution algorithm, of
  polynomial complexity, that produces temporal plans under uncertainty
  that are optimal with respect to fuzzy preferences.
\end{abstract}


\section{Introduction}
\label{intro} 

\label{sec:motiv}
Current research on temporal constraint reasoning, once exposed to the
difficulties of real-life problems, can be found lacking both
expressiveness and flexibility.  In rich application domains it is often
necessary to simultaneously handle not only temporal constraints, but also
preferences and uncertainty.

This need can be seen in many scheduling domains.  The motivation for
the line of research described in this paper is the domain of planning and
scheduling for NASA space missions.  NASA has tackled many scheduling
problems in which temporal constraints have been used with reasonable
success, while showing their limitations in their lack of capability to
deal with uncertainty and preferences.  For example, the Remote Agent
\cite{raj:remagent,mus:issues} experiments, which consisted of placing an
AI system on-board to plan and execute spacecraft activities, represents
one of the most interesting examples of this.  Remote Agent worked with
high level goals which specified, for example, the duration and frequency
of time windows within which the spacecraft had to take asteroid images to
be used for orbit determination for the on-board navigator. Remote Agent
dealt with both flexible time intervals and uncontrollable events; however,
it did not deal with preferences: all the temporal constraints are hard.
The benefit of adding preferences to this framework would be to allow the
planner to handle uncontrollable events while at the same time maximizing
the mission manager's preferences. 

A more recent NASA application is in the rovers domain
\cite{dea:rovers,bre:rovers:icaps05}.  NASA is interested in the generation
of optimal plans for rovers designed to explore a planetary surface (\eg
Spirit and Opportunity for Mars) \cite{bre:rovers:icaps05}.
Dearden et al. (2002) describe the problem of generating plans for planetary
rovers that handle uncertainty over time
and resources.  The approach involves first constructing a ``seed'' plan,
and then incrementally adding contingent branches to this plan in order to
improve its utility.  Again, preferences could be used to embed utilities
directly in the temporal model.

A third space application, which will be used several times in this paper
as a running example, concerns planning for fleets of Earth Observing
Satellites (EOS) \cite{fra:fleets}. This planning problem involves multiple
satellites, hundreds of requests, constraints on when and how to serve each
request, and resources such as instruments, recording devices, transmitters
and ground stations. In response to requests placed by scientists, image
data is acquired by an EOS.  The data can be either downlinked in real time
or recorded on board for playback at a later time. Ground stations or other
satellites are available to receive downlinked images. Different satellites
may be able to communicate only with a subset of these resources, and
transmission rates will differ from satellite to satellite and from station
to station. Further, there may be different financial costs associated with
using different communication resources.  In \cite{fra:fleets} the EOS
scheduling problem is dealt with by using a constraint-based interval
representation. Candidate plans are represented by variables and
constraints, which reflect the temporal relationship between actions
and the constraints on the parameters of
states or actions.  Also, temporal constraints are necessary to model
duration and ordering constraints associated with the data collection,
recording, and downlinking tasks.  Solutions are preferred based on
objectives (such as maximizing the number of high priority requests
served, maximizing the expected quality of the observations, and
minimizing the cost of downlink operations).  Uncertainty is present due to
weather: specifically due to duration and persistence of cloud cover, since
image quality is obviously affected by the amount of clouds over the
target. In addition, some of the events that need to be observed may happen
at unpredictable times and have uncertain durations (\eg fires or volcanic
eruptions).

Some existing frameworks, such as \defn{Simple Temporal Problems with
Preferences} (STPPs) \cite{kha:temporal_prefs}, address the lack of
expressiveness of hard temporal constraints by adding preferences to the
temporal framework, but do not take into account uncertainty.  Other
models, such as Simple Temporal Problems with Uncertainty (STPUs)
\cite{vid:stnu}, account for contingent events, but have no notion of
preferences.  In this paper we introduce a framework which allows us to
handle both preferences and uncertainty in Simple Temporal Problems.  The
proposed model, called \defn{Simple Temporal Problems with Preferences and
  Uncertainty} (STPPUs), merges the two pre-existing models of STPPs and
STPUs.

An STPPU instance represents a quantitative temporal problem with
preferences and uncertainty via a set of variables, representing the
starting or ending times of events (which can be controllable by the agent
or not), and a set of soft temporal constraints over such variables, each
of which includes an interval containing the allowed durations of the event
or the allowed 
times between events.  A preference function
associating each element of the interval with a value specifies how much
that value is preferred.  Such soft constraints can be defined on both
controllable and uncontrollable events.
In order to further clarify what is modeled by an STPPU, let us
emphasize that graduality is only allowed in terms of preferences
and not of uncertainty. In this sense,
the uncertainty represented by contingent STPPU constraints is the same as 
that of contingent STPU constraints: 
all durations are assumed to be equally possible.
In addition to expressing uncertainty, in STPPUs, contingent
constraints can be soft and different preference levels can be
associated to different durations of contingent events.

On these problems, we consider notions of controllability similar to those
defined for STPUs, to be used instead of consistency because of the
presence of uncertainty, and we adapt them to handle preferences.  
These notions, usually called \defn{strong}, \defn{weak}, and \defn{dynamic}
controllability, refer to the possibility of ``controlling'' the problem,
that is, of the executing agent assigning values to the controllable
variables, in a way that is \defn{optimal} \wrt what Nature has decided,
or will decide, for the uncontrollable variables.  The word \emph{optimal}
here is crucial, since in STPUs, where there are no preferences, we just
need to care about controllability, and not optimality.  In fact, the notions
we define in this paper that directly correspond to those for STPUs are
called strong, weak, and dynamic \defn{optimal} controllability.  

After defining these controllability notions and 
proving their properties, we then consider the same restrictions 
which 
have been shown to make temporal problems with preferences tractable 
\cite{kha:temporal_prefs,ros:stpp_learning}, i.e, semi-convex
preference functions and totally ordered preferences combined with an
idempotent operator. In this context, for each
of the above controllability notions, we give algorithms that check whether they hold, and we
show that adding preferences does not make the complexity of testing such
properties worse than in the case without preferences.  
Moreover, dealing
with different levels of preferences, we also define testing algorithms 
which refer to the possibility of controlling a
problem while maintaining a preference of at least a certain level
(called $\alpha$-controllability).
Finally, in the context of dynamic controllability, we also consider the
execution of dynamic optimal plans.

Parts of the content of this paper have appeared in
\cite{post1,change,yor:stppu,stppu-cp04}.  
This paper extends the previous work 
in at least two directions. 
First, while in those papers optimal 
and $\alpha$ controllability (strong or dynamic) 
were checked separately, now we can check optimal 
(strong or dynamic) controllability and, 
if it does not hold, the algorithm will return the 
highest $\alpha$ such that the given problem is $\alpha$-strong
or $\alpha$-dynamic controllable. 
Moreover, results are presented in a uniform technical 
environment, by providing a thorough theoretical study of the
properties of the algorithms and their computational aspects,
which makes use of several unpublished proofs. 

This paper is structured as follows.  In Section~\ref{stppu-back} we give
the background on temporal constraints with preference and with
uncertainty.  In Section~\ref{sec:defn} we define our formalism for Simple
Temporal Problems with both preferences and uncertainty and, in
Section~\ref{contdef},
we describe our new notions
of controllability.  Algorithms to test such notions are described
respectively in Section~\ref{osc} for Optimal Strong Controllability, in
Section~\ref{owc} for Optimal Weak Controllability, and in Section
\ref{odc} for Optimal Dynamic Controllability. In Section~\ref{comp-con} we
then give a general strategy for using such notions. Finally, in Section
\ref{stppu-rlw}, we discuss related work, and in Section~\ref{conclusion}
we summarize the main results and we point out some directions for future
developments. To make the paper more readable,
the proofs of all theorems are contained in the Appendix.


\section{Background}
\label{stppu-back}

In this section we give the main notions
of temporal constraints \cite{meiri} 
and the framework of Temporal Constraint
Satisfaction Problems with Preferences (TCSPPs)
\cite{kha:temporal_prefs,ros:stpp_learning}, which extend quantitative
temporal constraints \cite{meiri} with semiring-based preferences
\cite{jacm}. We also describe 
Simple Temporal Problems with Uncertainty
(STPUs) \cite{vid:stnu,mor:stnu}, which extend a tractable subclass of
temporal constraints to model agent-uncontrollable contingent events, and
we define the corresponding notions of controllability, introduced in
\cite{vid:stnu}.

\subsection{Temporal Constraint Satisfaction Problems}
\label{tcsp}
  
One of the requirements of a temporal reasoning system for planning and
scheduling problems is an ability to deal with metric information; in other
words, to handle quantitative information on duration of events (such as
``It will take from ten to twenty minutes to get home'').
Quantitative temporal networks provide a convenient formalism to deal with
such information.  They consider instantaneous events as the variables of
the problem, whose domains are the entire timeline. A variable may represent
either the beginning or an ending point of an event, or a neutral point of
time. An effective representation of quantitative temporal networks, based
on constraints, is within the framework of Temporal Constraint Satisfaction
Problems (TCSPs) \cite{meiri}.

In this paper we are interested in a particular subclass of TCSPs, known as
\defn{Simple Temporal Problems} (STPs) \cite{meiri}.  In such a problem, a
constraint between time-points $X_i$ and $X_j$ is represented in the
constraint graph as an edge $X_i \rightarrow X_j$, labeled by a single
interval $[a_{ij}, b_{ij}]$ that represents the constraint $a_{ij} \leq
X_j-X_i \leq b_{ij}$.  Solving an STP means finding an assignment of values
to variables such that all temporal constraints are satisfied.

Whereas the complexity of a general TCSP comes from having more than one
interval in a constraint, STPs can be solved in polynomial time.  Despite
the restriction to a single interval per constraint, STPs have been
shown to be valuable in many practical applications. 
This is why STPs have
attracted attention in the literature.

An STP can be associated with a directed weighted graph $G_d=(V,E_d)$,
called the \defn{distance graph}. It has the same set of nodes as the
constraint graph but twice the number of edges: for each binary constraint over
variables $X_i$ and $X_j$, the distance graph has an edge $X_i \rightarrow
X_j$ which is labeled by weight $b_{ij}$, representing the linear
inequality $X_j-X_i \leq b_{ij}$, as well as an edge $X_j \rightarrow
X_i$ which is labeled by weight $-a_{ij}$, representing the linear
inequality $X_i-X_j \leq -a_{ij}$.

Each path from $X_i$ to $X_j$ in the distance graph $G_d$, say through
variables $X_{i_0}=X_i, X_{i_1}, X_{i_2}, \dots$ $, X_{i_k}=X_j$ induces the
following \defn{path constraint}: $X_j-X_i \leq \sum_{h=1}^{k}
b_{i_{h-1}i_h}$. The intersection of all induced path constraints yields
the inequality $X_j-X_i \leq d_{ij}$, where $d_{ij}$ is the length of the
shortest path from $X_i$ to $X_j$, if such a length is defined, \ie if
there are no negative cycles in the distance graph.  An STP is consistent
if and only if its distance graph has no negative cycles
\cite{Shostak,Saxe}. This means that enforcing path consistency, by an
algorithm such as \solver{PC-2}, is sufficient for solving STPs
\cite{meiri}.  It follows that a given STP can be effectively specified by
another complete directed graph, called a \defn{$d$-graph}, where each edge
$X_i \rightarrow X_j$ is labeled by the shortest path length $d_{ij}$ in
the distance graph $G_d$.

In \cite{meiri} it is shown that any consistent STP is backtrack-free (that
is, decomposable) relative to the constraints in its \defn{$d$-graph}.
Moreover, the set of temporal constraints of the form $[-d_{ji},d_{ij}]$ is
the \defn{minimal STP} corresponding to the original STP and it is possible
to find one of its solutions using a backtrack-free search that simply
assigns to each variable any value that satisfies the minimal network
constraints compatibly with previous assignments.
Two specific solutions (usually called the \defn{latest} and the
\defn{earliest} assignments) are given by
$S_L=\{d_{01}, \dots, d_{0n} \}$ and $S_E=\{d_{10}, \dots, d_{n0}\}$, which
assign to each variable respectively its latest and earliest possible time
\cite{meiri}.

The \defn{$d$-graph} (and thus the \defn{minimal network}) of an STP can be found
by applying Floyd-Warshall's \defn{All Pairs Shortest Path} algorithm
\cite{floyd} to the distance graph with a complexity of $O(n^3)$ where $n$
is the number of variables.  If the graph is sparse, the Bellman-Ford
\defn{Single Source Shortest Path} algorithm can be used instead, with a
complexity equal to $O(nE)$, where $E$ is the number of edges.  We refer to
\cite{meiri,xu:triangle-stp} for more details on efficient STP solving.

\subsection{Temporal CSPs with Preferences}
\label{TCSPPs}

Although expressive, TCSPs model only hard temporal constraints. This means
that all constraints have to be satisfied, and that the solutions of a
constraint are all equally satisfying.  However, in many real-life
situations some solutions are preferred over others and, thus, the global
problem is to find a way to satisfy the constraints optimally, according to
the preferences specified.

To address this need, the TCSP framework has been generalized in
\cite{kha:temporal_prefs} to associate each temporal constraint with a
preference function which specifies the preference for each distance
allowed by the constraint.  This framework merges TCSPs and semiring-based
soft constraints \cite{jacm}.

\begin{definition}[soft temporal constraint]
\label{stc}
A \defn{soft temporal constraint} is a 4-tuple $\langle \{X,Y\},$ $I,$ $A,$ $f \rangle$ 
consisting of 
\begin{itemize}
\item a set of two variables $\{X,Y\}$ over the integers, called the scope
  of the constraint;
\item a set of disjoint intervals $I=\{ [a_1,b_1], \ldots, [a_n,b_n]\}$,
  where $a_i,b_i \in \mathbb{Z}$, and $a_i \leq b_i$ for all $i=1,\ldots,n$;
\item a set of preferences $A$;
\item a preference function $f : I \rightarrow A$, which is a
  mapping of the elements of $I$ into preference values, taken from the set
  $A$.
\end{itemize}

Given an assignment of the variables $X$ and $Y$, $X = v_x$ and $Y = v_y$,
we say that this assignment \defn{satisfies} 
the constraint $\langle
\{X,Y\}, I, A, f \rangle$ 
 iff there exists
$[a_i,b_i] \in I$ such that $a_i \leq v_y - v_x \leq b_i$.
In such a case, the preference associated with the assignment by the 
constraint is $f(v_y - v_x)=p$.$\Box$
\end{definition}

When the variables and the preference set of an STPP are apparent, we will
omit them and write a soft temporal constraint just as a pair $\langle I, f
\rangle$.

Following the soft constraint approach \cite{jacm}, the preference set is
the carrier of an algebraic structure known as a \defn{c-semiring}.
Informally a c-semiring $S = \langle A,+,\times,\0,\1 \rangle$ is a set
equipped with two operators satisfying some proscribed properties \cite<for
details, see>{jacm}). The additive operator $+$ is used to induce the
ordering on the preference set $A$; given two elements $a,b \in A$, $a \geq
b$ iff $a+b=a$. The multiplicative operator $\times$ is used to combine
preferences.

\begin{definition}[TCSPP]
  Given a semiring $S = \langle A,+,\times,\0,\1 \rangle$, a \defn{Temporal
  Constraint Satisfaction Problems with Preferences} (TCSPP) over $S$ is a
  pair $\langle V, C \rangle$, where $V$ is a set of variables and $C$ is a
  set of soft temporal constraint over pairs of variables in $V$ and with
  preferences in $A$.$\Box$
\end{definition}

\begin{definition}[solution]
  Given a TCSPP $\langle V, C \rangle$ over a semiring $S$, a
  \defn{solution} is a complete assignment of the variables in $V$.  A
  solution $t$ is said to satisfy a constraint $c$ in $C$ with preference
  $p$ if the projection of $t$ over the pair of variables of $c$'s scope
  satisfies $c$ with preference $p$.  We will write $pref(t,c) = p$.$\Box$
\end{definition}

Each solution has a \defn{global preference value}, obtained by combining,
via the $\times$ operator, the preference levels at which the solution
satisfies the constraints in $C$.

\begin{definition}[preference of a solution]
  Given a TCSPP $\langle V, C \rangle$ over a semiring $S$, the
  \defn{preference of a solution} $t = \langle v_1, \ldots, v_n \rangle$,
  denoted $val(t)$, is computed by $\Pi_{c \in C} pref(s,c)$.$\Box$
\end{definition}

The optimal solutions of a TCSPP are those solutions which have the best
global preference value, where ``best'' is determined by the ordering
$\leq$ of the values in the semiring.

\begin{definition}[optimal solutions]
Given a TCSPP $P =  \langle V, C \rangle$
over the semiring $S$, 
a solution $t$ of $P$ 
is \defn{optimal} if for every other solution $t'$ of $P$,
$t' \not \geq_S t$.$\Box$ 
\end{definition}

Choosing a specific semiring means selecting a class of preferences.  For
example, the semiring
\[S_{FCSP} = \langle [0,1], max, min, 0, 1 \rangle\]  
allows one to model the so-called \defn{fuzzy preferences}
\cite{Rut,schiex}, which associate to each element allowed by a constraint
a preference between $0$ and $1$ (with $0$ being the worst and $1$ being the
best preferences), and gives to each complete assignment the minimal among
all preferences selected in the constraints.  The optimal solutions are
then those solutions with the maximal preference.  Another example is the
semiring $S_{CSP} = \langle \{false, true\},\vee, \wedge, false, true
\rangle$, which allows one to model classical TCSPs, without preferences, in
the more general TCSPP framework.

In this paper we will refer to fuzzy temporal constraints.  However, the
absence of preferences in some temporal constraints can always be modelled
using just the two elements $0$ and $1$ in such constraints.  Thus preferences
can always coexists with hard constraints.

A special case occurs when each constraint of a TCSPP contains a single
interval. In analogy to what is done in the case without preferences, such
problems are called \defn{Simple Temporal Problems with Preferences}
(STPPs).  This class of temporal problems is interesting because, as noted
above, STPs are polynomially solvable while general TCSPs are NP-hard, and
the computational effect of adding preferences to STPs is not immediately
obvious.

\begin{figure}[tb]
\begin{center}
\resizebox{3.3in}{!}{\includegraphics{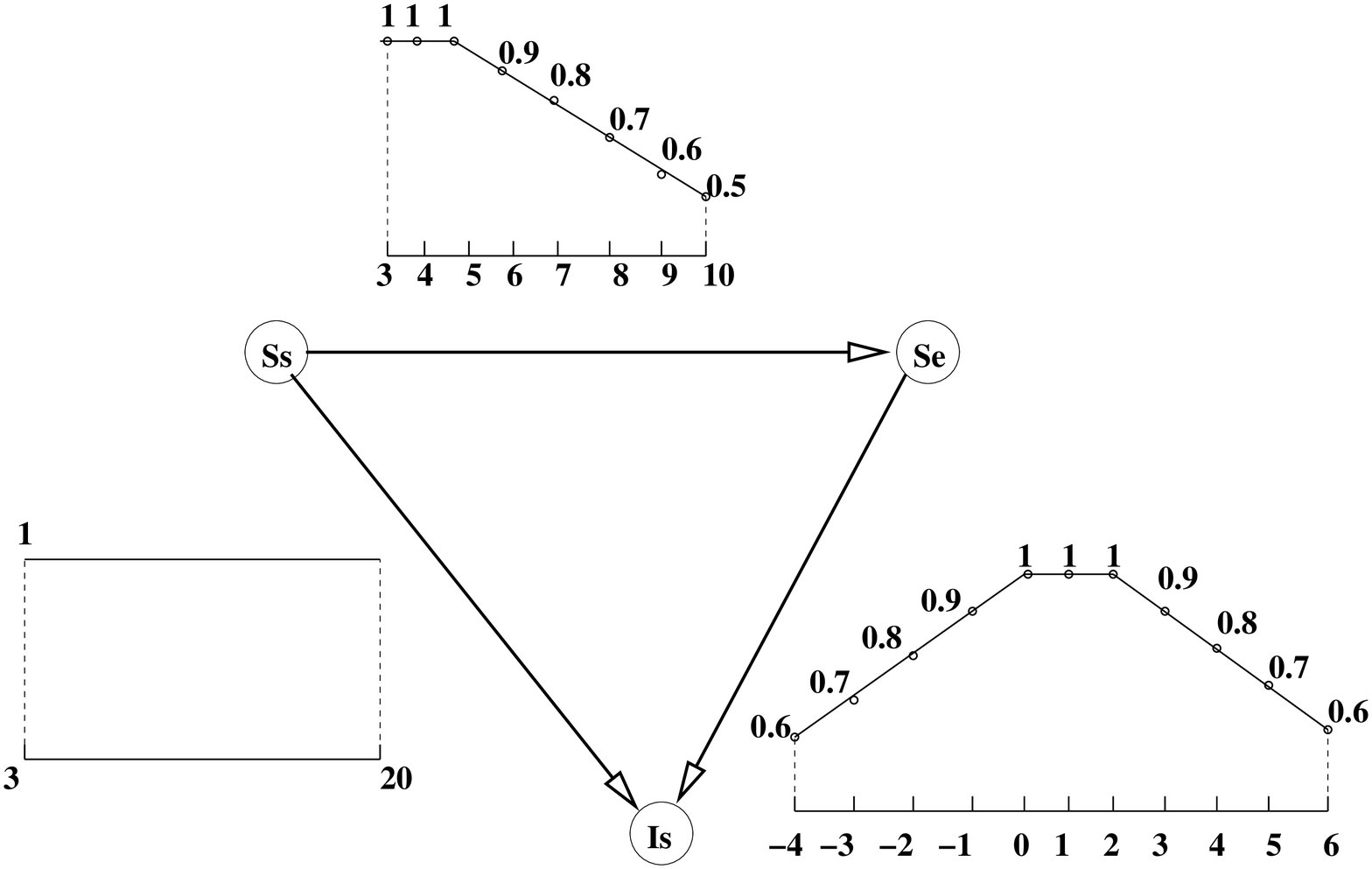}}
\end{center}
\caption{The STPP for \protect{Example~\ref{ex:landsat7}}.}
\label{triag}
\end{figure}

\begin{example}\label{ex:landsat7}
  Consider the EOS example given in Section~\ref{sec:motiv}.  In
  Figure~\ref{triag} we show an STPP that models the scenario in which
  there are three events to be scheduled on a satellite: the start time
  ($Ss$) and ending time ($Se$) of a slewing procedure and the starting
  time ($Is$) of an image retrieval. The slewing activity in this example
  can take from $3$ to $10$ units of time, ideally between $3$ to $5$ units
  of time, and the shortest time possible otherwise.
  The image taking can start any time between $3$ and $20$ units of time
  after the slewing has been initiated.  The third constraint, on variables
  $Is$ and $Se$, models the fact that it is better for the image
  taking to start as soon as the slewing has stopped.$\Box$
\end{example}

In the following example, instead, we consider an STPP which uses the
set-based semiring: $S_{set}=\langle \wp(A),\cup,\cap,\emptyset,A\rangle$.
Notice that, as in the fuzzy semiring, the multiplicative operator, \ie,
intersection, is idempotent, while the order induced by the additive
operator, \ie, union, is partial.

\begin{example}
  Consider a scenario where three friends, Alice, Bob, and Carol, want to
  meet for a drink and then for dinner and must decide at what time to meet
  and where to reserve dinner depending on how long it takes to get to the
  restaurant.  The variables involved in the problem are: the global start
  time $X_0$, with only the value $0$ in its domain, the start time of the
  drink ($Ds$), the time to leave for dinner ($De$), and the time of
  arrival at the restaurant ($Rs$). They can meet, for the drink, between 8
  and 9:00pm and they will leave for dinner after half an hour. Moreover,
  depending on the restaurant they choose, it will take from 20 to 40
  minutes to get to dinner. Alice prefers to meet early and have dinner
  early, like Carol. Bob prefers to meet at 8:30 and to go to the best
  restaurant which is the farthest.  Thus, we have the following two soft
  temporal constraints.  The first constraint is defined on the variable
  pair $(X_0,Ds)$, the interval is [8:00,9:00] and the preference function,
  $f_s$, is such that, $f_s(8:00)=\{Alice,Carol\}$, $f_s(8:30)=\{Bob\}$ and
  $f_s(9:00)=\emptyset$.  The second constraint is a binary constraint on
  pair ($De$,$Rs$), with interval $[20,40]$ and preference function
  $f_{se}$, such that, $f_{se}(20) =\{Alice,Carol\}$ and $f_{se}(20)
  =\emptyset$ and $f_{se}(20) =\{Bob\}$.  There is an additional ``hard''
  constraint on the variable pair $(Ds,De)$, which can be modeled by the
  interval $[30,30]$ and a single preference equal to $\{Alice, Carol,
  Bob\}$.  The optimal solution is $(X_0=0, Ds=8:00, De=8:30, Rs=8:50)$,
  with preference $\{Alice, Carol\}$. $\Box$
\end{example}
 
Although both TCSPPs and STPPs are NP-hard, in \cite{kha:temporal_prefs} a
tractable subclass of STPPs is described. The tractability assumptions are:
the semi-convexity of preference functions, the idempotence of the
combination operator of the semiring, and a totally ordered preference set.
A preference function $f$ of a soft temporal constraint $\langle I,f
\rangle$ is semi-convex iff for all $y \in \Re ^+$, the set $\{x \in I,
f(x) \geq y\}$ forms an interval.  Notice that semi-convex functions
include linear, convex, and also some step functions. The only aggregation
operator on a totally ordered set that is idempotent is $\min$
\cite{minidemp}, \ie the combination operator of the $S_{FCSP}$ semiring.

If such tractability assumptions are met, STPPs can be solved in polynomial
time.  In \cite{ros:stpp_learning} two polynomial solvers for this
tractable subclass of STPPs are proposed.  One solver is based on the
extension of path consistency to TCSPPs.  The second solver decomposes the
problem into solving a set of hard STPs.

\subsection{Simple Temporal Problems with Uncertainty}
\label{stpu-back}

When reasoning concerns activities that an agent performs interacting with
an external world, uncertainty is often unavoidable.  TCSPs assume that all
activities have durations under the control of the agent. Simple
Temporal Problems with Uncertainty (STPUs) \cite{vid:stnu} extend STPs 
by distinguishing \defn{contingent} events, whose occurrence is
controlled by exogenous factors often referred to as ``Nature''.

As in STPs, activity durations in STPUs are modelled by intervals.  The
start times of all activities are assumed to be controlled by the agent
(this brings no loss of generality). The end times, however, fall into two
classes: \defn{requirement} \cite<``free'' in>{vid:stnu} and
\defn{contingent}. The former, as in STPs, are decided by the agent,
but
the agent has no control over the latter: it only can observe their
occurrence after the event; observation is supposed to be known immediately
after the event. The only information known prior to observation of a
time-point is that nature will respect the interval on the duration.
Durations of contingent links are assumed to be independent.

In an STPU, the variables are thus divided into two sets depending on the
type of time-points they represent.

\begin{definition}[variables]
The variables of an STPU are divided into:
\begin{itemize}

\item \defn{executable time-points}: are those points, $b_i$, whose time
  is assigned by the executing agent;
  
\item \defn{contingent time-points}: are those points, $e_i$, whose 
time is assigned by the external world.$\Box$

  


\end{itemize}
\end{definition}

The distinction on variables leads to constraints which are also divided
into two sets, requirement and contingent, depending on the type of
variables they constrain. Note that as in STPs all the constraints are
binary.  Formally:

\begin{definition}
The constraints of an STPU are divided into:
\begin{itemize}
  
\item a \defn{requirement constraint (or link)} $r_{ij}$, on generic
  time-points $t_i$ and $t_j$ \footnote{In general $t_i$ and $t_j$ can be
    either contingent or executable time-points.}, is an interval $I_{ij} =
  [l_{ij},u_{ij}]$ such that $l_{ij} \leq \gamma(t_j)-\gamma(t_i) \leq
  u_{ij}$ where $\gamma(t_i)$ is a value assigned to variable $t_i$
  
\item a \defn{contingent link} $g_{hk}$, on executable point $b_h$
  and contingent point $e_k$, is 
  an interval ${I}_{hk}=[{l}_{ij},{u}_{ij}]$ which contains all the
  possible durations of the contingent event represented by $b_h$ and
  $e_k$.$\Box$
\end{itemize}
\end{definition}

The formal definition of an STPU is the following:

\begin{definition}[STPU]
  A \defn{Simple Temporal Problem with Uncertainty} (STPU) is a 4-tuple
  $N=\{ X_e, X_c, R_r, R_{c} \}$ such that:
\begin{itemize}
\item $X_e= \{ b_1, \dots, b_{n_e} \}$: is the set of executable
  time-points;
\item $X_c= \{ e_1, \dots, e_{n_c} \}$: is the set of contingent time-points;
\item $R_r = \{ c_{i_1j_1}, \dots, c_{i_Cj_C} \}$: is the set $C$ of
  requirement constraints;
\item $R_c = \{ g_{i_1j_1}, \dots, g_{i_Gj_G} \}$: is the set $G$ of
  contingent constraints.$\Box$
\end{itemize}
\end{definition}

\begin{figure}[tb] 
\begin{center}
\resizebox{3.8in}{!}{\includegraphics{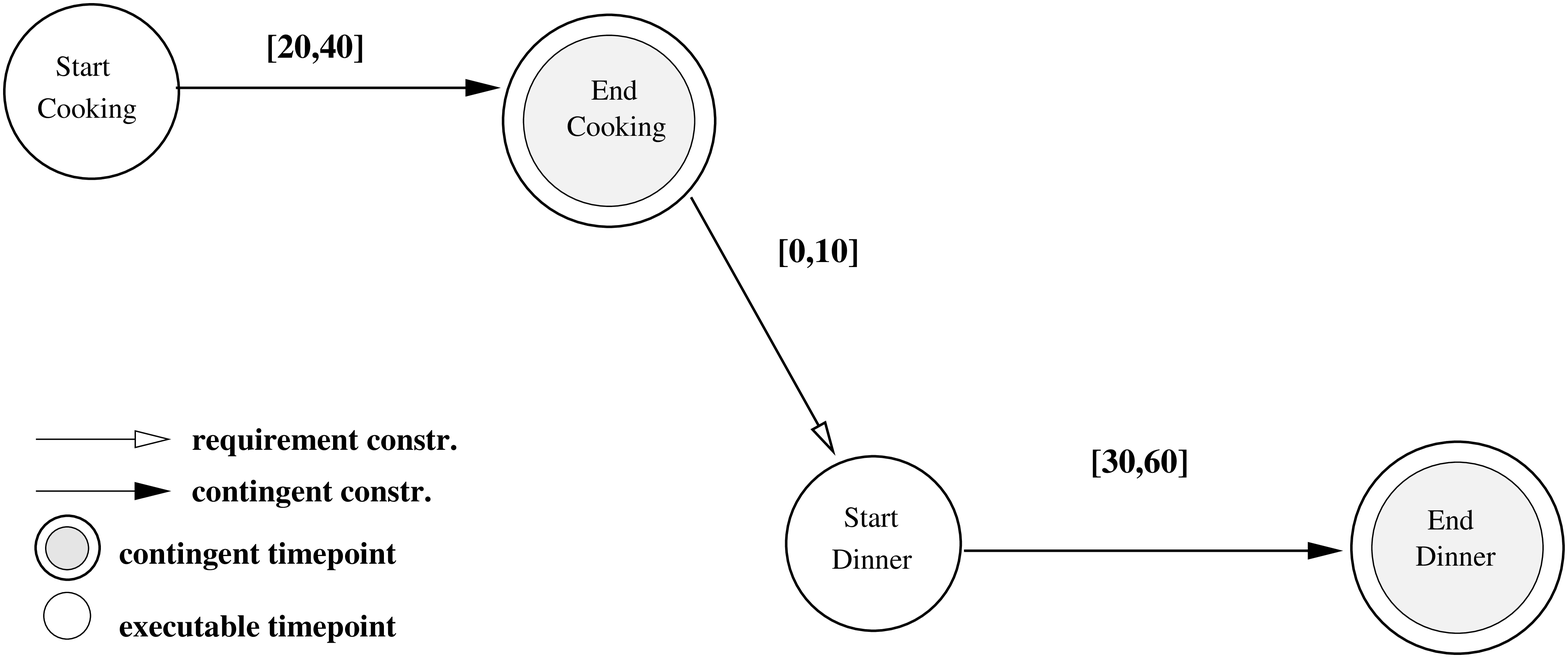}}
\end{center} 
\vspace{-0.8cm} 
\caption{\label{cookdinfig}
The STPU for Example~\ref{cookdinner}.}
\vspace{-0.3cm} 
\end{figure}

\begin{example}\label{cookdinner}
  This is an example taken from \cite{vid:stnu}, which describes a scenario
  which can be modeled using an STPU.  Consider two activities {\em
    Cooking} and {\em Having dinner}. Assume you don't want to eat your
  dinner cold.  Also, assume you can control when you start cooking and
  when the dinner starts but not when you finish cooking or when the dinner
  will be over. The STPU modeling this example is depicted in
  Figure~\ref{cookdinfig}. There are two executable time-points \{{\em
    Start-cooking, Start-dinner}\} and two contingent time-points \{{\em
    End-cooking, End-dinner}\}. Moreover, the contingent constraint on
  variables \{{\em Start-cooking, End-cooking}\} models the uncontrollable
  duration of fixing dinner which can take anywhere from 20 to 40 minutes;
  the contingent constraint on variables \{{\em Start-dinner, End-dinner}\}
  models the uncontrollable duration of the dinner that can last from 30 to
  60 minutes.  Finally, there is a requirement constraint on variables
  \{{\em End-cooking, Start-dinner}\} that simply bounds to 10 minutes the
time between when the food is ready and when the dinner
  starts.$\Box$
\end{example}

Assignments to executable variables and assignments to contingent variables
are distinguished:

\begin{definition}[control sequence]
  A \defn{control sequence} $\delta$ is an assignment to executable
  time-points.  It is said to be \defn{partial} if it assigns values to a
  proper subset of the executables, otherwise \defn{complete}.$\Box$
\end{definition}

\begin{definition}[situation]
  A situation $\omega$ is a set of durations on contingent constraints.  If
  not all the contingent constraints are assigned a duration it is said to
  be \defn{partial}, otherwise \defn{complete}.$\Box$
\end{definition} 

\begin{definition}[schedule]
  A \defn{schedule} is a complete assignment to all the time-points in
  $X_e$ and $X_c$.
  A schedule $T$ identifies a control sequence, $\delta_T$, consisting of
  all the assignments to the executable time-points, and a situation,
  $\omega_T$, which is the set of all the durations identified by the
  assignments in $T$ on the contingent constraints.  \defn{$Sol(P)$}
  denotes the set of all schedules of an STPU.$\Box$
\end{definition}

It is easy to see that to each situation corresponds an STP. In fact, once
the durations of the contingent constraints are fixed, there is no more
uncertainty in the problem, which becomes an STP, called the
\defn{underlying STP}.  This is formalized by the notion of
\defn{projection}.

\begin{definition}[projection]
  A \defn{projection $P_{\omega}$}, corresponding to a situation $\omega$,
  is the STP obtained leaving all requirement constraints unchanged and
  replacing each contingent constraint $g_{hk}$ with the constraint
  $\langle [{\omega}_{hk},{\omega}_{hk}] \rangle$, where ${\omega}_{hk}$ is
  the duration of event represented by $g_{hk}$ in $\omega$.
  \defn{$Proj(P)$} is the set of all projections of an STPU $P$.$\Box$
\end{definition}

\subsection{Controllability}

It is clear that in order to solve a problem with uncertainty all possible
situations must be considered.  The notion of consistency defined for STPs
does not apply since it requires the existence of a single schedule, which
is not sufficient in this case since all situations are equally
possible.\footnote{\citeA{tsamar2} has augmented STPUs to include
  probability distributions over the possible situations; in this paper we
  implicitly assume a uniform, independent distribution on each link.}  For
this reason, in \cite{vid:stnu}, the notion of controllability has been
introduced.  \defn{Controllability} of an STPU is, in some sense, the
analogue of consistency of an STP.  Controllable means the agent has a
means to execute the time-points under its control, subject to all
constraints.  The notion of controllability is expressed, in terms of the
ability of the agent to find, given a situation, an appropriate control
sequence. This ability is identified with having a strategy:

\begin{definition}[strategy]
  A \defn{strategy} $S$ is a map $S$ : $Proj(P) \rightarrow Sol(P)$, such that
  for every projection $P_\omega$, $S(P_\omega)$ is a schedule which
  induces the durations in $\omega$ on the contingent constraints.  
  Further, a strategy is \defn{viable } if, for every
  projection $P_\omega$, $S(P_\omega)$ is a solution of $P_\omega$.$\Box$
\end{definition}

We will write $[S(P_{\omega})]_x$ to indicate the value assigned to
executable time-point $x$ in schedule $S(P_{\omega})$, and
$[S(P_{\omega})]_{<x}$ the \defn{history} of $x$ in $S(P_{\omega})$, that
is, the set of durations of contingent constraints which occurred in
$S(P_{\omega})$ before the execution of $x$, \ie the partial solution so
far.


In \cite{vid:stnu}, three notions of controllability are introduced for 
STPUs.

\subsubsection{Strong Controllability}
The first notion is, as the name suggests, the most restrictive in terms of
the requirements that the control sequence must satisfy.

\begin{definition}[Strong Controllability]
\label{sc}
  An STPU $P$ is \defn{Strongly Controllable} (SC) iff there is
  an execution strategy $S$ s.t.\ $\forall P_\omega \in Proj(P)$,
  $S(P_\omega)$ is a solution of $P_\omega$,
  and $[S(P_1)]_x=[S(P_2)]_x$, $\forall P_1, P_2$ projections and for every
  executable time-point $x$.$\Box$
\end{definition}

In words, an STPU is \defn{strongly controllable} if there is a fixed
execution strategy that works in all situations. 
This means that there is a
fixed control sequence that will be consistent with any possible scenario
of the world. 
Thus, the notion of strong controllability is related to that of
conformant planning.
It is clearly a very strong requirement.
As \citeA{vid:stnu} suggest, SC may be relevant in some
applications where the situation is not observable at all or where the
complete control sequence must be known beforehand (for example in
cases in which other activities depend on the control sequence, 
as in the production planning area).

In \cite{vid:stnu} a polynomial time algorithm for checking if an STPU is
strongly controllable is proposed.  The main idea is to rewrite the STPU
given in input as an equivalent STP only on the executable variables.  What
is important to notice, for the contents of this paper, is that algorithm
\scstpu takes in input an STPU $P = \{ X_e, X_c, R_r, R_{c} \}$ and returns
in output an STP defined on variables $X_e$. The STPU in input is strongly
controllable iff the derived STP is consistent. Moreover, every
solution of the STP is a control sequence which guarantees strong
controllability for the STPU.  When the STP is consistent, the output of
\scstpu is its minimal form.

In \cite{vid:stnu} it is shown that the complexity of \scstpu is $O(n^3)$,
where $n$ is the number of variables. 

\subsubsection{Weak Controllability}
On the other hand, the notion of controllability with the fewest
restrictions on the control sequences is Weak Controllability.

\begin{definition}[Weak Controllability]
  An STPU $P$ is said to be \defn{Weakly Controllable} (WC) iff $\forall
  P_\omega \in Proj(P)$ there is a strategy $S_\omega$ s.t.\ 
  $S_\omega(P_\omega)$ is a solution of $P_\omega$.$\Box$
\end{definition}

In words, an STPU is \defn{weakly controllable} if there is a viable global
execution strategy: there exists at least one schedule for every situation.
This can be seen as a minimum requirement since, if this property
does not hold, then there are some situations such that there is no way
to execute the controllable events in a consistent way.
It also looks attractive since, once an STPU is shown to WC, as soon
as one knows the situation, one can pick out and apply the control
sequence that matches that situation. Unfortunately in \cite{vid:stnu}
it is shown that this property is not so useful in classical planning.
Nonetheless, WC may be relevant in specific applications 
(as large-scale warehouse scheduling) where the actual situation will
be totally observable before (possibly \emph{just before}) the
execution starts, but one wants to know in advance that, whatever the
situation, there will always be at least one feasible control
sequence.

In \cite{vid:stnu} it is conjectured and in \cite{mormusc} it is proven
that the complexity of checking weak controllability is co-NP-hard.
The algorithm proposed for testing WC in \cite{vid:stnu} is based on
a classical enumerative process and a lookahead technique. 

Strong Controllability implies Weak Controllability \cite{vid:stnu}.
Moreover, an STPU can be seen as an STP if the uncertainty is ignored. If
enforcing path consistency 
removes some elements from the contingent intervals, then these elements
belong to no solution.  If so, it is possible to conclude that the STPU is
not weakly controllable.

\begin{definition}[pseudo-controllability]
  An STPU is \defn{pseudo-controllable} if applying path consistency leaves
  the intervals on the contingent constraints unchanged.$\Box$
\end{definition}

Unfortunately, if path consistency leaves the contingent intervals
untouched, we cannot conclude that the STPU is weakly controllable.  That
is, WC implies pseudo-controllability but the converse is false.  In fact,
weak controllability requires that given any possible combination of
durations of all contingent constraints the STP corresponding to that
projection must be consistent.  Pseudo-controllability, instead, only
guarantees that for each possible duration on a contingent constraint there
is at least one projection that contains such a duration and it is a
consistent STP.

\subsubsection{Dynamic Controllability}
In dynamic applications domains, such as planning, the situation is
observed over a time.  Thus decisions have to be made even if the situation
remains partially unknown.  Indeed the distinction between Strong and
Dynamic Controllability is equivalent to that between conformant and
conditional planning.
The final notion of controllability defined in \cite{vid:stnu} address this
case.  Here we give the definition provided in \cite{mor:stnu} which is
equivalent but more compact.

\begin{definition}[Dynamic Controllability]
  An STPU $P$ is \defn{Dynamically Controllable} (DC) iff there is a
  strategy $S$ such that $\forall P_1, P_2$ in $Proj(P)$ and for any
  executable time-point $x$:
  \begin{enumerate}
  \item if $[S(P_1)]_{<x} = [S(P_2)]_{<x}$ then $[S(P_1)]_{x} =
    [S(P_2)]_{x}$;
  \item $S(P_1)$ is a solution of  $P_1$ and $S(P_2)$
    is a solution of  $P_2$.$\Box$
  \end{enumerate}
  \label{DC:defn}
\end{definition}

In words, an STPU is dynamically controllable if there exists a viable
strategy that can be built, step-by-step, depending only the observed
events at each step.  SC $\implies$ DC and that DC $\implies$ WC.
Dynamic Controllability, seen as the most useful controllability notion in
practice, is also the one that requires the most complicated algorithm.
Surprisingly, \citeA{mor:stnu} and \citeA{dc-rev05} proved DC is polynomial
in the size of the STPU representation.
In Figure~\ref{DCfig} the pseudocode of algorithm \DC is shown.

\begin{figure}[tbp]
\centering\begin{tabular}{|l|}
\hline
{\bfseries Pseudocode of} \DC \\ 
\hline
1. {\bf input} STPU W; \\
2. {\bf If} W is not pseudo-controllable {\bf then write} ``not
  DC'' and {\bf stop}; \\
3. Select all triangles ABC, C uncontrollable, A before C,\\ 
\hspace{3.5mm} such that the upper bound of the BC interval, $v$, is non-negative.\\ 
4. Introduce any tightenings required by the Precede case\\ 
\hspace{3.5mm} and any waits required by the Unordered case. \\
5. Do all possible regressions of waits,\\ \hspace{3.5mm} while converting 
unconditional waits to lower bounds.\\ \hspace{3.5mm} Also introduce 
lower bounds as provided by the general reduction. \\
6. If steps 3 and 4 do not produce any new (or tighter)\\ 
\hspace{3.5mm} constraints, then return true, otherwise go to 2.\\
\hline
\end{tabular}
\caption{\label{DCfig}
Algorithm \DC proposed in (Morris et al., 2001) 
for checking DC of an STPU.}
\end{figure}




In this paper we will extend the notion of dynamic controllability in
order to deal with preferences. The algorithm we will propose to test
this extended property will require
a good (even if not complete) understanding of the \DC algorithm.
Thus, we will now give the necessary details on this algorithm.

\begin{figure}[tb] 
\begin{center}
\resizebox{3.3in}{!}{\includegraphics{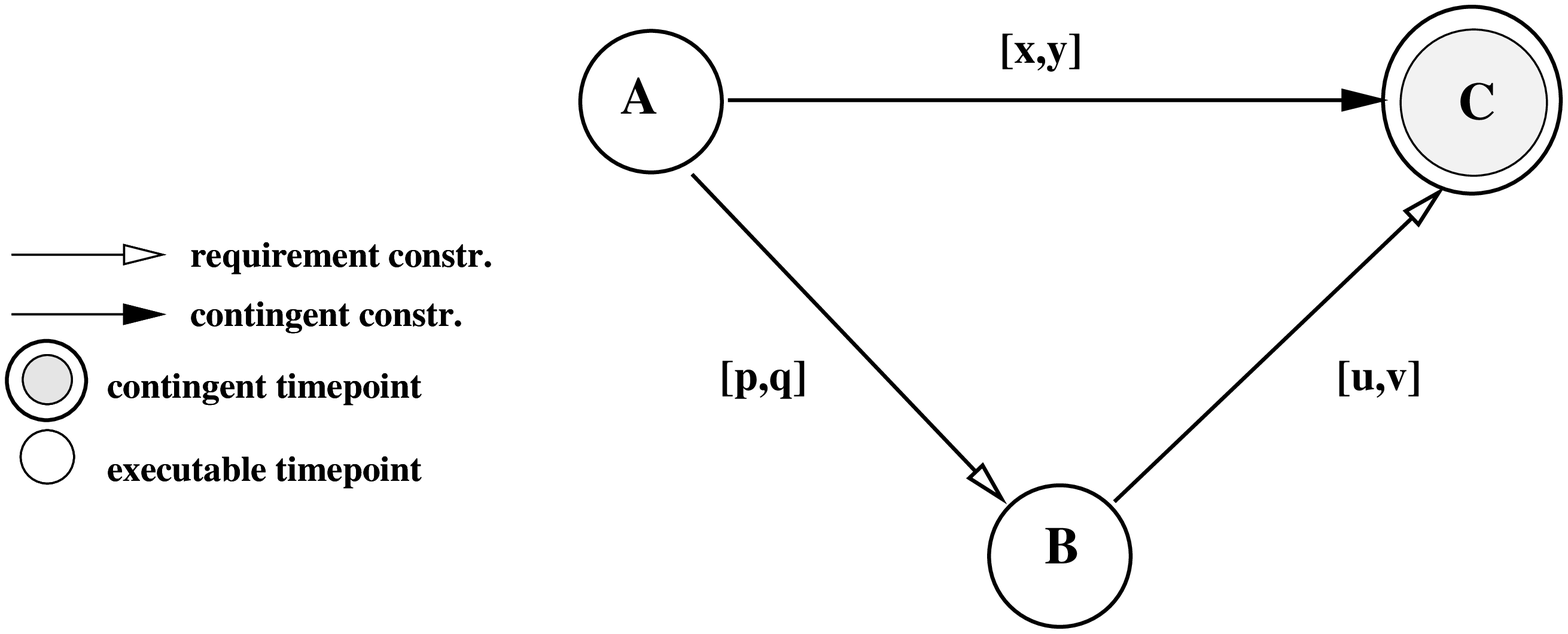}}
\end{center} 
\vspace{-0.8cm} 
\caption{\label{fig:triag}
A triangular STPU.}
\vspace{-0.3cm} 
\end{figure} 

As it can be seen, the algorithm is based on some considerations on
triangles of constraints.  The triangle shown in Figure~\ref{fig:triag} is
a triangular STPU with one contingent constraint, AC, two executable
time-points, A and B, and a contingent time-point C.  Based on the sign of
$u$ and $v$, three different cases can occur:

\begin{itemize}
\item {\it Follow case ($v<0$):} B will always follow C.  If the STPU is
  path consistent then it is also DC since, given the time at which C
  occurs after A, by definition of path consistency, it is always possible
  to find a consistent value for B.

\item {\it Precede case ($u \geq 0$):} B will always precede or happen
  simultaneously with C.  Then the STPU is dynamically controllable if $y-v
  \leq x-u$, and the interval $[p,q]$ on AB should be replaced by interval
  $[y-v,x-u]$, that is by the sub-interval containing all the elements of
  $[p,q]$ that are consistent with each element of $[x,y]$.
  
\item {\it Unordered case ($u<0$ and $v \geq 0$):} B can either follow or
  precede C.  To ensure dynamic controllability, B must wait either for C
  to occur first, or for $t=y-v$ units of time to go by after A.  In other
  words, either C occurs and B can be executed at the first value
  consistent with C's time, or B can safely be executed $t$ units of time
  after A's execution.  This can be described by an additional constraint
  which is expressed as a {\it wait} on AB and is written $<C,t>$, where
  $t=y-v$.  Of course if $x \geq y-v$ then we can raise the lower bound of
  AB, $p$, to $y-v$ ({\it Unconditional Unordered Reduction}), and in any
  case we can raise it to $x$ if $x>p$ ({\it General Unordered reduction})
  .
\end{itemize}

\begin{figure}[tb]
     \centering
     \subfigure[Regression 1]{
          \label{reg1}
          \includegraphics[width=.45\textwidth]{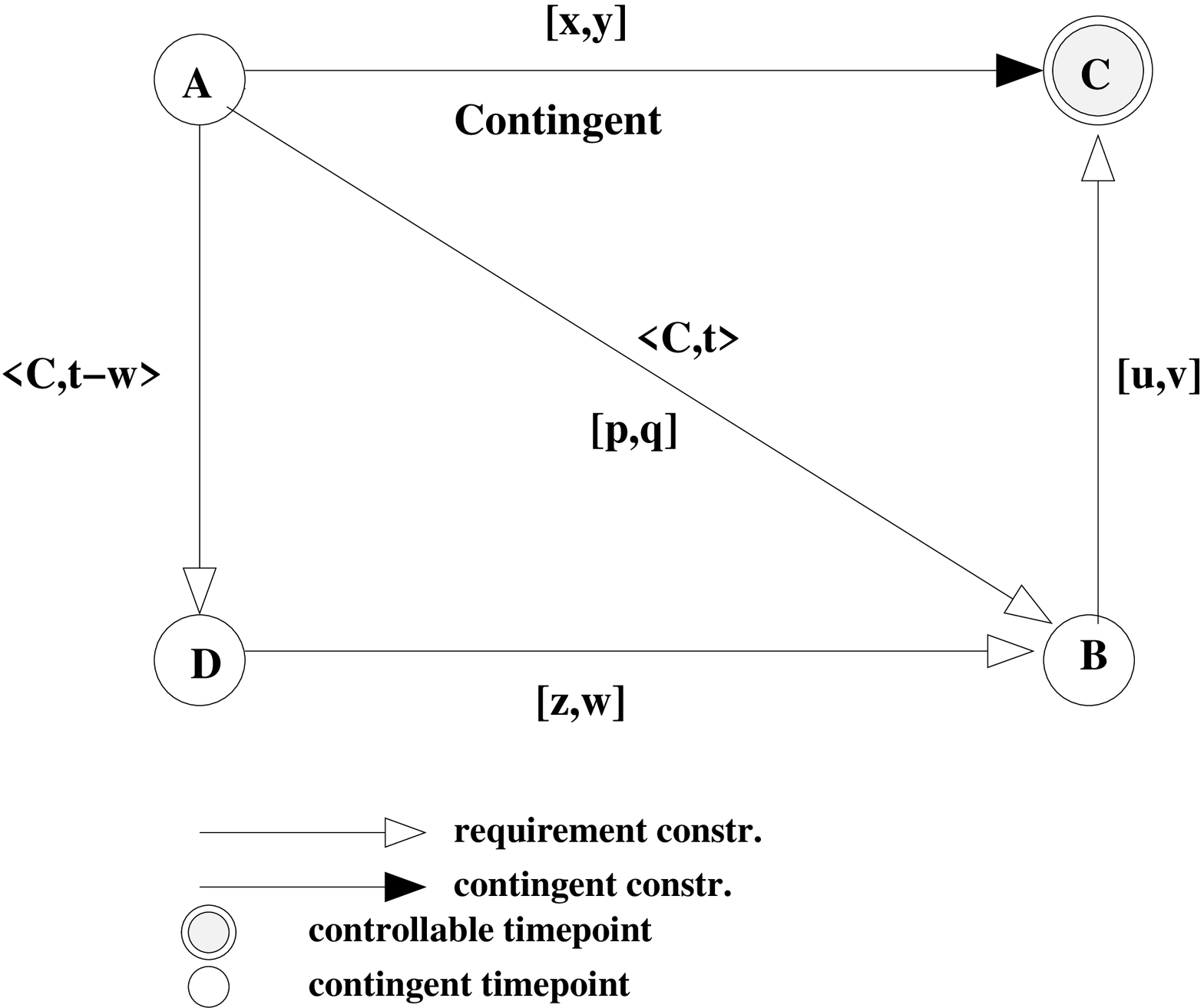}}
     \hspace{.3in}
     \subfigure[Regression 2]{
          \label{reg2}
          \includegraphics[width=.45\textwidth]{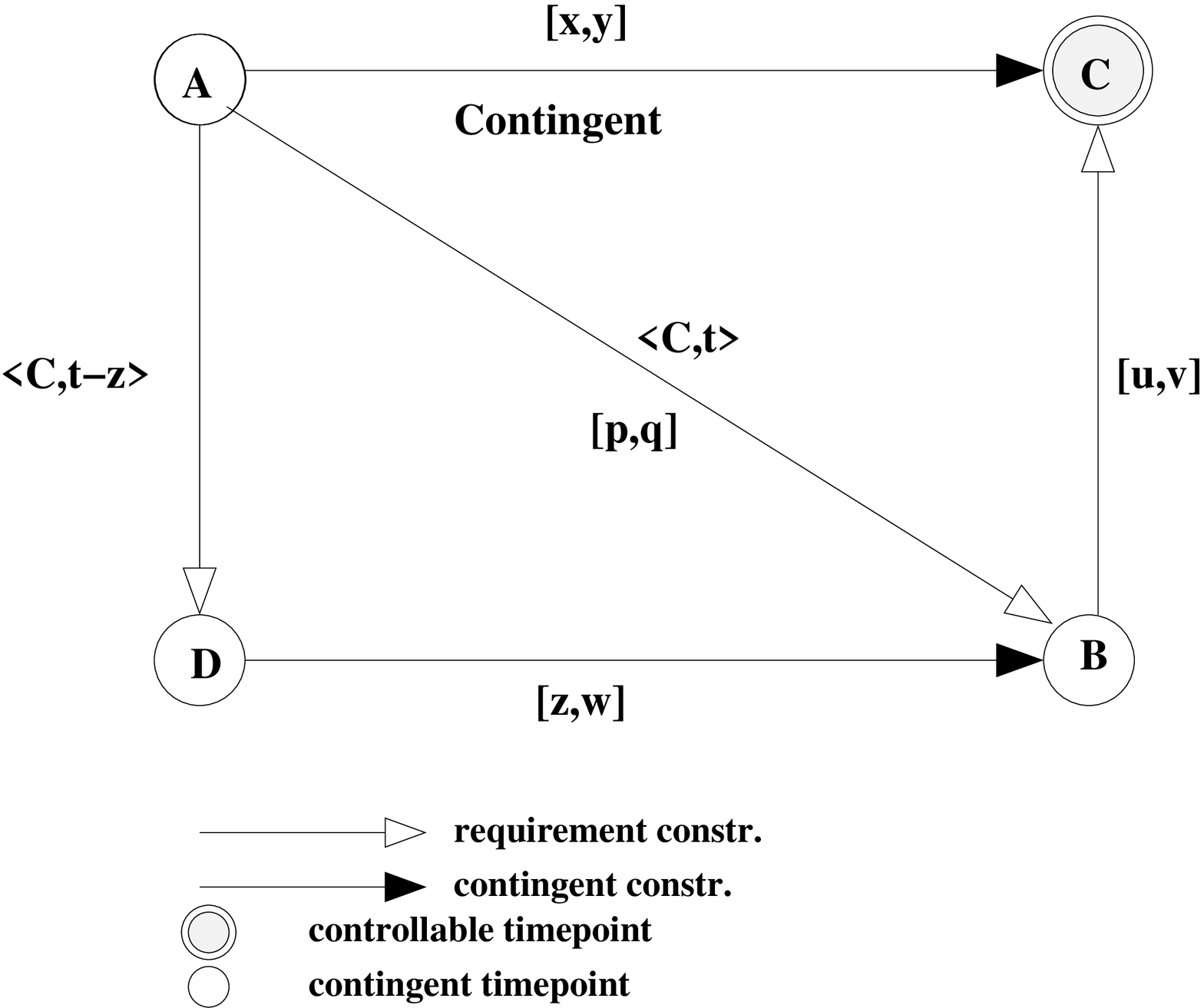}}
        \caption{Regressions in algorithm \DC.}
     \label{fig:2858multifig}
\end{figure}

It can be shown that waits can be propagated 
(in \citeauthor{mor:stnu}, 2001, the term
``regressed''is used ) from one constraint to another: a wait on AB induces a wait
on another constraint involving A, \eg AD, depending on the type of
constraint DB.  In particular, there are two possible ways in which the
waits can be regressed.
\begin{itemize}
\item \defn{Regression 1:} assume that the AB constraint has a wait $\langle
  C,t \rangle$. Then, if there is any DB constraint (including AB itself)
  with an upper bound, $w$, it is possible to deduce a wait $\langle C, t-w
  \rangle$ on $AD$.  Figure~\ref{reg1} shows this type of regression.
\item \defn{Regression 2:} assume that the AB constraint has a wait $\langle
  C,t \rangle$, where $t \geq 0$.  Then, if there is a contingent
  constraint DB with a lower bound, $z$, and such that $B \neq C$, it is
  possible to deduce a wait $\langle C, t-z \rangle$ on $AD$.  Figure
 ~\ref{reg2} shows this type of regression.
\end{itemize}

\begin{figure}[tb] 
\begin{center}
\resizebox{3.3in}{!}{\includegraphics{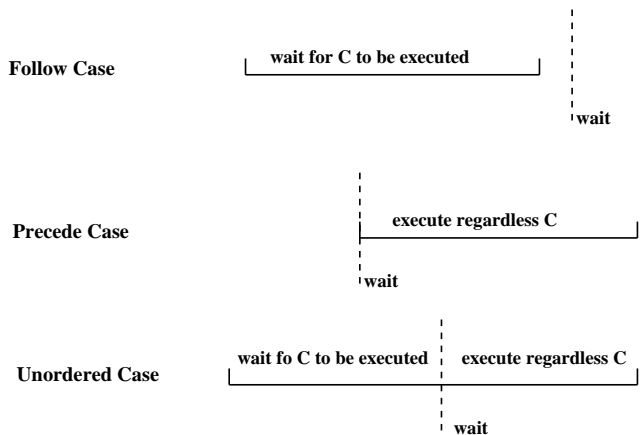}}
\end{center} 
\caption{The resulting AB interval constraint in the three cases considered by the \DC algorithm.}
\label{interv1}
\end{figure}

Assume for simplicity and without loss of generality that A is executed at
time 0.  Then, B can be executed before the wait only if C is executed
first. After the wait expires, B can safely be executed at any time left in
the interval.  As Figure~\ref{interv1} shows, it is possible to consider the
Follow and Precede cases as special cases of the Unordered. In the Follow
case we can put a ``dummy'' wait after the end of the interval, meaning that
B must wait for C to be executed in any case (Figure~\ref{interv1} (a)). In
the Precede case, we can set a wait that expires at the first element of
the interval meaning that B will be executed before C and any element in
the interval will be consistent with C (Figure~\ref{interv1} (b)). The
Unordered case can thus be seen as a combination of the two previous
states. The part of the interval before the wait can be seen as a Follow
case (in fact, B must wait for C until the wait expires), while the second
part including and following the wait can be seen as a Precede case (after
the wait has expired, B can be executed and any assignment to B that
corresponds to an element of this part of interval AB will be consistent
with any possible future value assigned to C).


The \DC algorithm applies these rules to all triangles in the STPU and
regresses all possible waits.  If no inconsistency is found, that is no
requirement interval becomes empty and no contingent interval is squeezed,
the STPU is DC and the algorithm returns an STPU where some constraints may
have waits to satisfy, and the intervals contain elements that appear in at
least one possible dynamic strategy.  This STPU can then be given to an
execution algorithm which dynamically assigns values to the executables
according to the current situation.

The pseudocode of the execution algorithm, \solver{DC-Execute}, is shown in
Figure~\ref{fig:dc:execution}.  The execution algorithm observes, as the
time goes by, the occurrence of the contingent events and accordingly
executes the controllables.  For any controllable B, its execution is
triggered if it is (1) \defn{live}, that is, if current time is within its
bounds, it is (2) \defn{enabled}, that is, all the executables constrained to
happen before have occurred, and (3) all the waits imposed by the
contingent time-points on B have expired.

\begin{figure}[tbp]
\centering\begin{tabular}{|l|}
\hline
{\bfseries Pseudocode for} {\solver{DC-Execute}}\\
\hline
1. {\bfseries input} STPU  $P$;\\
2. Perform initial propagation from the start time-point;\\
3.  {\bf repeat}\\
4.  immediately execute any executable time-points that\\
  \hspace{3.5 mm} have reached their upper bounds;\\
5.  arbitrarily pick an executable time-point $x$ that\\
   \hspace{3.5mm} is live and enabled and not yet executed, and whose waits,\\
   \hspace{3.5mm} if any, have all been satisfied;\\ 
6.  execute $x$;\\
7.  propagate the effect of the execution;\\
8.  {\bf if} network execution is complete {\bf then} return;\\
9.  {\bf else} advance current time,\\
   \hspace{3.5mm} propagating the effect of any contingent time-points that occur;\\
10. {\bf until} false;\\
\hline
\end{tabular}
\caption{\label{fig:dc:execution}
Algorithm that executes a dynamic strategy for an STPU.}
\end{figure}



\solver{DC-Execute} produces dynamically a consistent schedule on every
STPU on which algorithm \DC reports success \cite{mor:stnu}.  The complexity of the
algorithm is $O(n^3r)$, where $n$ is the number of variables and $r$ is the
number of elements in an interval. Since the 
polynomial complexity relies on the assumption of a bounded maximum
interval size,
\citeA{mor:stnu} conclude that \DC is
\emph{pseudo-}polynomial.  A DC algorithm of ``strong'' polynomial
complexity is presented in \cite{dc-rev05}.  The new algorithm differs from
the previous one mainly because it manipulates the distance graph rather
than the constraint graph of the STPU.  
It's complexity is $O(n^5)$.
What is important to notice for our
purposes is that, from the distance graph produced in output by the new
algorithm, it is possible to directly recover the intervals and waits of
the STPU produced in output by the original algorithm described in
\cite{mor:stnu}.


\section{Simple Temporal Problems with Preferences and Uncertainty (STPPUs)}
\label{sec:defn}

Consider a temporal problem that we would model naturally with preferences
in addition to hard constraints, but one also features uncertainty.
Neither an STPP nor an STPU is adequate to model such a problem.  Therefore
we propose what we will call \defn{Simple Temporal Problems with
  Preferences and Uncertainty}, or STPPUs for short.

Intuitively, an STPPU is an STPP for which time-points are partitioned into
two classes, requirement and contingent, just as in an STPU. Since some
time-points are not controllable by the agent, the notion of consistency of
an STP(P) is replaced by that of controllability, just as in an STPU. Every
solution to the STPPU has a global preference value, just as in an STPP, and
we seek a solution which maximizes this value, while satisfying
controllability requirements.

More precisely, we can extend some definitions given for STPPs and STPUs to
fit STPPUs in the following way.

\begin{definition}
In a context with preferences:

\begin{itemize}
\item an \defn{executable time-point} is a  variable, $x_i$, whose time
  is assigned by the agent;
  
\item a \defn{contingent time-point} is a variable, $e_i$, whose
  time is assigned by the external world;

  



 \item a \defn{soft requirement link} $r_{ij}$, on generic
  time-points $t_i$ and $t_j$ \footnote{Again, in general $t_i$ and $t_j$ can be 
  either contingent or executable time-points.}, 
  is a pair $\langle I_{ij}, f_{ij}
  \rangle$, where $I_{ij} = [l_{ij},u_{ij}]$ 
  such that $l_{ij} \leq
  \gamma(t_j)-\gamma(t_i) \leq u_{ij}$ where $\gamma(t_i)$ is a value 
  assigned to variable $t_i$, 
  and $f_{ij}:I_{ij} \rightarrow A$
  is a preference function mapping each element of
  the interval into an element of the preference set, $A$, of the semiring
  $S=\semiring$;
  
\item a \defn{soft contingent link} $g_{hk}$, on executable point $b_h$
  and contingent point $e_k$, is a pair $\langle {I}_{hk},
  {f}_{hk} \rangle$ where interval ${I}_{hk}=[{l}_{hk},{u}_{hk}]$
contains all the possible durations of the contingent event represented 
by $b_h$ and $e_k$  
  and ${f}_{hk}:{I}_{hk}
  \rightarrow A$ is a preference function that maps
  each element of the interval into an element of the preference set $A$.$\Box$
\end{itemize}
\end{definition}

In both types of constraints, the preference function represents the
preference of the agent on the duration of an event or on the distance
between two events.  However, while for soft requirement
constraints the agent has control and can be guided by the preferences in
choosing values for the time-points, for soft contingent constraints the
preference represents merely a desire of the agent on the possible outcomes
of Nature: there is no control on the outcomes.
It should be noticed that in STPPUs 
uncertainty is modeled, just like in STPUs, assuming 
``complete ignorance'' on when events are more likely to happen.
Thus, all durations of contingent events are assumed 
to be equally possible (or plausible) and different levels of
plausibility  are not allowed.

We can now state formally the definition of STPPUs, which combines
preferences from the definition of an STPP with contingency from the
definition of an STPU. 

\begin{definition}[STPPU]
  A \defn{Simple Temporal Problem with Preferences and Uncertainty} (STPPU)
  is a tuple $P=(N_e, N_c, L_r, L_c,S)$ where:
  \begin{itemize}
   \item $N_e$ is the set of executable time-points;
   \item $N_c$ is the set of contingent time-points;
   \item $S = \semiring$ is a c-semiring;
   \item $L_r$ is the set of soft requirement constraints over S;
   \item $L_c$ is the set of soft contingent constraints over S.$\Box$
  \end{itemize}
\end{definition}

Note that, as STPPs, also STPPUs can model hard constraints by
soft constraints in which each element of the
interval is mapped into the maximal element of the preference set.
Further, without loss of generality, and
following the assumptions made for STPUs \cite{mor:stnu}, we 
assume that no two contingent constraints end at the same time-point.

Once we have a complete assignment to all time-points we can compute its
global preference, as in STPPs.  This is done according to the
semiring-based soft constraint schema: first we project the assignment on
each soft constraint, obtaining an element of the interval and the
preference associated to that element; then we combine the preferences
obtained on all constraints with the multiplicative operator of the
semiring.
Given two assignments with their preference, the best is chosen using the additive
operator.
An assignment is \defn{optimal} if there is no other assignment with a
preference which is better in the semiring's ordering.

In the following we summarize some of the definitions given for STPUs,
extending them directly to STPPUs.

\begin{definition}
Given an STPPU $P$:
\begin{itemize}
\item A \defn{schedule} is a complete assignment to all the time-points in
  $N_e$ and $N_c$;
  
\item \defn{Sched(P)} is the set of all schedules of $P$; while
  \defn{Sol(P)} the set of all schedules of $P$ that are consistent with
  all the constraints of $P$ (see Definition~\ref{stc},
  Section~\ref{TCSPPs}); 
  
\item Given a schedule $s$ for $P$, a \defn{situation} (usually written
  $\omega_s$) is the set of durations of all contingent constraints in $s$;
  
\item Given a schedule $s$ for $P$, a \defn{control sequence} (usually
  written $\delta_s$ is the set of assignments to executable time-points in
  $s$;
  
\item $T_{\delta,\omega}$ is a schedule such that $[T_{\delta,\omega}]_x
  =[\delta]_x$\footnote{Regarding notation, as in the case with hard
    constraints, given an executable time-point $x$, we will write
    $[S(P_{\omega})]_x$ to indicate the value assigned to $x$ in
    $S(P_{\omega})$, and $[S(P_{\omega})]_{<x}$ to indicate the durations
    of the contingent events that finish prior to $x$ in $S(P_{\omega})$.},
  $\forall x \in N_e$, and for every contingent constraint, $g_{hk}\in
  L_c$, defined on executable $b_h$ and contingent time-point $e_k$,
  $[T_{\delta,\omega}]_{e_k}$-$[T_{\delta,\omega}]_{b_h}=\omega_{hk}$,
  where $\omega_{hk}$ is the duration of $g_{hk}$ in $\omega$;
  
\item A \defn{projection $P_{\omega}$} corresponding to a situation
  $\omega$ is the STPP obtained from $P$ by 
leaving all requirement constraints
  unchanged and replacing each contingent constraint $g_{hk}$ with the soft
  constraint $\langle [{\omega}_{hk},{\omega}_{hk}],
  f({\omega}_{hk}) \rangle$, where ${\omega}_{hk}$ is the duration of the
  event represented by $g_{hk}$ in $\omega$, and $f({\omega}_{hk})$ is the
  preference associated to such duration;

\item Given a projection $P_{\omega}$ we indicate with 
$Sol(P_{\omega})$ the set of solutions of $P_{\omega}$ 
and we define $OptSol(P_\omega)=\{s \in Sol(P_\omega) | \not \exists s' \in
  Sol(P_\omega)$, $pref(s') > pref(s) \}$; if the set of preferences 
is totally ordered 
we indicate with $opt(P_{\omega})$ the preference of any optimal solution of
$P_{\omega}$; 

\item \defn{Proj(P)} is the set of all projections of an STPPU P;
  
\item A \defn{strategy} $s$ is a map $s$ : $Proj(P) \rightarrow Sched(P)$
  such that for every projection $P_{\omega}$, $s(P_{\omega})$ is a
  schedule which includes $\omega$;
  
\item A strategy is \defn{viable} if $\forall \omega$, $S(P_{\omega})$ is a
  solution of $P_{\omega}$, that is, if it satisfies all its soft temporal
  constraints.  Thus a viable strategy is a mapping $S:$ $Proj(P)
  \longrightarrow Sol(P)$.  In this case we indicate with
  $pref(S(P_{\omega}))$ the global preference associated to schedule
  $S(P_{\omega})$ in STPP $P_{\omega}$.$\Box$
\end{itemize}

\end{definition}

\begin{figure}[tb] 
\begin{center}
\resizebox{3.1in}{!}{\includegraphics{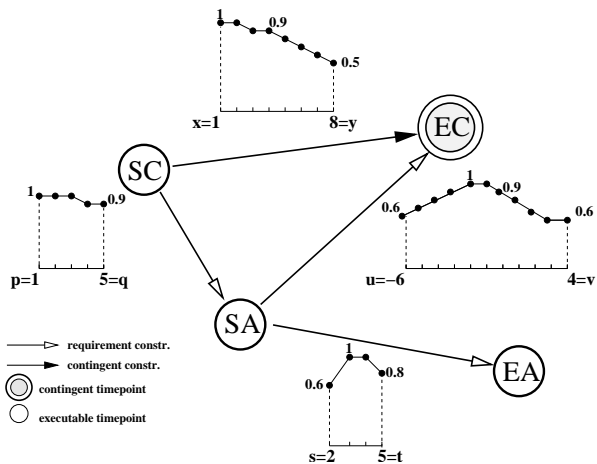}}
\end{center} 
\caption{Example STPPU from the Earth Observing Satellites domain.}
\label{fig:eos}
\end{figure} 

\begin{example}\label{eosexample}
  Consider as an example the following scenario from the Earth Observing
  Satellites domain \cite{fra:fleets} described in Section~\ref{sec:motiv}.
  Suppose a request for observing a region of interest has been received and
  accepted.  To collect the data, the instrument must be aimed at the
  target before images can be taken.  It might be, however, that for a
  certain period during the time window allocated for this observation, the
  region of interest is covered by clouds.  The earlier the cloud coverage
  ends the better, since it will maximise both the quality and the quantity
  of retrieved data; but coverage is not controllable.
     
  Suppose the time window reserved for an observation is from $1$ to $8$
  units of time and that we start counting time when the cloud occlusion on
  the region of interest is observable.  Also, suppose, in order for the
  observation to succeed, the aiming procedure must start before $5$ units
  after the starting time, ideally before $3$ units, and it actually can
  only begin after at least $1$ time unit after the weather becomes
  observable.  Ideally the aiming procedure should start slightly after
  the cloud coverage will end.  If it starts too early, then, since the
  instrument is activated immediately after it is aimed, clouds might still
  occlude the region and the image quality will be poor.  On the other
  hand, if it waits too long after 
  the clouds have disappeared then precious time
  during which there is no occlusion will be wasted aiming the instrument
  instead of taking images.  The aiming procedure can be controlled by the
  mission manager and it can take anywhere between $2$ and $5$ units of
  time.  An ideal duration is $3$ or $4$ units, since a short time of $2$
  units would put the instrument under pressure, while a long duration,
  like $5$ units, would waste energy.
    
  This scenario, rather tedious to describe in words, can be compactly
  represented by the STPPU shown in Figure~\ref{fig:eos} with the
  following features:

  \begin{itemize}
  \item a set of executable time-points $\event{SC}$ (Start Clouds), 
   $\event{SA}$ (Start Aiming), $\event{EA}$ (End Aiming);
  \item a contingent time-point $\event{EC}$ (End Clouds);
  \item a set of soft requirement constraints on $\{\event{SC}
    \rightarrow \event{SA},$ $\event{SA} \rightarrow \event{EC},$
    $\event{SA} \rightarrow \event{EA}\}$;
  \item a soft contingent constraint $\{\event{SC} \rightarrow
    \event{EC}\}$;
  \item the fuzzy semiring $S_{\text{FCSP}} = \langle [0,1], \max, \min,
    0, 1 \rangle$.
  \end{itemize}
  
  A solution of the STPPU in Figure~\ref{fig:eos} is the schedule
  $s =$ $\{\event{SC} = 0,$ $\event{SA} = 2,$ $\event{EC} = 5,$
  $\event{EA}=7\}$.  The situation associated with $s$ is the projection on
  the only contingent constraint, $\event{SC} \rightarrow \event{EC}$,
  \ie $\omega_s=5$, while the control sequence is the assignment to the
  executable time-points, \ie $\delta_s = \{\event{SC}=0, \event{SA}=2,
  \event{EA}=7\}$.  The global preference is obtained by considering the
  preferences associated with the projections on all constraints, that is
  $\pref{2}=1$ on $\event{SC} \rightarrow \event{SA}$, $\pref{3}=0.6$ on
  $\event{SA} \rightarrow \event{EC}$, $\pref{5}=0.9$ on $\event{SA}
  \rightarrow \event{EA}$, and $\pref{5}=0.8$ on $\event{SC} \rightarrow
  \event{EC}$.  The preferences must then be combined using the
  multiplicative operator of the semiring, which is $\min$, so the global
  preference of $s$ is $0.6$.  Another solution $s' =$ $\{\event{SC}=0,$
  $\event{SA}=4,$ $\event{EC}=5,$ $\event{EA}=9\}$ has global preference
  $0.8$.  Thus $s'$ is a better solution than $s$ according to the semiring
  ordering since $\max(0.6,0.8)=0.8$.$\Box$ 
\end{example}


\section{Controllability with Preferences}
\label{contdef}

We now consider how it is possible to extend the notion of controllability
to accommodate preferences.  In general we are interested in the ability of
the agent to execute the time-points under its control, not only subject to
all constraints but also in the best possible way \wrt preferences.

It transpires that the meaning of `best possible way' depends on the types of
controllability required.  In particular, the concept of optimality must be
reinterpreted due to the presence of uncontrollable events.  In fact, the
distinction on the nature of the events induces a difference on the meaning
of the preferences expressed on them, as mentioned in the previous section.
Once a scenario is given it will  
have  a certain level of desirability, expressing how much the agent
likes such a situation. 
Then, the
agent often has several choices for the events he controls that are
consistent with that scenario. 
Some of these choices might be preferable
with respect to others. This is expressed by the preferences on the
requirement constraints and such information should guide the agent in
choosing the best possible actions to take.  Thus, the concept of
optimality is now `relative' to the specific scenario. 
The final preference
of a complete assignment is an overall value which combines how much the
corresponding scenario is desirable for the agent and how well the agent
has reacted in that scenario.  

The concepts of controllability we will
propose here are, thus, based on the possibility of the agent to execute
the events under her control in the best possible way given the actual
situation. Acting in an optimal way can be seen as not lowering
further the preference given by the uncontrollable events.

\subsection{Strong Controllability with Preferences}

We start by considering the strongest notion of controllability.  We extend
this notion, taking into account preferences, in two ways, obtaining
\defn{Optimal Strong Controllability} and \defn{$\alpha$-Strong
  Controllability}, where $\alpha \in A$ is a preference level.  As we will see,
the first notion corresponds to 
a stronger requirement, since it assumes the existence of a
fixed unique assignment for all the executable time-points that is optimal
in every projection.  The second notion requires such a fixed assignment to
be optimal only in those projections that have a maximum preference value
not greater than $\alpha$, and to yield a preference $\not < \alpha$
in all other cases.


\begin{definition}[Optimal Strong Controllability]
  An STPPU $P$ is \defn{Optimally Strongly Controllable} (OSC) iff there is
  a viable execution strategy $S$ s.t.
\begin{enumerate}
\item $[S(P_1)]_x=[S(P_2)]_x$, $\forall P_1, P_2 \in Proj(P)$
  and for every executable time-point $x$;
\item $S(P_{\omega})\in OptSol(P_\omega)$, 
      $\forall P_{\omega} \in Proj(P)$. $\Box$ 
\end{enumerate}
\end{definition}

In other words, an STPPU is OSC if there is a fixed control sequence that
works in all possible situations and is optimal in each of them.  In the
definition, `optimal' means that there is no other assignment the agent can
choose for the executable time-points that could yield a higher preference
in any situation.  Since this is a powerful restriction, as mentioned
before, we can instead look at just reaching a certain quality threshold:

\begin{definition}[$\alpha$-Strong Controllability]\label{asc}
  An STPPU $P$ is \defn{$\alpha$-Strongly Controllable} ($\alpha$-SC), with
  $\alpha \in A$ a preference, iff there is a viable strategy $S$ s.t.
  \begin{enumerate} 
  \item $[S(P_1)]_x=[S(P_2)]_x$, $\forall P_1, P_2 \in Proj(P)$
   and for every executable time-point $x$;
  \item  $S(P_{\omega})\in OptSol(P_\omega)$,$\forall P_{\omega} \in Proj(P)$ such that $\not \exists s' \in
  OptSol(P_\omega)$ with $pref(s') > \alpha$;
  \item $pref(S(P_{\omega})) \not < \alpha$ otherwise.$\Box$
  \end{enumerate}
\end{definition}

In other words, an STPPU is 
$\alpha$-SC if there is a fixed control sequence that works in all
situations and results in optimal schedules for those situations where the
optimal preference level of the projection is 
not $> \alpha$
in a schedule with preference not smaller than  $\alpha$ in all other cases.  

\subsection{Weak Controllability with Preferences}

Secondly, we extend similarly the least restrictive notion of
controllability.  Weak Controllability requires the existence of a solution
in any possible situation, possibly a different one in each situation.  We
extend this definition by requiring the existence of an optimal solution in
every situation.
   
\begin{definition}[Optimal Weak Controllability]
  An STPPU $P$ is \defn{Optimally Weakly Controllable}  (OWC) iff $\forall
  P_{\omega} \in Proj(P)$ there is a strategy $S_\omega$ s.t.\ 
  $S_\omega(P_{\omega})$ is an optimal solution of $P_{\omega}$.$\Box$
\end{definition}

In other words, an STPPU is OWC if, for every situation, there is a 
control sequence that results in an optimal schedule for that situation.

Optimal Weak Controllability of an STPPU is equivalent to Weak
Controllability of the corresponding STPU obtained by ignoring preferences, 
as we will formally prove in Section~\ref{owc}.
The reason is that if a projection $P_\omega$ has at least one solution
then it must have an optimal solution. 
Moreover, any STPPU is such that its
underlying STPU is either WC or not.  Hence it does not make sense to
define a notion of $\alpha$-Weak Controllability. 

\subsection{Dynamic Controllability with Preferences}

Dynamic Controllability (DC) 
addresses the ability of the agent to execute a schedule by choosing
incrementally the values to be assigned to executable time-points, looking
only at the past.  When preferences are available, it is desirable that the
agent acts not only in a way that is guaranteed to be consistent with any
possible future outcome but also in a way that ensures the absence of
regrets w.r.t. preferences.

\begin{definition}[Optimal Dynamic Controllability]
  An STPPU $P$ is \defn{Optimally Dynamically Controllable} (ODC) 
  iff there is a viable strategy $S$ such that $\forall P_1, P_2 \in Proj(P)$ 
  and for any executable time-point $x$:
  \begin{enumerate}
  \item if $[S(P_1)]_{<x} = [S(P_2)]_{<x}$ then $[S(P_1)]_{x} =
    [S(P_2)]_{x}$;
  \item $S(P_1) \in OptSol(P_1)$ and $S(P_2)=OptSol(P_2)$.$\Box$
  \end{enumerate}
  \label{ODC:defn}
\end{definition}

In other words, an STPPU is ODC if there exists a means of extending any
current partial control sequence to a complete control sequence in the
future in such a way that the resulting schedule will be optimal.  As
before, we also soften the optimality requirement to having a preference
reaching a certain threshold.

\begin{definition}[$\alpha$-Dynamic Controllability]
  An STPPU $P$ is \defn{$\alpha$-Dynamically Controllable} ($\alpha$-DC) iff
  there is a viable strategy $S$ such that $\forall P_1, P_2 \in
  Proj(P)$ and for every executable time-point $x$: 
  \begin{enumerate}
  \item if $[S(P_1)]_{<x} = [S(P_2)]_{<x}$ then $[S(P_1)]_{x} =
    [S(P_2)]_{x}$;
  \item $S(P_1) \in OptSol(P_1)$ and $S(P_2) \in OptSol(P_2)$ 
    if $\not \exists s_1 \in
  OptSol(P_1)$ with $pref(s_1) > \alpha$ and $\not \exists s_2 \in
  OptSol(P_2)$ with $pref(s_2) > \alpha$;
 \item $\pref{S(P_1)} \not <\alpha$ and
    $\pref{S(P_2)}\not < \alpha$ otherwise.$\Box$
  \end{enumerate}
\label{alphaDC:defn}
\end{definition}

In other words, an STPPU is $\alpha$-DC if there is a means of extending
any current partial control sequence to a complete sequence; but optimality
is guaranteed only for situations with preference $\not > \alpha$. For all
other projections the resulting dynamic schedule will have preference at
not smaller than $\alpha$.

\subsection{Comparing the Controllability Notions}
\label{compcont}

We will now consider the relation among the different notions of
controllability for STPPUs.

Recall that for STPUs, $SC \implies DC \implies WC$ (see Section
\ref{stppu-back}).  We start by giving a similar result that holds for the
definitions of optimal controllability with preferences.  Intuitively, if
there is a single control sequence that will be optimal in all situations,
then clearly it can be executed dynamically, just assigning the values in
the control sequence when the current time reaches them. Moreover if,
whatever the final situation will be, we know we can consistently assign
values to executables, just looking at the past assignments, and never
having to backtrack on preferences, then it is clear that every situation
has at least an optimal solution.

\begin{theorem}
\label{OSCiODCiOWC}
If an STPPU $P$ is OSC, then it is ODC; if it is ODC, then it is OWC.
\end{theorem}

Proofs of theorems are given in the appendix.  The opposite implications of
Theorem~\ref{OSCiODCiOWC} do not hold in general.  It is in fact sufficient
to recall that hard constraints are a special case of soft constraints and
to use the known result for STPUs \cite{mor:stnu}.

\begin{figure}[tbp] 
\begin{center} \ \setlength{\epsfxsize}{4.3in} 
\epsfbox{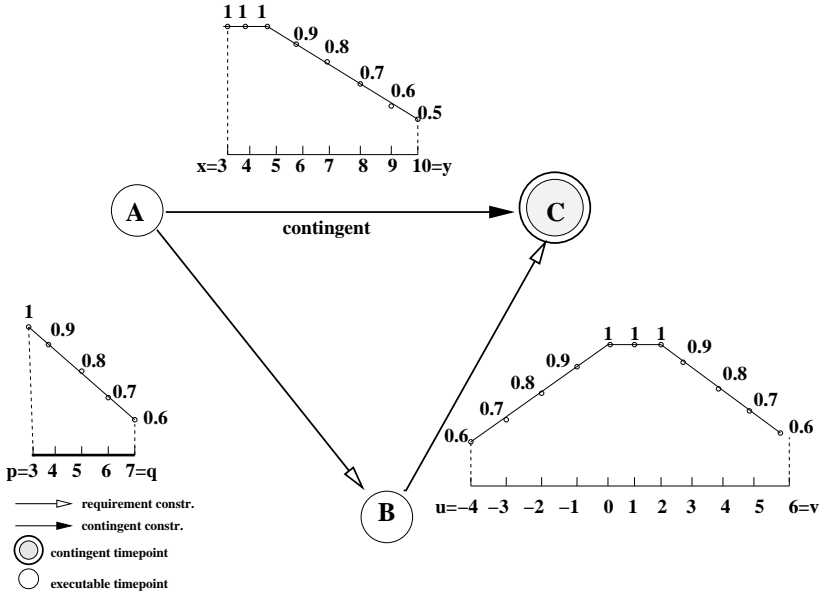} 
\end{center} 
\caption{An STPPU which is OWC but not ODC.}\label{fail5} 
\vspace{-0.3cm} 
\end{figure}

As examples consider the following two, both defined on the fuzzy
semiring.  Figure~\ref{fail5} shows an STPPU which is OWC but is not ODC.
It is, in fact, easy to see that any assignment to A and C, which is a
projection of the STPPU can be consistently extended to an assignment of B.
However, we will show in Section~\ref{odc} that the STPPU depicted is not
ODC.

\begin{figure}[tbp] 
\begin{center} \ \setlength{\epsfxsize}{2.3in} 
\epsfbox{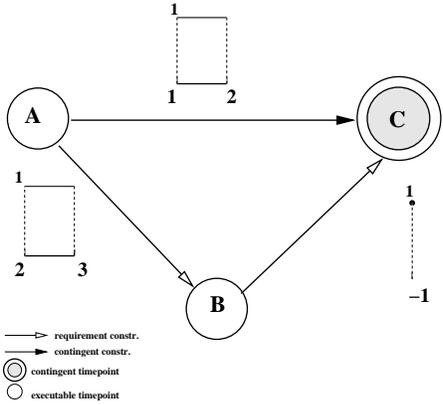} 
\end{center} 
\caption{\label{nosc}
An STPPU which is ODC but not OSC.}
\vspace{-0.3cm} 
\end{figure} 

Figure~\ref{nosc}, instead, shows an ODC STPPU which is not OSC.
A and B are two executable time-points and C is a contingent time-point.
There are only two projections, say $P_1$ and $P_2$, corresponding
respectively to point 1 and point 2 in the AC interval.  The optimal
preference level for both is 1. In fact, $\langle A=0, C=1, B=2 \rangle$ is
a solution of $P_1$ with preference 1 and $\langle A=0, C=2, B=3 \rangle$
is a solution of $P_2$ with preference 1.  The STPPU is ODC. In fact, there
is a dynamic strategy $S$ that assigns to B value 2, if C occurs at 1, and
value 3, if C occurs at 2 (assuming A is always assigned 0). However there
is no single value for B that will be optimal in both scenarios.

Similar results apply in the case of $\alpha$-controllability, 
as the following formal treatment shows.

\begin{theorem}
\label{aSCiaDC}
For any given preference level $\alpha$, if an STPPU $P$ is $\alpha$-SC
then it is $\alpha$-DC.
\end{theorem}

Again, the converse does not hold in general.  As an example consider again
the STPPU shown in Figure~\ref{nosc} and $\alpha=1$.  Assuming $\alpha=1$,
such an STPPU is 1-DC but, as we have shown above, it is not $1$-SC.

Another useful  result is that if a controllability property holds 
at a given preference level, say $\beta$, then it holds also 
$\forall \alpha < \beta$, as stated in the following theorem. 

\begin{theorem}
\label{lowerp}
Given an STPPU $P$ and a preference level $\beta$, 
if $P$ is  $\beta$-SC (resp. $\beta$-DC), then it is 
$\alpha$-SC (resp. $\alpha$-DC), $\forall \alpha < \beta$. 
\end{theorem}

Let us now consider case in which the preference set is totally ordered. 
If we eliminate the uncertainty in an STPPU, by regarding all contingent
time-points as executables, we obtain an STPP.  Such an STPP can be solved
obtaining its optimal preference value $opt$.  This preference level,
$opt$, will be useful to relate optimal controllability to
$\alpha$-controllability.  As stated in the following theorem, an STPPU is
optimally strongly or dynamically controllable if and only if it satisfies
the corresponding notion of $\alpha$-controllability at $\alpha=opt$.

\begin{theorem}
\label{oopt}
Given an STPPU $P$ defined on a c-semiring with totally ordered
preferences, let $opt=max_{T \in Sol(P)}pref(T)$. Then, 
P is OSC (resp. ODC) iff it is $opt$-SC (resp. $opt$-DC).
\end{theorem}

For OWC, we will formally prove in Section~\ref{owc} that
an STPPU is OWC iff the STPU obtained by ignoring the preference
functions is WC.
As for the relation between 
$\alpha_{min}$-controllability and controllability without preferences, we
recall that considering the elements of the intervals mapped in a
preference $\geq \alpha_{min}$ coincides by definition to considering the
underlying STPU obtained by ignoring the preference functions of the
STPPU. Thus, $\alpha_{min}$-X holds iff X holds, where X is either SC or DC.

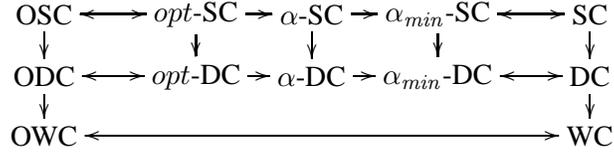
\begin{figure}[tbp]
\vspace{-0.3cm}
\begin{center}
$$
\xymatrix@-=8pt{
\text{OSC}\ar@{<->}[rr]\ar@{->}[d] & & \text{$opt$-SC}\ar@{->}[r]
\ar@{->}[d]
 &
\text{$\alpha$-SC}\ar@{->}[r]\ar@{->}[d] &
\text{$\alpha_{\mathit{min}}$-SC}\ar@{<->}[rr]\ar@{->}[d]  & &
\text{SC}\ar@{->}[d]\\ 
\text{ODC}\ar@{<->}[rr]\ar@{->}[d] & & \text{$opt$-DC}\ar@{->}[r] &
\text{$\alpha$-DC}\ar@{->}[r] &
\text{$\alpha_{\mathit{min}}$-DC}\ar@{<->}[rr] &
&
\text{DC}\ar@{->}[d]\\
\text{OWC}\ar@{<->}[rrrrrr] & & & & & & \text{WC}
}
$$
\end{center}
\caption{Comparison of controllability notions
for total orders. $\alpha_{\mathit{min}}$ is the smallest
  preference over any constraint: $opt \geq \alpha \geq
\alpha_{\mathit{min}}
$.}
\vspace{-0.3cm}
\label{fig:schema}
\end{figure}

In Figure~\ref{fig:schema} we summarize the relationships 
holding among the various controllability notions when preferences
are totally ordered. When instead they are partially ordered, 
the relationships $opt-X$ and $\alpha_{min}-X$, where 
$X$ is a controllability notion, do not make sense.
In fact, in the partially ordered case, there can be several 
optimal elements and several minimal elements, not just one.


\section{Determining Optimal Strong Controllability and $\alpha$-Strong Controllability}
\label{osc}

In the next sections we give methods to determine which levels of
controllability hold for an STPPU.  Strong Controllability fits when
off-line scheduling is allowed, in the sense that the fixed optimal control
sequence is computed before execution begins.  This approach is reasonable
if the planning algorithm has no knowledge on the possible outcomes, other
than the agent's preferences. Such a situation requires us to find a fixed way
to execute controllable events that will be consistent with any possible
outcome of the uncontrollables and that will give the best possible final
preference.

\subsection{Algorithm \bestsc}

The algorithm described in this section checks whether an STPPU is OSC.  If
it is not OSC, the algorithm will detect this and will also return the
highest preference level $\alpha$ such that the problem is $\alpha$-SC.

All the algorithms we will present in this paper rely on the following
tractability assumptions, inherited from STPPs: (1) the underlying semiring
is the fuzzy semiring $S_{FCSP}$ defined in Section~\ref{TCSPPs}, (2) the
preference functions are semi-convex, and (3) the set of preferences
$[0,1]$ is discretized in a finite number of elements according to a given
granularity.

The algorithm \bestsc is based on a simple idea: for each preference level
$\beta$, it finds all the control sequences that guarantee strong
controllability for all projections such that their optimal preference is
$\geq \beta$, and optimality for those with optimal preference $\beta$.
Then, it keeps only those control sequences that do the same for all
preference levels $> \beta$.

\begin{figure}[tbp]
\centering\begin{tabular}{|l|}
\hline
{\bfseries Pseudocode for} \textsf{\bfseries{Best-SC}}\\ \hline
1. {\bfseries input} STPPU $P$;\\
2. compute $\alpha_{min}$;\\
3. STPU $Q^{\alpha_{min}} \receives$ $\alpha_{min}$-\solverns{Cut}($P$);\\
4. {\bf if} (\scstpu($Q^{\alpha_{min}}$) inconsistent) {\bf write} ``not $\alpha_{min}$-SC'' and {\bf stop};\\
5. {\bf else} \{ \\
6. \hspace{2mm} STP $P^{\alpha_{min}}$ $\receives$ \scstpu($Q^{\alpha_{min}}$);\\
7. \hspace{2mm} preference $\beta \receives \alpha_{min}+1$;\\
8. \hspace{2mm} bool OSC$\receives$false, bool $\alpha$-SC$\receives$false;\\
9. \hspace{2mm} {\bf do} \{\\ 
10. \hspace{4mm} STPU $Q^{\beta} \receives$ $\beta$-\solverns{Cut}($P$);\\
11. \hspace{4mm} {\bf if} (\pca($Q^{\beta}$) inconsistent) OSC$\receives$true;\\
12. \hspace{4mm} {\bf else} \{ \\
13.  \hspace{6mm}{\bf if} (\scstpua(\pca($Q^{\beta})$) inconsistent)
$\alpha$-SC$\receives$ true; \\
14.  \hspace{6mm}{\bf else} \{\\
15.   \hspace{8mm} STP $P^{\beta}$ $\receives$ $P^{\beta-1} \bigotimes $ \scstpua(\pca($Q^{\beta})$) ;\\
16.   \hspace{8mm} {\bf if} ($P^{\beta}$ inconsistent) \{ $\alpha$-SC
$\receives$ true \};\\
17.   \hspace{8mm} {\bf else} \{ $\beta \receives \beta+1$ \};\\
18.   \hspace{12mm} \}\\
19.   \hspace{10mm}\}\\
20.   \hspace{6mm}\}{\bf while} (OSC=false and $\alpha$-SC=false);\\
21. \hspace{2mm}{\bf if} (OSC=true) {\bf write} ``P is OSC'';\\
22. \hspace{2mm}{\bf if} ($\alpha$-SC=true) {\bf write} ``P is'' $(\beta-1)$ ''-SC'';\\
23. \hspace{2mm} $s_e$=\solverns{Earliest-Solution($P^{\beta-1}$)}, 
     $s_l$=\solverns{Latest-Solution($P^{\beta-1}$)};\\    
24. \hspace{2mm}{\bf return} $P^{\beta-1}$, $s_e$, $s_l$;\\
25. \hspace{3mm} \};\\
\hline
\end{tabular}
\caption{\label{hasc} 
Algorithm \bestsc: it tests if an STPPU is OSC and 
finds the highest $\alpha$ such that STPPU $P$ is $\alpha$-SC.}
\end{figure}
   
The pseudocode is shown in Figure~\ref{hasc}.  The algorithm takes in input
an STPPU $P$ (line 1).  As a first step, the lowest preference
$\alpha_{min}$
is computed.  Notice that, to do this
efficiently, the analytical structure of the preference functions
(semi-convexity) can be exploited.


In line 3 the STPU obtained from $P$ by cutting it at preference level
$\alpha_{min}$ is considered.  Such STPU is obtained by applying function
$\alpha_{min}$-\solverns{Cut}(STPPU $G$) with $G$=$P$
\footnote{Notice that function $\beta$-\solverns{Cut} can be
  applied to both STPPs and STPPUs: in the first case the output is an STP,
  while in the latter case an STPU. Notice also that,
  $\alpha$-\solverns{Cut} is a known concept in fuzzy literature.}.
In general, the result of 
$\beta$-\solverns{Cut}($P$)
is the STPU $Q^{\beta}$
(\ie, a temporal problem with uncertainty but not preferences) defined as
follows:
\begin{itemize}
\item $Q^{\beta}$ has the same variables with the same domains as in P;
\item for every soft temporal constraint (requirement or contingent) 
in $P$ on variables $X_i$, and $X_j$, say  
$c=\langle I, f \rangle$, there is, 
in $Q^{\beta}$, a simple temporal constraint on the same variables defined as 
$\{x \in I | f(x) \geq \beta\}$.
\end{itemize}  
Notice that the semi-convexity of the preference functions guarantees that
the set $\{x \in I | f(x) \geq \beta\}$ forms an interval.  The
intervals in $Q^{\beta}$ contain all the durations of requirement and
contingent events that have a local preference of at least $\beta$.

Once STPU $Q^{\alpha_{min}}$ is obtained, the algorithm checks if it is
strongly controllable.  If the STP obtained applying algorithm \scstpu
\cite{vid:stnu} to STPU $Q^{\alpha_{min}}$ is not consistent, then,
according to Theorem~\ref{lowerp}, there is no hope for any higher
preference, and the algorithm can stop (line 4), reporting that the STPPU is
not $\alpha$-SC $\forall \alpha \geq 0$ and thus is not OSC as well.  If,
instead, no inconsistency is found, \bestsc stores the resulting STP (lines
5-6) and proceeds moving to the next preference level $\alpha_{min}+1$
\footnote{By writing $\alpha_{min} +1$ we mean the next preference level
  higher than $\alpha_{min}$ defined in terms of the granularity of the
  preferences in the [0,1] interval.} (line 7).

In the remaining part of the algorithm (lines 9-21), three steps are
performed at each preference level considered: 
\begin{itemize}
\item Cut STPPU $P$ and obtain STPU $Q^{\beta}$ (line 10);
\item Apply path consistency to $Q^{\beta}$ considering it as an STP:
  \pca($Q^{\beta}$) (line 11);
\item Apply strong controllability to STPU \pca($Q^{\beta}$) (line 13).
\end{itemize}

Let us now consider the last two steps in detail.

Applying path consistency to STPU $Q^{\beta}$ means considering it as an
STP, that is, treating contingent constraints as requirement constraints.
We denote as algorithm \pc any algorithm enforcing path-consistency on
the temporal network (see Section \ref{tcsp} and \citeauthor{meiri}, 
1991).  
It returns the minimal network leaving in the intervals only
values that are contained in at least one solution. This allows us to
identify all the situations, $\omega$, that correspond to contingent
durations that locally have preference $\geq \beta$ and that are consistent
with at least one control sequence of elements in $Q^{\beta}$.  In other
words, applying path consistency to $Q^{\beta}$ leaves in the contingent
intervals only durations that belong to situations such that the
corresponding projections have optimal value at least $\beta$.  If such a
test gives an inconsistency, it means that the given STPU, seen as an STP,
has no solution, and hence that all the projections corresponding to
scenarios of STPPU $P$ have optimal preference $< \beta$ (line 11).

The third and last step applies \scstpu to path-consistent STPU
\pca($Q^{\beta}$), reintroducing the information on uncertainty on the
contingent constraints.  Recall that the algorithm rewrites all the
contingent constraints in terms of constraints on only executable
time-points. If the STPU is strongly controllable, \scstpu will leave in
the requirement intervals only elements that identify control sequences
that are consistent with any possible situation. In our case, applying
\scstpu to \pca($Q^{\beta}$) will find, if any, all the control sequences
of \pca($Q^{\beta}$) that are consistent with any possible situation in
\pca($Q^{\beta}$).

However, if STPU \pca($Q^{\beta}$) is strongly controllable, some of the
control sequences found might not be optimal for scenarios with optimal
preference lower than $\beta$.  In order to keep only those control
sequences that guarantee optimal strong controllability for all preference
levels up to $\beta$, the STP obtained by \scstpua(\pca($Q^{\beta}$)) is
intersected with the corresponding STP found in the previous step (at
preference level $\beta-1$), that is $P^{\beta-1}$ (line 15).  We recall
that given two two STPs, $P_1$ and $P_2$, defined on the same set of
variables, the STP $P_3=P_1 \otimes P_2$ has the same variables as $P_1$
and $P_2$ and each temporal constraint, $c_{ij}^3=c_{ij}^1 \otimes
c_{ij}^2$, is the intersection of the corresponding intervals of $P_1$ and
$P_2$.  If the intersection becomes empty on some constraint or the STP
obtained is inconsistent, we can conclude that there is no control sequence
that will guarantee strong controllability and optimality for preference
level $\beta$ and, at the same time, for all preferences $< \beta$ (line
16).  If, instead, the STP obtained is consistent, algorithm \bestsc
considers the next preference level, $\beta +1$, and performs the three
steps again.

The output of the algorithm is the STP, $P^{\beta-1}$, 
obtained in the iteration previous to the one causing the execution to
stop (lines 23-24) and two of its solutions, $s_e$ and $s_l$.  
This STP, as we will show shortly, contains all the control
sequences that guarantee $\alpha$-SC up to $\alpha=\beta-1$. Only if
$\beta-1$ is the highest preference level at which cutting gives a
consistent problem, then the STPPU is OSC.
The solutions provided by the algorithm are respectively the 
the earliest, $s_e$, and the latest, $s_l$, solutions of $P^{\beta-1}$. 
In fact, as proved in \cite{meiri}
and mentioned in Section \ref{tcsp}, since $P^{\beta-1}$ is minimal,
the earliest (resp. latest) solution 
corresponds to assigning to each variable the lower (resp. upper) bound
of the interval on the constraint defined on $X_0$ and the variable.
This is indicated in the algorithm 
by procedures \solverns{Earliest-Solution} and
\solverns{Latest-Solution}. Let us also recall that every other
solution can be found from $P^{\beta-1}$ without backtracking.

Before formally proving the correctness of algorithm \bestsc, we give
an example.

\begin{table}[tb]
{\small\caption{
In this table each row corresponds to 
a preference level $\beta$ and represents
the intervals of STPU $Q^{\beta}$ obtained by cutting 
the STPPU in Figure~\ref{fig:eos} at level $\beta$.} 
\begin{center}
\begin{tabular}{cc ccc}
\hline\noalign{\smallskip}
STPU && $(\event{SC} \rightarrow \event{EC})$  
& $(\event{SC} \rightarrow \event{SA})$ &  
$(\event{SA} \rightarrow \event{EC})$ 
\\
\noalign{\smallskip}\hline\noalign{\smallskip}
$Q^{0.5}$ && $[1,8]$ & $[1,5]$ & $[-6,4] $ \\
$Q^{0.6}$ && $[1,7]$ & $[1,5]$ & $[-6,4] $ \\
$Q^{0.7}$ && $[1,6]$ & $[1,5]$ & $[-5,2] $ \\
$Q^{0.8}$ && $[1,5]$ & $[1,5]$ & $[-4,1] $ \\
$Q^{0.9}$ && $[1,4]$ & $[1,5]$ & $[-3,0] $ \\
$Q^1$   && $[1,2]$ & $[1,3]$ & $[-2,-1]$ \\
\hline
\end{tabular}
\end{center}\vspace{-0.3cm}
\label{table-int1}}
\vspace{-0.3cm}
\end{table}

\begin{table}[tb]
{\small\caption{
In this table each row corresponds to 
a preference level $\beta$ and represents
the intervals of STPU \pca($Q^{\beta}$) obtained applying 
path consistency to the STPUs in Table~\ref{table-int1}.} 
\begin{center}
\begin{tabular}{cc ccc}
\hline\noalign{\smallskip}
STPU && $(\event{SC} \rightarrow \event{EC})$  
& $(\event{SC} \rightarrow \event{SA})$ &  
$(\event{SA} \rightarrow \event{EC})$ 
\\
\noalign{\smallskip}\hline\noalign{\smallskip}
$\pca(Q^{0.5})$ && $[1,8]$ & $[1,5]$ & $[-4,4] $ \\
$\pca(Q^{0.6})$ && $[1,7]$ & $[1,5]$ & $[-4,4] $ \\
$\pca(Q^{0.7})$ && $[1,6]$ & $[1,5]$ & $[-4,2] $ \\
$\pca(Q^{0.8})$ && $[1,5]$ & $[1,5]$ & $[-4,1] $ \\
$\pca(Q^{0.9})$ && $[1,4]$ & $[1,5]$ & $[-3,0] $ \\
$\pca(Q^1)  $   && $[1,2]$ & $[2,3]$ & $[-2,-1]$ \\
\hline
\end{tabular}
\end{center}\vspace{-0.3cm}
\label{table-int2}}
\vspace{-0.3cm}
\end{table}

\begin{table}[tb]
{\small\caption{
In this table each row corresponds to 
a preference level $\beta$ and represents
the intervals of STP \scstpu\pca($Q^{\beta}$) obtained applying 
the strong controllability check to the STPUs 
in Table~\ref{table-int2}.} 
\begin{center}
\begin{tabular}{cc}
\hline\noalign{\smallskip}
STP & $(\event{SC} \rightarrow \event{SA})$ \\
\noalign{\smallskip}\hline\noalign{\smallskip}
\scstpua(\pca($Q^{0.5}$)) & $[4,5]$\\
\scstpua(\pca($Q^{0.6}$)) & $[3,5]$\\
\scstpua(\pca($Q^{0.7}$)) & $[4,5]$\\
\scstpua(\pca($Q^{0.8}$)) & $[4,5]$\\
\scstpua(\pca($Q^{0.9}$)) & $[4,4]$ \\
\scstpua(\pca($Q^1$))     & $[3,3]$ \\
\hline
\end{tabular}
\end{center}\vspace{-0.3cm}
\label{table-int3}}
\vspace{-0.3cm}
\end{table}

\begin{example}\label{eosexampleOSC}
  Consider the STPPU described in Example~\ref{eosexample}, and depicted in
  Figure~\ref{fig:eos}.  For simplicity we focus on the triangular
  sub-problem on variables $\event{SC}$, $\event{SA}$, and $\event{EC}$.
  In their example, $\alpha_{min}=0.5$.  Table~\ref{table-int1} 
  shows the STPUs $Q^{\beta}$ obtained cutting the problem at each
  preference level $\beta=0.5, \dots, 1$.  Table~\ref{table-int2} shows the
  result of applying path consistency (line 11) to each of the STPUs shown
  in Table~\ref{table-int1}.  As can be seen, all of the
  STPUs are consistent. Finally, Table~\ref{table-int3} shows the STPs
  defined only on executable variables, $\event{SC}$ and $\event{SA}$, that
  are obtained applying \scstpu to the STPUs in Table~\ref{table-int2}.
  
  By looking at Tables~\ref{table-int2}~and~\ref{table-int3} it is easy to
  deduce that the \bestsc will stop at preference level 1.  In fact, by
  looking more carefully at Table~\ref{table-int3}, we can see that STP
  $P^{0.9}$ consists of interval $[4,4]$ on the constraint $\event{SC}
  \rightarrow \event{SA}$, while \scstpua(\pca($Q^{1}$)) consist of
  interval $[3,3]$ on the same constraint.  Obviously intersecting the two
  gives an inconsistency, causing the condition in line 16 of
  Figure~\ref{hasc} to be satisfied.
  
  The conclusion of executing \bestsc on the example depicted in
  Figure\ref{fig:eos} is that it is $0.9$-SC but not OSC.  Let us now see
  why this is correct. Without loss of generality we can assume that
  $\event{SC}$ is assigned value 0. From the last line of
  Table~\ref{table-int3} observe that the only value that can be assigned
  to $\event{SA}$ that is optimal with both scenarios that have optimal
  preference 1 (that is when $\event{EC}$ is assigned 1 or 2) is 3.
  However, assigning 3 to $\event{SA}$ is not optimal if $\event{EC}$
  happens at 6, since this scenario has optimal preference value 0.7 (\eg
  if $\event{SA}$ is assigned 5) while in this case it would have a global
  preference 0.6 (given in constraint $\event{SA} \rightarrow
  \event{EC}$)\footnote{Recall that in the fuzzy semiring context the
    global preference of any assignment is computed taking the minimum
    preference assigned to any of its projections.}.$\Box$
\end{example}

\subsection{Properties of \bestsc}

We will now prove that algorithm \bestsc correctly identifies whether an
STPPU $P$ is OSC, and, if not, finds the highest preference level at
which $P$ is $\alpha$-SC. Let us first consider the events in which \bestsc
stops.
\begin{itemize}
\item {\bf Event 1.} \scstpua($Q^{\alpha_{min}}$) is inconsistent
  (line 4); 

\item {\bf Event 2.} \pca($Q^{\gamma}$) returns an inconsistency (line
  11); 

\item {\bf Event 3.} \pca($Q^{\gamma}$) is consistent but
it is not strongly controllable (line 13); 

\item {\bf Event 4.} \pca($Q^{\gamma}$) is strongly controllable, 
however the intersection 
of the STP obtained by \scstpua(\pca($Q^{\gamma})$) 
with the STP obtained at the previous preference level, 
$P^{\gamma-1}$, is inconsistent (line 16).

\end{itemize}

First notice  that the algorithm terminates.

\begin{theorem} 
\label{th1sc}
Given any STPPU P with a finite number of preference levels, 
the execution of algorithm \bestsc over $P$ terminates.
\end{theorem}

Intuitively, either one of the termination events occur or all the
preference levels will be exhausted. 

Next, let us show 
that \bestdc is a sound and complete algorithm for 
checking if an STPPU is OSC and for finding the highest preference
level at which it is $\alpha$-SC. 

As we have said before, cutting an STPPU $P$ at a preference level
$\gamma$ gives an STPU $Q^{\gamma}$. Moreover, every situation
$\omega=\{\omega_1, \dots, \omega_l\}$ of $Q^{\gamma}$ 
can be seen as a situation of $P$   
such that 
$f_j(\omega_j) \geq \gamma, \forall j$.
This implies that every projection  
$P_\omega \in Proj(Q^{\gamma})$, which is an STP, corresponds 
to  a projection  
$P_{\omega} \in Proj(P)$ which is an STPP.
For all situations $\omega$ of $Q^{\gamma}$, 
in what follows we will write always $P_{\omega}$ which should be
interpreted as an STP when seen as a projection of $Q^{\gamma}$ and as
an STPP when seen as a projection of $P$.    
In the following lemmas we state properties which 
relate the solutions of such projections in the two contexts: without
and with preferences.

\begin{theorem} \label{lem1}
Consider an STPPU $P=\langle N_e, N_c, L_r, L_c, S_{FCSP} \rangle$ 
and preference level $\gamma$, and consider the 
STPU $Q^{\gamma}=\langle N_e, N_c, L'_r,L'_c \rangle$ 
obtained by cutting $P$ at $\gamma$, and STPU 
\pca($Q^{\gamma}$)=$\langle N_e, N_c,$ $L''_r,L''_c \rangle$.
Then: 
\begin{enumerate}
\item $\forall \omega$ situation of $P$, 
$P_{\omega} \in Proj$(\pca($Q^{\gamma}$)) iff
$opt_P(P_{\omega} ) \geq \gamma$;
\item for every control sequence $\delta$, $\delta$ is a 
solution of  $T^{\gamma}$= \scstpua (\pca ($Q^{\gamma})$,  
iff $\forall P_{\omega} \in$ Proj(\pc ($Q^{\gamma}$)),
$T_{\delta,\omega} \in Sol(P_{\omega})$ and
$pref(T_{\delta,\omega})\geq \gamma$.
\end{enumerate}
\end{theorem}

The first part of the theorem states that, by
applying path consistency to STPU $Q^\gamma$,
we remove those situations that cannot be extended to complete
solutions in $Q^\gamma$, and thus correspond to projections having optimal
preference strictly less than $\gamma$.  The second part of the lemma
considers the STP $T^{\gamma}$ obtained applying \scstpu after path
consistency.  In particular it is stated that all the solutions of
$T^{\gamma}$ result, for all the projections of \pc ($Q^{\gamma}$), in
solutions with preference at least $\gamma$. Notice that this implies that
they result in optimal solutions for those projections of $P$ having
optimal preference exactly $\gamma$.
They might
not be optimal, however, for some projections with optimal preference
strictly greater than $\gamma$.

From the above theorem, we get the following corollary, which clarifies the 
relation between the STPU obtained cutting an STPPU at preference 
level $\gamma$, and the $\gamma$-SC of the STPPU.

\begin{corollary}
\label{corsc}
Consider an STPPU $P$ and a preference level $\gamma$ and assume that 
$\exists$ $\omega$, situation of $P$, such that 
$opt(P_{\omega}) \geq \gamma$, where $P_{\omega}$ is the 
corresponding projection. Then, if 
STPU \pca($Q^{\gamma}$), 
obtained by cutting $P$ at $\gamma$, and then 
applying path consistency, is not SC the $P$ is not $\gamma$-SC.
\end{corollary}

Now if we consider all preference levels between $\alpha_{min}$ and
$\gamma$, and compute the corresponding STPs, say $T^{\alpha_{min}}, \dots,
T^{\gamma}$, each such STP will identify the assignments to executable
variables guaranteeing strong controllability and optimality at each 
level. By intersecting all these STPs we 
keep only the common solutions and thus
those which guarantee strong controllability and optimality for all the
situations of $P$ with optimal preference smaller than or equal to $\gamma$.

\begin{theorem}
\label{lem2}
Consider an STPPU $P$, and all preference levels from $\alpha_{min}$ to
$\gamma$, and assume that the corresponding STPs, $T^{\alpha_{min}}, \dots,
T^{\gamma}$ obtained by cutting $P$ at preference levels $\alpha_{min},
\dots, \gamma$, and enforcing strong controllability are consistent.  Then,
$\delta \in Sol (P^\gamma)$, where $P^{\gamma} =\bigotimes_{i=\alpha_{min},
  \dots, \gamma} T^{i}$, iff $\mbox{ } \forall P_{\omega} \in Proj(P)$:
$T_{\delta,\omega} \in Sol(P_{\omega})$, if $opt(P_{\omega}) \leq \gamma$,
then $pref(T_{\delta,\omega})=opt(P_{\omega})$, otherwise
$pref(T_{\delta,\omega})\geq \gamma$.
\end{theorem}

We now consider each of the events in which \bestsc can 
stop and for each of them we prove which of the strong 
controllability properties hold. 

\begin{theorem}
If the execution of algorithm 
\bestsc on STPPU $P$ stops due to the occurrence of Event 1 (line 4),
then $P$ is not $\alpha$-SC $\forall \alpha \geq 0$.
\end{theorem}

This is the case when the underlying STPU 
obtained from the STPPU by ignoring the preference functions is not
strongly controllable. Since cutting at higher preferences will
give even smaller intervals there is no hope for 
controllability at any level and the execution can halt.

\begin{theorem}
If the execution of algorithm 
\bestsc on STPPU $P$ stops due to the occurrence of Event 2 (line 11)
at preference level $\gamma$, then 
\begin{itemize}
\item $\gamma-1=opt=max_{T\in Sol(P)} pref(T)$;
\item $P$ is $OSC$ and a control sequence $\delta$ is a solution of STP
  $P^{opt}$ (returned by the algorithm) iff it is optimal in any scenario
  of $P$.
\end{itemize}
\end{theorem}

This event occurs when the algorithm cuts the STPPU at a given preference
level and the STPU obtained, seen as an STP, is inconsistent.  In
particular, this means that no projection of $Proj(P)$ has an optimal
preference equal to or greater than this preference level.  However, if such a
level has been reached, then up to the previous level, assignments
guaranteeing SC and optimality had been found. Moreover, this previous
level must have been also the highest preference of any solution of $P$,
$opt(P)$. This means that $opt(P)$-SC has been established, which by
Theorem~\ref{oopt} is equivalent to OSC.

\begin{theorem}
If the execution of algorithm 
\bestsc on STPPU $P$ stops due to the occurrence of Event 3 (line 13) 
or Event 4 
(line 16)
at preference level $\gamma$, then $P$ is not OSC but it is  
$(\gamma-1)$-SC and any solution $\delta$ of 
STP $P^{\gamma-1}$ (returned by the algorithm) is 
such that, $\mbox{ } \forall P_{\omega} \in Proj(P)$: 
$T_{\delta,\omega} \in Sol(P_{\omega})$, 
if $opt(P_{\omega}) \leq \gamma-1$, then 
$pref(T_{\delta,\omega})=opt(P_{\omega})$, otherwise 
$pref(T_{\delta,\omega})\geq \gamma-1$.
\end{theorem}

Intuitively, if the algorithm reaches $\gamma$ and stops in line 13, then
there are projections of $P$ with optimal preference $\geq \gamma$ but the
corresponding set of situations is not SC.  
Notice that this is exactly the situation considered in 
Corollary \ref{corsc}.
If instead it stops in line 16,
then this set of situations is SC, but none of the assignments guaranteeing
SC for these situations does the same and is optimal %
for situations at all preference levels up to $\gamma$.  In both cases the
problem is not $\gamma$-SC. However, assuming that $\gamma$ is the first
level at which the execution is stopped the problem is $\gamma-1$-SC.

We conclude this section considering the complexity of \bestsc.

\begin{theorem}
  Determining the OSC or the highest preference level of $\alpha$-SC of an
  STPPU with \defn{n} variables and  \defn{$\ell$} preference levels 
can be achieved in time $O(n^3 \ell)$.
\label{OSCcomp}
\end{theorem}

Notice that we cannot use a binary search over preference levels (in
contrast to algorithms for STPPs), since the correctness of the procedure
is based on the intersection of the result obtained at a given preference
level, $\gamma$, with those obtained at $all$ preference levels $< \gamma$.

The above theorem allows us to conclude that the cost of adding
preferences, and thus a considerable expressive power, is low.  In
fact, the complexity is still
polynomial and it has grown only of a factor equal to the number of
preference levels.



\section{Determining Optimal Weak Controllability}
\label{owc}

Optimal Weak Controllability is the least useful property in practice and
also the property in which adding preferences has the smallest impact in
terms of expressiveness.  What OWC requires is the existence of an optimal
solution in every possible scenario. This is equivalent to requiring the
existence of a solution for every situation, as stated in the
following theorem.


\begin{theorem}
\label{thowc1}
STPPU $P$ is OWC iff the STPU $Q$, obtained by simply ignoring the
preference functions on all the constraints WC.
\end{theorem}

By ignoring the preference functions we mean mapping each soft
constraint $\langle I,f \rangle$ into a hard constraint 
$\langle I \rangle$ defined on the same variables. 
This theorem allows us to conclude that, to check OWC, it is enough to
apply algorithm \weakcon as proposed in \cite{vid:ecai96} and described in
Section~\ref{stppu-back}.  If, instead, we are given a scenario $\omega$,
then we can find an optimal solution of its projection, STPP
$Proj({\omega})$, by using one of the solvers described in
\cite{ros:stpp_learning}.

Let us now consider Example~\ref{eosexample} again. 
Section~\ref{osc} showed that the STPU obtained by cutting the STPPU of
Figure~\ref{fig:eos} at preference level $\alpha_{min}$ 
is strongly controllable. Since SC implies WC, we can conclude that the
STPU is weakly controllable and, thus, that the STPPU in
Figure~\ref{fig:eos} is Optimally Weakly Controllable.


\section{Determining Optimal Dynamic Controllability and $\alpha$-Dynamic
Controllability}
\label{odc}

Optimal Dynamic Controllability (ODC) is the most interesting and useful
property in practice. As described in Section~\ref{sec:motiv}, many industrial
applications can only be solved in a dynamic fashion, making decisions in
response to occurrences of events during the execution of the plan.  This
is true in the space application domains, where planning for a mission is
handled by decomposing the problem into a set of scheduling subproblems,
most of which depend on the occurrence of semi-predictable, contingent
events \cite{fra:fleets}.

In this section we describe an algorithm that tests whether an STPPU $P$ is
ODC and, if not ODC, it finds the highest $\alpha$ at which $P$ is
$\alpha$-DC.  The algorithm presented bears some similarities with \bestsc,
in the sense it decomposes the problem into STPUs corresponding to
different preference levels and performs a bottom up search for dynamically
controllable problems in this space.

Notice that the algorithm is attractive also in practice, since its output
is the minimal form of the problem where only assignments belonging to at
least one optimal solution are left in the domains of the executable
time-points.  This minimal form is to be given as input to an execution
algorithm, which we also describe, that assigns feasible values to
executable time-points dynamically while observing the current situation
(\ie, the values of the contingent time-points that have occurred).

\subsection{A Necessary and Sufficient Condition for Testing ODC}

We now define a necessary and sufficient condition for ODC, which is
defined on the intervals of the STPPU.  We then propose an algorithm which
tests such a condition, and we show that it is a sound and complete
algorithm for testing ODC.

The first claim is that, given an STPPU, the dynamic controllability of the
STPUs obtained by cutting the STPPU and applying \pc at every preference
level is a necessary but not sufficient condition for the optimal dynamic
controllability of the given STPPU.

\begin{theorem}
Given an STPPU $P$, consider any preference level $\alpha$ such that
STPU $Q^{\alpha}$, obtained cutting $P$ at $\alpha$, is
consistent. 
If STPU \pca($Q^{\alpha}$) is not DC 
then $P$ is not ODC and it is not $\beta$-DC, $\forall \beta \geq \alpha$.
\label{thnecc1}
\end{theorem}

Unfortunately this condition is not sufficient, since an STPPU can 
still be not ODC even if at every preference level the STPU obtained
after \pc is DC. An example was shown in Figure~\ref{fail5} 
and is described below.

\begin{example}
  Another potential application of STPPUs is scheduling for aircraft
  analysis of airborne particles \cite{coggiola00}. As an example consider
  an aircraft which is equipped with instruments as the Small Ice Detector
  and a Nevzorov probe, both of which are used to discriminate between liquid
  and ice in given types of clouds.  Such analysis is important
  for the prediction of the evolution of precipitatory systems and of the
  occurrence and severity of aircraft icing \cite{Met03}. Both instruments
  need an uncertain amount of time to determine which is the predominant
  state, between liquid and ice, when activated inside a cloud. 
  
  In the example shown in Figure~\ref{fail5} we consider the sensing event
  represented by variables $\event{A}$ and $\event{C}$ and the start time
  of a maneuver of the aircraft represented by variable $\event{B}$.  Due
  to how the instruments function, an aircraft maneuver can impact the
  analysis. In the example constraint AC represents the duration of the
  sensing event and the preference function models the fact that the
  earlier the predominant state is determined the better.  Constraint AB
  models instead the fact that the maneuver should start as soon as
  possible, for example, due to time constraints imposed by the aircraft's
  fuel availability.  Constraint BC models the fact that the maneuver
  should ideally start just before or just after the sensing event has
  ended.

  
  Let us call $P$ the STPPU depicted in Figure~\ref{fail5}.  In order to
  determine the highest preference level of any schedule of $P$ we can, for
  example use algorithm \stppchop \cite{ros:stpp_learning}.  The highest
  preference level at which cutting the functions gives a consistent STP
  is 1 (interval $[3,3]$ on AB, $[3,5]$ on AC and interval $[0,2]$
  on BC is a consistent STP).  The optimal solutions of $P$,
  regarded as an STPP, will have global preference $1$.
  
  Consider all the STPUs obtained by cutting at every preference level
  from the highest, 1, to the lowest 0.5.
  The minimum preference on any constraint in $P$ is $\alpha_{min}=0.5$
  and, it is easy to see, that all the STPUs obtained by cutting $P$ and
  applying \pc at all preference levels from 0.5 to 1 are DC.  However, $P$
  is not ODC. In fact, the only dynamic assignment to B that belongs to an
  optimal solution of projections corresponding to elements 3, 4 and 5 in
  $[x,y]$ is 3.  But executing B at 3 will cause an inconsistency if C
  happens at 10, since $10-3=7$ doesn't belong to $[u,v]$.$\Box$
\end{example}

We now elaborate on this example to find a sufficient condition for ODC.
Consider the intervals on AB, $[p^{\alpha},q^{\alpha}]$, and the waits $<C,
t^{\alpha}>$ obtained applying the DC checking algorithm at preference
level $\alpha$.  These are shown in Table~\ref{tab:odcsuff}.

\begin{table}[tb]
{\small\caption{In this table each row corresponds to a preference
    level $\alpha$ and represents the corresponding 
interval and wait on the AB constraint of the STPPU shown in 
Figure~\ref{fail5}.}
\begin{center}
\begin{tabular}{ccc}
\hline\noalign{\smallskip}
$\alpha$ & $[p^{\alpha},q^{\alpha}]$ & wait \\
\noalign{\smallskip}\hline\noalign{\smallskip}
$1  $ & $[3,3]$ & \\
$0.9$ & $[3,4]$ & $<C,3>$ \\
$0.8$ & $[3,5]$ & $<C,3>$ \\
$0.7$ & $[3,6]$ & $<C,3>$ \\
$0.6$ & $[3,7]$ & $<C,3>$ \\
$0.5$ & $[3,7]$ & $<C,4>$ \\
\hline
\end{tabular}
\end{center}
\vspace{-0.3cm}\label{tab:odcsuff}}
\vspace{-0.3cm}
\end{table}




If we look at the first and last intervals, resp., at $\alpha=1$ and
$\alpha=0.5$, there is no way to assign a value to B that at the same time
induces a preference $1$ on constraints AB and BC, if C occurs at 3, 4 or
5, and that also satisfies the wait $<C,4>$, ensuring consistency if C
occurs at 10.  This depends on the fact that the intersection of
$[p^1,q^1]$, \ie, $[3]$, and the sub interval of $[p^{0.5},q^{0.5}]$ that
satisfies $<C,4>$, that is, $[4,7]$, is empty.

We claim that the non-emptiness of such an intersection, together with
the DC of the STPUs obtained by cutting the problem at all preference
levels is a necessary \emph{and sufficient} condition for ODC. 
In the following section we will describe an algorithm which tests
such a condition. Then, in Section~\ref{soco}, we will prove that such an
algorithm is sound and complete w.r.t. testing ODC and finding the
highest level of $\alpha$-DC.

\subsection{Algorithm \bestdc}

The algorithm \bestdc echoes Section~\ref{osc}'s algorithm for checking
Optimal Strong Controllability. As done by \bestsc, it considers the STPUs
obtained by cutting the STPPU at various preference levels. For each
preference level, first it tests whether the STPU obtained considering it
as an STP is path consistent.  Then, it checks if the path consistent STPU
obtained is dynamically controllable, using the algorithm proposed in
\cite{mor:stnu}.  Thus, the control sequences that guarantee DC for
scenarios having different optimal preferences are found.  The next step is
to select only those sequences that satisfy the DC requirement and are
optimal at all preference levels.

\begin{figure}[tbp]
\centering\begin{tabular}{|l|}
\hline
{\bfseries Pseudocode for} \textsf{\bfseries{Best-DC}}\\ \hline
1. {\bfseries input} STPPU $P$;\\
2. compute $\alpha_{min}$;\\
3. STPU $Q^{\alpha_{min}} \receives$ $\alpha_{min}$-\solverns{Cut}($P$);\\
4. {\bf if} (\DCA($Q^{\alpha_{min}}$) inconsistent) {\bf write} ``not $\alpha_{min}$-DC'' and {\bf stop};\\
5. {\bf else} \{ \\
6. \hspace{2mm} STP $P^{\alpha_{min}}$ $\receives$ \DCA($Q^{\alpha_{min}}$);\\
7. \hspace{2mm} preference $\beta \receives \alpha_{min}+1$;\\
8. \hspace{2mm} bool ODC $\receives$ false, bool $\alpha$-DC
$\receives$ false;\\
9. \hspace{2mm} {\bf do} \{\\ 
10. \hspace{4mm} STPU $Q^{\beta} \receives$ $\beta$-\solverns{Cut}($P$);\\
11. \hspace{4mm} {\bf if} (\pca($Q^{\beta}$) inconsistent) ODC
$\receives$ true;\\
12. \hspace{4mm} {\bf else} \{ \\
13.  \hspace{6mm}{\bf if} (\DCA(\pca($Q^{\beta})$) inconsistent) 
$\alpha$-DC $\receives$ true; \\
14.  \hspace{6mm}{\bf else} \{\\
15.     \hspace{8mm} STPU $T^\beta$ $\receives$ \DCA(\pca($Q^{\beta}$));\\
16.  \hspace{8mm} {\bf if}(\solverns{Merge}($P^{\beta-1},T^{\beta}$) FAILS)
\{ $\alpha$-DC $\receives$ true \}\\ 
17. \hspace{8mm} {\bf else} \{\\
18. \hspace{10mm} STPU $P^{\beta} \receives $\solverns{Merge}($P^{\beta-1},T^{\beta}$); \\
19.       \hspace{10mm} $\beta$ $\receives$ $\beta+1$;\\   
20.    \hspace{13mm} \};\\
21.   \hspace{10mm} \};\\
22.   \hspace{8mm} \};\\
23. \hspace{4mm}\}{\bf while} (ODC=false and $\alpha$-DC=false);\\
24. \hspace{2mm}{\bf if} (ODC=true) {\bf write} ``P is ODC'';\\
25. \hspace{2mm}{\bf if} ($\alpha$-DC=true) {\bf write} ``P is'' $(\beta-1)$ ''-DC'';\\
26. \hspace{2mm}{\bf return} STPPU $F^{\beta-1} \receives$ \solverns{resulting\_STPPU}($P$,$P^{\beta-1}$);\\
27. \hspace{2mm} \};\\
\hline
\end{tabular}
\caption{\label{hadc} 
Algorithm that tests if an STPPU is ODC and, if not,
finds the highest $\gamma$ such that STPPU $P$ is $\gamma$-DC.}
\end{figure}
   
The pseudocode is given in Figure~\ref{hadc}.  Algorithm \bestdc takes as
input an STPPU $P$ (line 1) and then computes the minimum preference,
$\alpha_{min}$, assigned on any constraint (line 2).

Once $\alpha_{min}$ is known, the STPU obtained by cutting $P$ at
$\alpha_{min}$ is computed (line 3).  This STPU can be seen as the STPPU $P$
with the same variables and intervals on the constraints as $P$ but with no
preferences.  Such an STPU, which is denoted as $Q^{\alpha_{min}}$, is
given as input to algorithm \DC.  If $Q^{\alpha_{min}}$ is not dynamically
controllable, then $P$ is not ODC nor $\gamma$-DC (for any $\gamma \geq
\alpha_{min}$, hence for all $\gamma$), as shown in Theorem~\ref{thnecc1}.
The algorithm detects the inconsistency and halts (line 4).  If, instead,
$Q^{\alpha_{min}}$ is dynamically controllable, then the STPU that is
returned in output by \DC is saved and denoted with $P^{\alpha_{min}}$
(line 6).  Notice that this STPU is minimal, in the sense that in the
intervals there are only elements belonging to at least one dynamic
schedule \cite{mor:stnu}.  In addition, since we have preferences, the
elements of the requirement intervals, as well as belonging to at least one
dynamic schedule, are part of optimal schedules for scenarios which have a
projection with optimal preference equal to $\alpha_{min}$\footnote{In
  fact, they all have preference at least $\alpha_{min}$ by definition.}.

In line 7 the preference level is updated to the next value in the ordering to be considered 
(according to the given preference granularity). 
In line 8 two Boolean flags, $ODC$ and $\alpha$-$DC$ are defined. 
Setting flag $ODC$ to $true$ will signal that the algorithm has established %
that the problem is ODC, while setting flag $\alpha$-$DC$ to $true$ will 
signal that the algorithm has found the highest preference level at which the 
STPPU is $\alpha$-DC.
 
Lines 9-25 contain the main loop of the algorithm. 
In short, each time the loop is executed, it cuts $P$ at the current
preference level and  
looks if the cutting has produced a path consistent STPU (seen as an STP).
If so, it checks if the path consistent version of the STPU is also dynamically 
controllable and, if also this test is passed, 
then
a new STPU is created 
by `merging' the current results with those of previous levels. 

We now describe each step in detail.
Line 10 cuts $P$ at the current preference level $\beta$.
In line 11 the consistency of the STPU $Q^{\beta}$ is tested applying 
algorithm \pc. If \pc returns an inconsistency, then we can conclude that 
P has no schedule  with preference $\beta$ (or greater). 

The next step is to check if STPU \pca($Q^{\beta}$) is DC.  Notice that
this is required for all preference levels up to the optimal level in order
for $P$ to be ODC, and up to $\gamma$ in order for $P$ to be $\gamma$-DC
(Theorem~\ref{thnecc1}).
If applying algorithm \DC detects that \pca($Q^{\beta}$) is not 
dynamically controllable, then the algorithm sets flag $\alpha$-$DC$ to
true.
If, instead, \pca($Q^{\beta}$) is dynamically controllable the resulting minimal 
STPU is saved and denoted $T^{\beta}$ (line 15). 

In line 16, the output of procedure \solver{Merge} is tested.  This
procedure is used to combine the results up to preference $\beta-1$ with
those at preference $\beta$, by applying it to
the STPU obtained at the end of the previous $while$ iteration,
$P^{\beta-1}$, and STPU $T^{\beta}$.  The pseudocode for \solver{Merge} is
shown in Figure~\ref{merge}, and we will describe it in detail shortly.
If 
no inconsistency is found, the new STPU obtained by the merging procedure
is denoted with $P^{\beta}$ (line 18) and a new preference level is
considered (line 19).

Lines 24-27 take care of the output. Lines 24 and 25 will write in output
if P is ODC or, if not, the highest $\gamma$ at which it is $\gamma$-DC.
In line 27 the final STPPU, $F$, to be given in output, is obtained from
STPU $P^{\beta-1}$, that is, the STPU obtained by the last iteration of the
$while$ cycle which was completed with success (\ie, it had reached line
20).  Function \solver{Resulting\_STPPU} restores all the preferences on
all the intervals of $P^{\beta-1}$ by setting them as they are in $P$.  We
will show that the requirement constraints of $F$ will contain only
elements corresponding to dynamic schedules that are always optimal, if the
result is that $P$ is ODC, or are optimal for scenarios corresponding to
projections with optimal preference $\leq \gamma$ and guarantee a global
preference level of at least $\gamma$ in all others, if the result is that
P is $\gamma$-DC.




\begin{figure}[tbp]
\centering\begin{tabular}{|l|}
\hline
{\bfseries Pseudocode for} \solver{Merge}\\ 
\hline
1. {\bf  input} (STPU $T^{\beta}$, STPU $T^{\beta+1}$);\\

2. STPU $P^{\beta+1} \receives T^{\beta}$; \\

3. {\bfseries for each constraint AB, A and B executables, 
in $P^{\beta+1}$}\\ define interval $[p',q']$ and wait $t'$,\\ 
given \{ interval $[p^{\beta}, q^{\beta}]$, wait
$t^\beta$ in $T^{\beta}$ \}\\
and \{ interval $[p^{\beta+1}, q^{\beta+1}]$, wait $t^{\beta+1}$ in 
$T^{\beta+1}$ \}, as follows:;\\

4. {\bf if} ($t^{\beta}=p^{\beta}$ and $t^{\beta+1}=p^{\beta+1}$) (Precede - Precede)\\   
5. \hspace{2mm} $p'\receives max(p^{\beta}, p^{\beta+1})$,
$q'\receives min(q^{\beta}, q^{\beta+1})$, 
$t'\receives max (t^{\beta},t^{\beta+1})$;\\
6. \hspace{2mm} {\bf if} ($q'<p'$) {\bf return} FAILED; \\
7. {\bf if} ($p^{\beta} < t^{\beta} < q^{\beta}$ and 
$p^{\beta+1} \leq t^{\beta+1} < q^{\beta+1}$) (Unordered - Unordered
or Precede) \\  
8. \hspace{2mm} 
$t'\receives max (t^{\beta},t^{\beta+1})$,
$q'\receives min(q^{\beta}, q^{\beta+1})$;\\ 
9. \hspace{2mm} {\bf if} ($q'<t'$) {\bf return} FAILED; \\ 
10. {\bf output} $P^{\beta+1}$.\\
\hline
\end{tabular}
\caption{\label{merge} 
Algorithm \solver{Merge}.}
\end{figure}


The pseudocode of procedure \solver{Merge} is given in Figure~\ref{merge}.
The input consists of two STPUs defined on the same set of variables.  In
describing how \solver{Merge} works, we will assume it is given in input
two STPUs, $T^{\beta}$ and $T^{\beta+1}$, obtained by cutting two STPPUs at
preference levels $\beta$ and $\beta+1$ and applying, by hypothesis with
success, \pc and \DC (line 1 Figure~\ref{merge}).  

In line 2, \solver{Merge} initializes the STPU which will be given in
output to $T^{\beta}$.  As will be 
formally proven in Theorem~\ref{themerge}, due
to the semi-convexity of the preference functions we have that
$Proj(T^{\beta+1}) \subseteq Proj(T^{\beta})$. Notice that \solver{Merge}
leaves all contingent constraints unaltered. Thus, all the projection with
optimal preference $\beta$ or $\beta+1$ are contained in the set of
projections of $P^{\beta+1}$.

\solver{Merge} considers every requirement constraint defined on any two
executables, say A and B, respectively in $T^{\beta}$ and
$T^{\beta+1}$. 
Since we are assuming that algorithm \DC has been applied to
both STPUs, there can be some waits on the intervals.
Figure~\ref{interv1} illustrates the three cases in which the interval
on AB can be. If the wait expires after the upper bound of the
interval (Figure~\ref{interv1} (a)), then the execution of B must
follow the execution of every contingent time-point ({\em Follow
  case}). If the wait coincides with the lower bound of the interval 
(Figure~\ref{interv1} (b)), then the execution of B must precede that
of any contingent time-point ({\em Precede case}). Finally, as shown
in Figure~\ref{interv1} (c), if the wait is within the interval, 
then B is in the {\em Unordered case} with at least a
contingent time-point, say C.

\solver{Merge} considers in which case the corresponding intervals are 
in $T^{\beta}$ and in $T^{\beta+1}$ (line 3). 
Such intervals are respectively
indicated as $[p^{\beta}, q^{\beta}]$, with wait
$t^\beta$,  and $[p^{\beta+1}, q^{\beta+1}]$, with wait $t^{\beta+1}$.
\solver{Merge} obtains a new interval $[p',q']$ and new wait $t'$,\
which will replace the old wait in $T^{\beta+1}$.
Interval $[p'q']$ will contain all and only the values which are
projections on the AB constraint of some optimal solution of some STPP
corresponding to a situation in $T^{\beta}$ or $T^{\beta+1}$. 
Wait $t'$ is the wait that should be respected during a dynamic
execution in order to guarantee that the solution obtained is optimal,
if the projection corresponding to the final scenario has preference
$\beta$ or $\beta+1$.

Due to the semi-convexity of the preference functions
it cannot be the case that: 
\begin{itemize}
\item AB is a Follow or a Precede case in $T^{\beta}$ and an 
Unordered case in $T^{\beta+1}$;
\item AB is a Follow case in $T^{\beta}$ and a 
Precede case in $T^{\beta+1}$; 
\item AB is a Precede case in $T^{\beta}$ and a 
Follow case in $T^{\beta+1}$;
\end{itemize}

This means that the cases which should be considered are:
\begin{itemize}
\item AB is a Follow case in both $T^{\beta}$ and  $T^{\beta+1}$;
\item AB is a Precede case in $T^{\beta}$ and in $T^{\beta+1}$;
\item AB is a Unordered case in $T^{\beta}$ and a 
Precede or an Unordered case in $T^{\beta+1}$;
\end{itemize}

In the first two cases the AB interval is left as it is in
$T^{\beta+1}$. A formal motivation of this is contained in the proof
of Theorem~\ref{themerge}. However, informally, we can say that the AB
interval in $T^{\beta+1}$ already satisfies the desired property.
   
In lines 4 and 5 the case in which AB is in a {\bf Precede case in both}
STPUs is examined. 
Here, B will always occur before any contingent time-point. The values in
the $[p^{\beta}, q^{\beta}]$ (resp. $[p^{\beta+1}, q^{\beta+1}]$) are
assignments for B that will be consistent with any future occurrence of C
mapped into a preference $\geq \beta$ (resp. $\geq \beta+1$). Clearly the
intersection should be taken in order not to lower the preference if C
occurs with preference $\geq \beta+1$.  Line 6 considers the event in which
such intersection is empty. This means that there is no common assignment
to B, given that of A, that will be optimal both in scenarios with optimal
preference $\beta$ and in scenarios with optimal preference $\beta+1$.

\begin{figure}[tbp] 
\begin{center} \ \setlength{\epsfxsize}{4.3in} 
\epsfbox{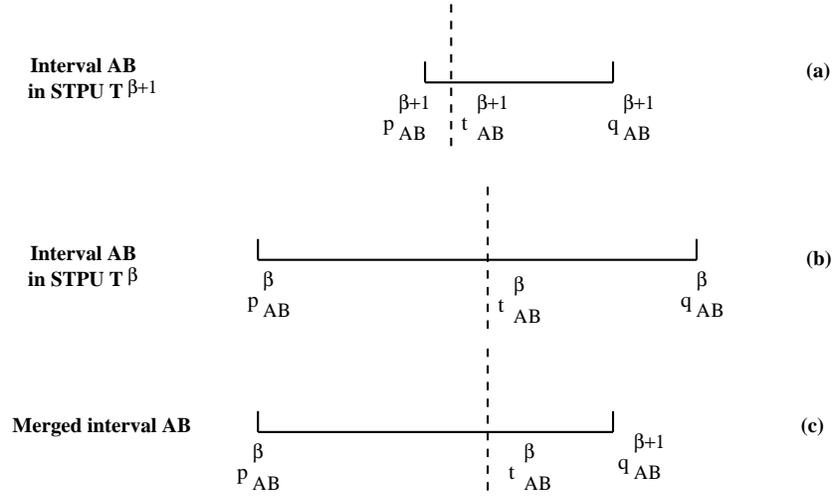} 
\end{center} 
\caption{Merging two intervals in the \emph{Unordered} case.}
\label{intervmerge} 
\vspace{-0.3cm} 
\end{figure}

In lines 7 and 8 two scenarios are considered:
when AB is in the {\bf Unordered}
case in $T^{\beta}$ and in the {\bf Precede} case in $T^{\beta+1}$ and  
when AB is in the {\bf Unordered case in both} STPUs.
Figure~\ref{intervmerge} shows the second case. 
\solver{Merge} takes the union of the parts of the intervals 
preceding the wait and the intersection of the parts following the wait.
The intuition underlying this is that any execution of B identifying an 
element of either $[p^{\beta},t^{\beta}[$ or 
$[p^{\beta+1},t^{\beta+1}[$ will be 
preceded by the execution of all the contingent time-points for which
it has to wait. This means that 
when B is executed, for any such contingent time-point C, 
both the time at which C has been executed, say $t_C$, 
and the associated preference, say $f_{AC}(t_C)$,
on constraint AC in STPPU $P$ will be known. The propagation of this
information will allow us to identify those elements of
$[p^{\beta},t^{\beta}[$ (resp.\ $[p^{\beta+1},t^{\beta+1}[$) that have a
preference $\geq f_{AC}(t_C)$ and thus an optimal assignment for B.  This
means that all the elements in both interval $[p^{\beta},t^{\beta} [$ and
interval $[p^{\beta+1},t^{\beta+1} [$ are eligible to be chosen.  For
example, if $f_{AC}(t_C)=\beta$ there might be values for B with preference
equal to $\beta$ that are optimal in this case but would not if C occurred
at a time such that $f_{AC}(t_C)>\beta$. But since in any case we know when
and with what preference C has occurred, it will be the propagation step
that will prune non-optimal choices for B. In short, leaving all elements
allows more flexibility in the propagation step.  Moreover, as will
be proven in
Theorem~\ref{themerge}, $p^{\beta} \leq p^{\beta+1}$.

If instead we consider elements of interval $[t^{\beta},q^{\beta} ]$, we
know that they identify assignments for B that can be executed regardless
of when C will happen (however we know it will happen with a preference
greater $\geq \beta$). This means that we must take the intersection of
this part with the corresponding one, $[t^{\beta+1},q^{\beta+1}]$, in order
to guarantee consistency and optimality also when C occurs at any time with
preference $= \beta+1$. An easy way to see this is that interval
$[t^{\beta},q^{\beta} ]$ may contain elements that in $P$ are mapped into
preference $\beta$. These elements can be optimal in scenarios in
which C happens at a time associated with a preference $=\beta$ in the AC
constraint; however, they cannot be optimal in scenarios with C occurring at
a time with preference $\beta+1$.

Line 9 handles the case in which the two parts of the intervals, following
the waits, have an empty intersection. In this case, optimality cannot be
guaranteed neither at level $\beta$ nor $\beta+1$, in particular if the
contingent events occur after the waits expire.

\subsection{Properties of \bestdc}
\label{soco}

We will now show that \bestdc is a sound and complete algorithm for testing
ODC and for finding the highest preference level at which the STPPU given
in input is $\alpha$-DC.  We recall, once more, that all the results that
follow rely on the tractability assumptions requiring semi-convex
preference functions and the fuzzy semiring $\langle [0,1],max ,min, 0, 1
\rangle$ as underlying structure.

Let us consider STPPU $P$ and STPUs $T^{\beta}$ and $T^{\beta+1}$, as
defined in the previous section.  Then, STPU
$P^{\beta+1}=$\solver{Merge}($T^{\beta}$, $T^{\beta+1}$) will have the
same contingent constraints as $T^{\beta}$ \footnote{We recall that the
  projections of $T^{\beta}$ coincide with the projections of STPPU $P$
  with optimal preference $\geq \beta$ (see Theorem~\ref{lem1}), and that, due
  to the semi-convexity of the preference functions, $Proj(T^{\beta+1})
  \subseteq Proj(T^{\beta})$.} and requirement constraints as defined by
the merging procedure.  We start by proving that \solver{Merge} is a sound
and complete algorithm for testing the existence of a viable dynamic
strategy, common to both such STPUs, which is optimal for projections
having optimal preference equal to either $\beta$ or $\beta+1$.

\begin{theorem}
  Consider STPPU $P$ and STPUs, $T^{\beta}$ and $T^{\beta+1}$, obtained by
  cutting $P$ respectively at level $\beta$ and $\beta +1$ and applying
  \pc , without finding inconsistencies, and \DC with success.  Consider
  STPU $P^{\beta+1}=$ \solverns{Merge}($T^{\beta}$ , $T^{\beta+1}$).

Then, 
\solverns{Merge}($T^{\beta}$ , $T^{\beta+1}$) does not fail
if and only if 
\begin{itemize}
\item $P^{\beta+1}$ is dynamically controllable and 
\item there is a viable dynamic strategy $S$ 
such that for every projection 
$P_i \in Proj(P^{\beta+1})$, 
\begin{itemize} 
\item if $opt(P_i)=\beta$ or $opt(P_i)=\beta+1$ in
$P$, $pref (S(P_i))=opt(P_i)$;
\item otherwise $pref(S(P_i)) \geq \beta+1$.
\end{itemize}
\end{itemize}
\label{themerge}
\end{theorem}

The following theorem extends the result for the merging procedure to more
than two preference levels, in particular to all preference levels smaller
or equal to a given threshold $\beta$.

\begin{theorem}
\label{cor2}
Consider STPPU $P$ and for every preference level, $\alpha$, define
$T^{\alpha}$ as the STPU obtained by cutting $P$ at $\alpha$, then
applying \pc and then \DC. Assume that $\forall \alpha \leq \beta$,
$T^{\alpha}$ is DC.  Consider STPU $P^{\beta}$:
$$
P^{\beta}=\solverns{Merge}( \solverns{Merge}(\dots \solverns{Merge}(
\solverns{Merge}(T^{\alpha_{min}},T^{\alpha_{min}+1}), T^{\alpha_{min}+2}),
\dots), T^{\beta})
$$
with $\alpha_{min}$ the minimum preference on any constraint in P.
Assume that, when applied, \solverns{Merge} always returned a consistent STPU.
Then, there is a viable dynamic strategy $S$, 
such that $\forall P_i \in Proj(P)$, if $opt(P_i) \leq \beta$ 
then $S(P_i)$ is an optimal solution of $P_i$, otherwise 
$pref(S(P_i)) \geq \beta+1$.
\end{theorem}


Theorem~\ref{cor2} allows us to prove the main result.
Informally, \bestdc applies
\solver{Merge} from the lowest preference to the highest threshold
$\gamma$, above which the returned problem becomes inconsistent.  If there
is no projection of the STPPU with an optimal solution higher than $\gamma$,
then, by using Theorem~\ref{cor2}, we can conclude that the STPPU is ODC;
otherwise it is $\gamma$-DC.

Let us start by enumerating the conditions at which \bestdc terminates:
\begin{itemize}
\item {\bf Event 1.} \bestdc stops because the STPU obtained at level $\alpha_{min}$
  is not DC (line 4);

\item {\bf Event 2.} \bestdc exits because it has reached a preference level
  $\beta$ at which the STPU (seen as an STP) is not path consistent (line 11);

\item {\bf Event 3.} \bestdc stops because it has reached a preference level
  $\beta$ at which the path consistent STPU is not dynamically controllable (line 13);

\item {\bf Event 4.} \bestdc stops because procedure \solver{Merge} has found
 an inconsistency (line 16). 
\end{itemize}

The following theorem shows that the execution of \bestdc always terminates.

\begin{theorem}
\label{stop}
Given an STPPU P, the execution of algorithm \bestdc on $P$ 
terminates.
\end{theorem} 

\bestdc considers each preference level, starting from the lowest and
moving up each time of one level according to the granularity of the
preference set.  It stops either when an inconsistency is found or when all
levels, which are assumed to be finite, have been precessed.

We are now ready to prove the soundness and completeness of \bestdc .
We split the proof into three theorems, 
each considering a different terminating condition.
The first theorem considers the case in which the underlying STPU 
obtained from $P$, by ignoring the preferences, is not DC. 
In such a case the output is that the STPPU is not $\alpha$-DC at any
level and thus is not ODC.

\begin{theorem}
\label{stop4}
Given an STPPU $P$ as input, \bestdc terminates in line 4 iff 
$\not \exists \alpha \geq 0$  such that 
P is $\alpha$-DC.
\end{theorem}

The next theorem considers the case in which the highest preference level
reached with success by the merging procedure is also the highest optimal
preference of any projection of $P$. In such a case, the problem is ODC.

\begin{theorem}
\label{stop11}
Given an STPPU $P$ as input, \bestdc terminates in line 11 iff 
P is ODC.
\end{theorem}

The last result considers the case in which there is at least a projection 
with an optimal preference strictly higher than the highest reached
with success by the merging procedure.  
In such case the problem is not ODC and  \bestdc has found 
the highest level at which the STPPU $\alpha$-DC. 

\begin{theorem}
\label{stop13-17}
Given STPPU $P$ in input, \bestdc stops at lines 13 or 16 
at preference level $\beta$ iff $P$ is $(\beta-1)$-DC and not ODC.
\end{theorem}



As mentioned in Section \ref{stpu-back}, in \cite{dc-rev05}, 
it is proven that checking DC of
an STPU can be done in $O(n^5)$, where $n$ is the number of variables.
The revised algorithm processes the distance
graph of the STPU, rather than its constraint graph. It also maintains
additional information, in the form of additional labeled edges which
correspond to waits.  The main feature of the new algorithm, as noted
earlier, it is a strongly polynomial algorithm for determining the dynamic
controllability of an STPU.  What is important in our context is to stress
the fact that the output of the two algorithms, presented in 
\cite{mor:stnu} and \cite{dc-rev05}, is essentially the same. In
fact it is easy to obtain, in polynomial time $O(n^2)$,
the constraint graph with waits produced by \DC starting from the distance
graph produced by the new algorithm, and vice versa.

\begin{theorem}
  The complexity of determining ODC or the highest preference level
  $\alpha$ of $\alpha$-DC of an STPPU with \defn{n} variables, a bounded
  number of preference levels \defn{$\ell$} 
is time $O(n^5 \ell)$.
\label{ODCcomp}
\end{theorem}

The complexity result given in Theorem~\ref{ODCcomp} is unexpectedly
good. In fact, it shows that the cost of adding a considerable
expressive power through preferences to STPUs is a factor equal to the
number of different preference levels. This implies that solving the
optimization problem and, at the same time,
the controllability problem,
remains in P, if the number of different preference levels is bounded.

\subsection{The Execution Algorithm}

The execution algorithm we propose is very similar to that for STPUs
presented in \cite{mor:stnu}, which we described in Section~\ref{stppu-back}
and shown in Figure~\ref{fig:dc:execution}.  Of course the execution
algorithm for STPPUs will take in input an STPPU to which \bestdc has been
successfully applied.  In line 2 of Figure~\ref{fig:dc:execution}, the
algorithm performs the initial propagation from the starting point.  The
main difference between our STPPU execution algorithm and the STPU
algorithm in \cite{mor:stnu} is that the definition of `propagation' also
involves preferences.

\begin{definition}[soft temporal propagation]
\label{softtempprop}
Consider an STPPU $P$ and a variable $Y \in P$ and a value $v_Y \in D(Y)$.
Then \defn{propagating} the assignment $Y=v_Y$ in $P$, means:
\begin{itemize}
\item for all constraints, $c_{XY}$ involving $Y$ such that $X$ is
  already assigned value $v_X \in  D(X)$: 
  replace the interval on $c_{XY}$ with interval 
  $\langle [v_Y-v_X,v_Y-v_X] \rangle$;
\item cut $P$ at preference level $min_{X} \{f_{c_{XY}}(v_Y-v_X)\}$.$\Box$
\end{itemize}
\end{definition}

We will call \solver{ODC-Execute} the algorithm \solver{DC-Execute} where
propagation is defined as in Definition~\ref{softtempprop}.  Assume we
apply \solver{ODC-Execute} to an ODC or $\alpha$-DC STPPU $P$ 
to which \bestdc has been
applied.  If, up to a given time $T$, the preference of the partial schedule
was $\beta$, then we know that if $P$ was ODC or $\alpha$-DC with $\alpha
\geq \beta$, by Theorem~\ref{themerge} and 
Theorem~\ref{cor2}, the execution
algorithm has been assigning values in $T^{\beta+1}$.  Assume now that a
contingent event occurs and lowers the preference to $\beta-2$. This
will be propagated and the STPPU will be cut at preference level
$\beta-2$.  From now on, the execution algorithm will assign values in
$T^{\beta-2}$ and, by Theorem~\ref{themerge} and Theorem~\ref{cor2}, the new
waits imposed will be such that the assignments for the executables will be
optimal in any situation where the optimal preference is $\leq \beta-2$. In
all other situations such assignments guarantee a preference of at least
$\beta-2$.

 
\section{Using the Algorithms}
\label{comp-con}

Section~\ref{compcont} described the relations between our notions of
controllability.  As a general strategy, given an STPPU, the first property
to consider is OSC.  If it holds, the solution obtained is feasible and
optimal in all possible scenarios. However, OSC is a strong property and
holds infrequently.  If the STPPU is not OSC, but we still need to have a
control sequence before execution begins, \bestsc will find the best
solution that is consistent with all possible future situations.

Most commonly, dynamic controllability will be more useful.  If the control
sequence needs not be known before execution begins, ODC is ideal.  Notice
that, from the results in Section~\ref{compcont}, an STPPU may be not OSC
and still be ODC.  If, however, the STPPU is not even ODC, then \bestdc
will give a dynamic solution with the highest preference.  Recall, as we
have shown in Section~\ref{compcont}, that for any given preference level
$\alpha$, $\alpha$-SC implies $\alpha$-DC but not vice versa.  Thus, it may
be that a given STPPU is $\beta$-SC and $\gamma$-DC with $\gamma > \beta$.
Being $\beta$-SC means that there is a fixed way to assign values to the
executables such that it will be optimal only in situations with optimal
preference $\leq \beta$ and will give a preference at least $\beta$ in all
other cases.  On the other hand, $\gamma$-DC implies that a solution
obtained dynamically, by the \solver{ODC-Execute} algorithm, will be
optimal for all those situations where the best solution has preference
$\leq \gamma$ and will yield a preference $\geq \gamma$ in all other cases.
Thus, if $\gamma > \beta$, using the dynamic strategy
will guarantee optimality in more situations and a higher preference in all
others.

The last possibility is to check OWC.  This will at least allow the
executing agent to know in advance if there is some situation that has no
solution.  Moreover, if the situation is revealed just before the execution
begins, using any of the solvers for STPPs described in
\cite{ros:stpp_learning} will allow us to find an optimal assignment for
that scenario.


\section{Related Work}
\label{stppu-rlw}

In this section we survey work which we regard as closely related to ours.
Temporal uncertainty has been studied before, but it has been defined 
in different ways according to the different contexts where it has been used.

We start considering the work proposed by \citeA{vila-fuzzy}. 
They propose \defn{Fuzzy Temporal Constraint Networks}, which are STPs
where the interval in each constraint is mapped into a possibility
distribution. In fact, they handle temporal uncertainty using possibility
theory \cite{zadeh75}, using the term `uncertainty' to describe vagueness
in the temporal information available. Their aim is to model statements as
``He called me more or less an hour ago'', where the uncertainty is the
lack of precise information on a temporal event. Their goal thus is
completely different from ours. In fact, we are in a scenario where an
agent must execute some activities at certain times, and such activities
are constrained by temporal relations with uncertain events. Our goal is to
find a way to execute what is in the agents control in a way that will be
consistent whatever nature decides in the future.

In \cite{vila-fuzzy}, instead, they assume to have imprecise temporal
information on events happened in the past. Their aim is to check if such
information is consistent, that is, if there are no contradictions implied
and to study what is entailed by the set of constraints. In order to model
such imprecise knowledge, possibilities are again used.  Every element of
an interval is mapped into a value that indicates how possible that event
is or how certain it is. Thus, another major difference with their approach
is that they do not consider preferences, only possibilities. On the other
hand, in the work presented here we do not allow to express information on
how possible or probable a value is for a contingent time-point. This is
one of the lines of research we want to pursue in the future. Moreover, in
\cite{vila-fuzzy}, they are concerned with the classical notion of
consistency (consistency level) rather than with controllability.

Another work related to the way we handle uncertainty is that of
\citeA{badaloni}.  They introduce \defn{Flexible Temporal Constraints} where
soft constraints are used to express preferences among feasible solutions
and prioritized constraints are used to express the degree of necessity of
the constraints' satisfaction.  In particular, they consider qualitative
Allen-style temporal relations and they associate each such relation to a
preference. The uncertainty they deal with is not on the time of occurrence
of an event but is on whether a constraint belongs or not to the constraint
problem.  In their model, information coming from plausibility and
information coming from preferences is mixed and is not distinguishable by
the solver. In other words, it is not possible to say whether a solution is
bad due to its poor preference on some relation or due to it violating a
constraint with a high priority. In our approach, instead, uncertainty and
preferences are separated. The compatibility with an uncertain event does
not change the preference of an assignment to an executable. The robustness
to temporal uncertainty is handled intrinsically by 
the different degrees of controllability.

In \cite{prade-allen} the authors consider fuzziness and uncertainty
in temporal reasoning by introducing \defn{Fuzzy Allen Relations}. 
More precisely, they present an extension of
Allen relational calculus, based on fuzzy comparators expressing
linguistic tolerance. 
\citeA{prade-allen} want to handle 
situations in which the information about
dates and relative positions of intervals is complete but, for some
reason, there is no interest in describing it in a precise manner. For
example, when one wants to speak only in terms of ``approximate
equality'', or proximity rather that in terms of precise equality.
Secondly, they want to be able to deal with available information
pervaded with imprecision, vagueness or uncertainty. 
In the framework we have presented we restrict the 
uncertainty to when an event will occur within a range. On the
other hand, we put ourselves into a ``complete ignorance'' position,
that would be equivalent, in the context of \cite{prade-allen}, 
to setting to 1 all possibilities of all contingent events. 
Moreover, in \cite{prade-allen} they do not allow preferences nor
address controllability. Instead, they consider, similarly to
\cite{vila-fuzzy}, the notions of 
consistency and entailment. The first notion is checked by
computing the transitive closure of the fuzzy temporal relations using
inference rules appropriately defined. The second notion 
is checked by defining several
patterns of inference.

Another work which addresses also temporal uncertainty is presented
in \cite{fuzzyjob-shop} and in \cite{fuzzysched}.  In this work both
preferences and activities with ill-known durations in the classical
job-shop scheduling problem are handled using the fuzzy framework. There
are three types of constraints: precedence constraints, capacity constraints
and due dates, and release time constraints.
In order to model such unpredictable events they use
possibility theory.
As the authors mention in \cite{fuzzyjob-shop}, possibility
distributions can be viewed as modeling uncertainty as well as
preference \cite<see>{dubois93}.  Everything depends on whether the
variable $X$ on which the possibility distribution is defined is
controllable or not. 
Thus \citeA{fuzzyjob-shop} distinguish between controllable and
uncontrollable variables.  However they do not allow to specify
preferences on uncontrollable events. Our preference functions
over contingent constraints would be interpreted 
as possibility distributions in their framework. In some sense, our work is
complementary to theirs. We
 assume a constraint possibility
distribution on contingent events 
always equal to 1 and we allow no representation of any further 
information on more or less possible values; on the other hand, we 
allow to specify 
preferences also on uncontrollable events. They, on the contrary,
allow to put possibility distributions on contingent events, but not
preferences.


Finally, \citeA{fuzzyjob-shop} show that a scheduling
problem with uncertain durations can be formally expressed by the same kind
of constraints as a problem involving what they call flexible durations
(\ie durations with fuzzy preferences).  However the interpretation is
quite different: in the case of flexible durations, the fuzzy information
comes from the specifications of preferences and represents the possible
values that can be assigned to a variable representing a duration. In the
case of imprecisely known durations, the fuzzy information comes from
uncertainty about the real value of some durations.  The formal
correspondence between the two constraints is so close that the authors do
not distinguish among them when describing the solving procedure.  Further,
the problem they solve is to find the starting times of activities such
that these activities take place within the global feasibility window
\emph{whatever the actual values of the unpredictable durations will be}.
Clearly this is equivalent to Optimal Strong Controllability. They do not
address the problem of dynamic or weak controllability with preferences.


\section{Summary and Future Work}
\label{conclusion}

We have defined a formalism to model problems with quantitative temporal
constraints with both preferences and uncertainty, and we have generalized
to this formalism three classical notions of controllability (that is,
strong, weak and dynamic).  We have then focused on a tractable class of
such problems, and we have developed algorithms that check the presence of
these properties.

This work advances the state of the art in temporal reasoning and
uncertainty since it provides a way to handle preferences in this context,
and to select the best solution (rather than a feasible one) in the
presence of uncontrollable events.  Moreover, it shows that the
computational properties of the controllability checking algorithms do not
change by adding preferences. In particular, dynamic controllability can
still be checked in polynomial time for the considered class of problems,
producing dynamically temporal plans under uncertainty that are optimal
\wrt preferences.

Among the future directions we want to pursue within this line of research,
the first is a deeper study of methods and algorithms for adding
preferences different from fuzzy ones.  Notice that the framework that we
have proposed here is able to represent any kind on preference within the
soft constraint framework. However, our algorithms apply only to fuzzy
preferences and semi-convex functions.  In particular, we would like to
consider the impact on the design and complexity of algorithms when there
are uncontrollable events and the underlying preference structures is the
weighted or the probabilistic semiring. Both of these semirings are
characterized by non-idempotent multiplicative operators. This can be a
problem when applying constraint propagation \cite{jacm}, such as
path-consistency, in such constraints. Thus search and propagation
techniques will have to be adapted to an environment featuring uncertainty
as well.  It should be noticed that in \cite{stput} some algorithms for
finding optimal solutions of STPs with preferences in the weighted semiring
have been proposed.  Another interesting class of preferences are
utilitarian ones.  In such a context each preference represents a utility
and the goal is to maximize the sum of the utilities.  Such preferences
have been used in a temporal context without uncertainty for example in
\cite{Se}.

Recently, another approach for handling temporal uncertainty has been
introduced in \cite{tsamar2,tsamar3}: \defn{Probabilistic Simple Temporal
  Problems} (PSTPs); similar ideas are presented in \cite{lau:robust_stp}.
In the PSTP framework, rather than bounding the occurrence of an
uncontrollable event within an interval, as in STPUs, a probability
distribution describing when the event is more likely to occur is defined
on the entire set of reals.
As in STPUs, the way the problem is solved depends on the assumptions made
regarding the knowledge about the uncontrollable variables. In particular
they define the \defn{Static Scheduling Optimization Problem}, which is the
equivalent to finding an execution satisfying SC in STPUs, and the
\defn{Dynamic Scheduling Optimization Problem}, equivalent to finding a
dynamic execution strategy in the context of STPUs.
In the above framework, optimal means ``with the highest probability of
satisfying all the constraints''. Preferences are not considered in this
framework. We believe it would
be interesting to add preferences also to this approach.
A first step could consists of keeping, for each strategy, separately its
global preference and its probability of success.  In this way we could use
the existing frameworks for handling the two aspects. Then, we can order
the strategies by giving priority to preferences, thus taking in some sense
a risky attitude, or, on the contrary, by giving priority to probabilities,
adopting a more cautious attitude.
A step in this direction has been recently proposed in \cite{stppp}, where,
however, the authors, rather than actually extending the notions of
consistency of PSTPs to handle preferences, consider inducing preferences
from probabilities.  In contrast, our approach is preliminary advanced in
\cite{pin:uncert_soft}.

Up to now we have focused our attention on non-disjunctive temporal
problems, that is, with only one interval per constraint.  We would like to
consider adding uncertainty to \defn{Disjunctive Temporal Problems}
\cite{dtps}, and to consider scenarios where there are both preferences and
uncertainty.  Such problems are not polynomial even without preferences or
uncertainty but it has been shown that the cost of adding preferences is
small \cite{dtpp}, so we hope that the same will hold in environments with
uncertainty as well.  Surprisingly, uncertainty in Disjoint Temporal
Problems has not been considered yet, although it is easy to see how
allowing multiple intervals on a constraint is itself a form of
uncontrollability. We, thus, plan to start defining DTPUs (preliminary
results are in \citeauthor{dtpu}, 2005) and then to merge this approach with the
existing one for DTPPs.

Extending \defn{Conditional Temporal Problems}, a framework proposed in
\cite{ctps}, is also a topic of interest for us.  In such model a Boolean
formula is attached to each temporal variable.  These formulae represent
the conditions which must be satisfied in order for the execution of events
to be enabled. In this framework the uncertainty is on which temporal
variables will be executed.  We believe that it would be interesting to
extend this approach in order to allow for conditional preferences:
allowing preference functions on constraints to have different shapes
according to the truth values of some formulas, or the occurrence of some
event at some time.  This would provide an additional gain in
expressiveness, allowing one to express the dynamic aspect of preferences
that change over time.



\section*{Appendix A}

\begin{theorem2}
If an STPPU $P$ is OSC, then it is ODC; if it is ODC, then it is OWC.
\end{theorem2}
\proof{Let us assume that $P$ is OSC.
Then there is
a viable execution strategy $S$ such that, 
$\forall P_1, P_2 \in Proj(P)$ and for every
  executable time-point $x$, $[S(P_1)]_x=[S(P_2)]_x$ and
$S(P_1) \in OptSol(P_1)$ and $S(P_2) \in OptSol(P_2)$.
Thus, in particular, $[S(P_1)]_x=[S(P_2)]_x$ for every pair f
  projections such that $[S(P_1)]_{<x}=[S(P_2)]_{<x}$.
This allows us to conclude that if $P$ is OSC 
then it is also ODC and any strategy which is a
witness of OSC is also a witness of ODC. 

Let us now assume that $P$ is ODC.
Then, in particular, there is a viable
dynamic strategy $S$ such that 
$\forall P_1 \in Proj(P)$, $S(P_1)$ is an optimal 
solution of $P_1$. This clearly means that every
projection has at least an optimal solution. Thus $P$ is OWC. $\Box$}

\begin{theorem2}
For any given preference level $\alpha$, 
if an STPPU $P$ is $\alpha$-SC then it is $\alpha$-DC. 
\end{theorem2}
\proof{Assume that $P$ is $\alpha$-SC. Then there
is a viable strategy $S$ such that:
$[S(P_1)]_x=[S(P_2)]_x$, $\forall P_1, P_2 \in Proj (P)$
    and for every executable time-point $x$, and
    $S(P_\omega)$ is an optimal solution of 
    projection $P_\omega$, if there is no optimal solution of $P_\omega$
    with preference $> \alpha$
    and 
    $pref(S(P_{\omega})) \not < \alpha$, otherwise. 

Thus, $[S(P_1)]_x=[S(P_2)]_x$ also for all pairs of projections, 
$P_1$ and $P_2$ such that $[S(P_1)]_{<x}=[S(P_2)]_{<x}$. This implies
that $P$ is $\alpha$-DC. $\Box$}

\begin{theorem2}
\label{lowerp2}
Given an STPPU $P$ and a preference level $\beta$, 
if $P$ is  $\beta$-SC (resp. $\beta$-DC), then it is 
$\alpha$-SC (resp. $\alpha$-DC), $\forall \alpha < \beta$. 
\end{theorem2}
\proof{If $P$ is $\beta$-SC then there
is a viable strategy $S$ such that:
$[S(P_1)]_x=[S(P_2)]_x$, $\forall P_1, P_2 \in Proj(P)$
    and for every executable time-point $x$, and
    $S(P_{\omega})$ is an optimal solution of 
    $P_{\omega}$ if there is no optimal solution of $P_\omega$
    with preference $> \beta$ and 
    $pref(S(P_{\omega})) \not < \beta$, otherwise. But, of course,
$\forall \alpha < \beta$ the set of projections 
with no optimal solution with preference $> \alpha$
is included in that of projections 
with no optimal solution with preference
$> \beta$. 
Moreover, for all the other projections, $P_z$, 
$pref(S(P_{z})) \not < \beta$ implies that $pref(S(P_{z})) \not < \alpha$
since $\beta > \alpha$.
Similarly for $\beta$-DC.$\Box$}

\begin{theorem2}
\label{oopt2}
Given an STPPU $P$, let $opt=max_{T \in Sol(P)}pref(T)$. Then, 
P is OSC (resp. ODC) iff it is $opt$-SC (resp. $opt$-DC).
\end{theorem2}
\proof{The result comes directly from the fact that $\forall P_i \in
  Proj(P)$, $opt(P_i) \leq opt$, and there is always at least a projection,
$P_j$, such that $opt(P_j)=opt$.$\Box$}


\begin{theorem2} 
Given any STPPU P with a finite number of preference levels, 
the execution of algorithm \bestsc over $P$ terminates.
\end{theorem2}
\proof{
Consider STPPU $P$ and its optimal preference value 
$opt =max_{T\in Sol(P)} pref(T)$, that is, 
the highest preference assigned to any of its solutions. By definition, 
$Q^{opt+1}$ is not consistent. 
This means that if the algorithm reaches 
level $opt+1$ (that is, the next preference level higher than $opt$ in
the granularity of the preferences)  
then the condition in line 11 will be satisfied and the execution will halt.
By looking at lines 9-20 we can see that either one of the events that cause the execution 
to terminate occurs or the preference level is incremented in line 16.
Since there is a finite number of preference levels, 
this allows us to conclude that the algorithm will terminate in a finite 
number of steps. $\Box$}

\begin{theorem2} \label{lem12}
Consider an STPPU $P=\langle N_e, N_c, L_r, L_c, S_{FCSP} \rangle$ 
and preference level $\gamma$, and consider the 
STPU $Q^{\gamma}=\langle N_e, N_c, L'_r,L'_c \rangle$ 
obtained by cutting $P$ at $\gamma$, and STPU 
\pca($Q^{\gamma}$)=$\langle N_e, N_c,$ $L''_r,L''_c \rangle$.
Then: 
\begin{enumerate}
\item $\forall \omega$ situation of $P$, 
$P_{\omega} \in Proj$(\pca($Q^{\gamma}$)) iff
$opt_P(P_{\omega} ) \geq \gamma$;
\item for every control sequence $\delta$, $\delta$ is a 
solution of  $T^{\gamma}$= \scstpua (\pca ($Q^{\gamma})$ iff,  
$\forall P_{\omega} \in$ Proj(\pc ($Q^{\gamma}$)),
$T_{\delta,\omega} \in Sol(P_{\omega})$ and
$pref(T_{\delta,\omega})\geq \gamma$.
\end{enumerate}
\end{theorem2}

\proof{We will prove each item of the theorem.
\begin{enumerate}
\item ($\Rightarrow$): 
  Consider any
  situation $\omega$ such that $P_{\omega} \in
  Proj$(\pca($Q^\gamma$)). Since \pca($Q^{\gamma}$) is path consistent,
  any consistent partial assignment (\eg that defined by $\omega$)
  can be extended to a complete consistent assignment, say
  $T_{\delta,\omega}$ of 
  \pca($Q^{\gamma}$). Moreover, 
  $T_{\delta, \omega} \in Sol(P_{\omega})$,
  and $pref(T_{\delta,\omega}) \geq \gamma$, since 
  the preference functions are semi-convex and every interval of
  \pca($Q^\gamma$) is a subinterval of the corresponding one 
  in $Q^{\gamma}$. Thus, $opt(P_{\omega}) \geq \gamma$ in $P$. 
  ($\Leftarrow$): Consider a situation $\omega$ such that
  $opt(P_{\omega}) \geq \gamma$. This implies that $\exists
  T_{\delta,\omega} \in Sol(P_{\omega})$ such that $pref(T_{\delta,\omega})
  \geq \gamma$. Since we are in the fuzzy semiring, this happens iff 
  $min_{c_{ij} \in L_r \cup L_c} f_{ij}(T_{\delta,\omega})
  \downarrow_{c_{ij}}) \geq \gamma$. Thus it must be that 
  $f_{ij}(T_{\delta,\omega} \downarrow_{c_{ij}}) \geq \gamma$,
  $\forall c_{ij} \in L_r \cup L_c$ and thus 
  $(T_{\delta,\omega}) \downarrow_{c_{ij}} \in c'_{ij}$, where $c'_{ij} 
  \in L'_r \cup L'_c$. This implies that $P_{\omega} \in Proj(Q^{\gamma})$. 
  Moreover, since $T_{\delta,\omega}$ is a consistent solution of
  $P_{\omega}$ in $Q^{\gamma}$, $P_{\omega} \in
  Proj$(\pca($Q^{\gamma}$)).
 \item By construction of $T^{\gamma}$, 
  $\delta \in Sol(T^{\gamma})$ iff, 
  $\forall P_{\omega} \in Proj$(\pca($Q^{\gamma}$)),
  $\mbox{ } T_{\delta,\omega} \in
  Sol(P_{\omega}) \cap Sol$(\pca($Q^{\gamma}$)).
  Notice that  the fact that $T_{\delta,\omega} \in
  Sol$(\pca($Q^{\gamma}$)) implies that 
  $pref(T_{\delta,\omega}) \geq \gamma$. $\Box$
\end{enumerate}  
}


\begin{corollary2}
\label{corsc2}
Consider an STPPU $P$ and a preference level $\gamma$ and assume that 
$\exists$ $\omega$, situation of $P$, such that 
$opt(P_{\omega}) \geq \gamma$, where $P_{\omega}$ is the 
corresponding projection. Then, if 
STPU \pca($Q^{\gamma}$), 
obtained by cutting $P$ at $\gamma$, and then 
applying path consistency, is not SC the $P$ is not $\gamma$-SC.
\end{corollary2}
  \proof{From item 1 of Theorem \ref{lem12} we get that 
$P_{\omega}$ is a projection of $P$ such that $opt(P_{\omega}) \geq \gamma$ 
iff $P_{\omega} \in Proj$(\pca($Q^{\gamma}$)). Thus, there are complete
assignments to controllable and contingent variables of $P$ 
with global preference 
$\geq \gamma$ iff \pca($Q^{\gamma}$) is consistent, \ie, 
iff $Q^{\gamma}$ is path consistent.
Let us now assume that \pca($Q^{\gamma}$) is not SC. Then by item 2 
of Theorem \ref{lem12}, there is no fixed assignment to controllable variables 
such that it is a solution of every projection in  
$Proj$(\pca($Q^{\gamma}$)) and, for every such projection,  
it gives a global preference $\geq \gamma$.

This means that either such set of projections has no common solution in $P$ 
or every common solution gives a preference strictly lower that $\gamma$.
Thus, $P$ is not $\gamma$-SC since this requires the 
existence of a fixed assignment to controllable variables which 
must be an optimal solution for projections with preference at most 
$\gamma$ (Definition \ref{asc}, Item 1 and 2) and give a preference 
$\geq \gamma$ in all other projections (Definition \ref{asc}, Item 3).}

\begin{theorem2}
\label{lem22}
Consider an 
STPPU $P$, and all preference levels from $\alpha_{min}$ to $\gamma$, and 
assume that the corresponding STPs, $T^{\alpha_{min}}, \dots,
T^{\gamma}$ obtained by cutting $P$ at preference levels 
$\alpha_{min}, \dots, \gamma$, and enforcing strong controllability 
are consistent. 
Then, $\delta \in Sol
(P^\gamma)$, where $P^{\gamma}
=\bigotimes_{i=\alpha_{min}, \dots, \gamma} T^{i}$, 
iff $\mbox{ } \forall P_{\omega} \in Proj(P)$: $T_{\delta,\omega} \in
Sol(P_{\omega})$, if $opt(P_{\omega}) \leq \gamma$, then 
$pref(T_{\delta,\omega})=opt(P_{\omega})$, otherwise 
$pref(T_{\delta,\omega})\geq \gamma$.
\end{theorem2}
\proof{($\Rightarrow$):  
Let us first recall that given two STPs, $P_1$ and $P_2$, defined on
the same set of variables, the STP 
$P_3=P_1 \otimes P_2$ has the same variables as $P_1$ and $P_2$ and 
each temporal constraint $c_{ij}^3=c_{ij}^1 \otimes c_{ij}^2$, that
is, the intervals of $P_3$ are the intersection of the corresponding
intervals of $P_1$ and $P_2$. 
Given this, 
and the fact that the set of projections of P is the same as the set
of projections of the STPU obtained cutting P at $\alpha_{min}$,
we can immediately derive from Theorem~\ref{lem12} 
that any solution of $P^\gamma$ satisfies the condition. 
($\Leftarrow$): Let us now consider 
a control sequence $\delta$ of $P$ such that 
$\delta \not \in Sol(P^{\gamma})$. Then, $\exists j 
\in \{\alpha_{min} \dots \gamma \}$ such that $\delta \not \in
Sol(T^j)$. From Theorem~\ref{lem12} we can conclude that $\exists
P_{\omega}$ such that $opt(P_{\omega})=j \leq \gamma$ such that 
$T_{\delta,\omega}$
is not an optimal solution of $P_{\omega}$. $\Box$}

\begin{theorem2}
If the execution of algorithm 
\bestsc on STPPU $P$ stops due to the occurrence of Event 1 (line 4),
then $P$ is not $\alpha$-SC $\forall \alpha \geq 0$.
\end{theorem2}
\proof{For every preference level $\gamma \leq \alpha_{min}$,
$Q^{\gamma}$=$\gamma$-\solverns{Cut}($P$),
=$\alpha_{min}$-\solverns{Cut}($P$)=$Q^{\alpha_{min}}$. 
The occurrence of Event 1 
implies that $Q^{\alpha_{min}}$ is not strongly controllable.
So it must be the same for all $Q^{\gamma}$, $\gamma \leq
\alpha_{min}$. And thus $P$ is not $\alpha$-SC 
$\forall \alpha \leq \alpha_{min}$. Theorem~\ref{lowerp2} allows us to
conclude the same $\forall \gamma > \alpha_{min}$. $\Box$}

\begin{theorem2}
If the execution of algorithm 
\bestsc on STPPU $P$ stops due to the occurrence of Event 2 (line 11)
at preference level $\gamma$, then 
\begin{itemize}
\item $\gamma-1=opt=max_{T\in Sol(P)} pref(T)$;
\item $P$ is $OSC$ and a control sequence
$\delta$ is a solution of
STP $P^{opt}$ (returned by the algorithm) iff it 
is optimal in any scenario of $P$.
\end{itemize} 
\end{theorem2}
\proof{If the condition of line 11 is satisfied by STPU $Q^{\gamma}$, 
it means that 
there are no schedules of $P$ that have preference $\gamma$. However, 
the same condition was 
not satisfied at the previous preference level, $\gamma-1$, which 
means that there are schedules with preference $\gamma-1$. 
This allows us to conclude that 
$\gamma-1$ is the optimal preference for STPPU $P$ seen as an STPP, 
that is, $\gamma-1=opt=max_{T\in Sol(P)} pref(T)$.
Since we are assuming that line 11 is executed by \bestsc at level 
$opt+1$, the conditions in lines 13 and 16 
must have not been satisfied at preference 
$opt$. This means that at 
level $opt$ the STP $P^{opt}$ (line 15) is consistent.
By looking at line 15, we can see that STP $P^{opt}$ satisfies the hypothesis 
of Theorem~\ref{lem22} from preference  $\alpha_{min}$ 
to preference $opt$. 
This allows us to
conclude that any solution of 
$P^{opt}$ is optimal in any scenario of $P$ and vice versa. 
Thus, P is $opt$-SC and, by Theorem~\ref{oopt2}, it is OSC.
$\Box$}

\begin{theorem2}
If the execution of algorithm 
\bestsc on STPPU $P$ stops due to the occurrence of Event 3 (line 13) 
or Event 4 
(line 16)
at preference level $\gamma$ then $P$ is not OSC but it is  
$(\gamma-1)$-SC and any solution $\delta$ of 
STP $P^{\gamma-1}$ (returned by the algorithm) is 
such that, $\mbox{ } \forall P_{\omega} \in Proj(P)$: 
$T_{\delta,\omega} \in Sol(P_{\omega})$, 
if $opt(P_{\omega}) \leq \gamma-1$, then 
$pref(T_{\delta,\omega})=opt(P_{\omega})$, otherwise 
$pref(T_{\delta,\omega})\geq \gamma-1$.
\end{theorem2}
\proof{If Event 3 or 
Event 4 occurs the condition in line 11 must have not been 
satisfied at preference level $\gamma$. 
This means that STPU \pca($Q^{\gamma}$) is consistent and thus 
there are schedules of $P$ with preference $\gamma$. 
If Event 3 occurs, then the condition in line 13 must be satisfied.
The STPU obtained by cutting $P$ at preference level $\gamma$ and applying 
path consistency is not strongly controllable.  
We can thus conclude, using Corollary \ref{corsc2},  
that $P$ is not OSC.
However since the algorithm had executed line 11 at 
preference level $\gamma$, at $\gamma-1$ it must have reached line 18.
By looking at line 15 we can see that STP $P^{\gamma-1}$ satisfies the hypothesis 
of Theorem~\ref{lem22} 
from preference $\alpha_{min}$ to
preference level $\gamma-1$. 
This allows us to conclude that P is $\gamma-1$-SC. 

If instead Event 4 occurs then it is $P^{\gamma}$ to be inconsistent 
which (by Theorem~\ref{lem22}) means that there 
is no common assignment to executables that is 
optimal for all scenarios with preference $< \gamma$ and at the same time 
 for those with preference equal to $\gamma$. 
However since the execution has reached line 16 
at preference level $\gamma$, 
again we can assume it had successfully completed 
the loop at preference $\gamma-1$ and conclude as above that P is 
$\gamma-1$-SC.$\Box$}

\begin{theorem2}
Determining the optimal strong controllability or 
the highest preference level of $\alpha$-SC 
of an STPPU with \defn{n}
variables and $\ell$ preference levels 
can be achieved in $O(n^3 \ell)$.
\end{theorem2}
\proof{ 
Notice first that the complexity of procedure $\alpha$-\solverns{Cut} (lines 3
and 10) and of 
intersecting two STPs (line 15) is linear in the number of constraints
and thus $O(n^2)$. 
Assuming we have at most $\ell$ different 
preference levels, 
we can conclude 
that the complexity of \bestsc is bounded by that 
of applying $\ell$ times \scstpu, that is 
$\bigoh{n^3 \ell}$ (see Section~\ref{stppu-back}).$\Box$} 

\begin{theorem2}
STPPU $P$ is OWC iff the STPU $Q$, obtained by simply ignoring the
preference functions on all the constraints WC.
\end{theorem2}
\proof{If $P$ is OWC, then for every situation
 $\omega$ of $P$
 there exists a control sequence $\delta$ such that schedule 
 $T_{\delta, \omega}$ is consistent and optimal for projection
 $P_{\omega}$. For every projection $P_{\omega}$ of $P$ there is a
 corresponding projection of $Q$, say $Q_{\omega}$, 
 which is the STP obtained from the
 $P_{\omega}$ by ignoring the preference functions.
 It is easy to see that Definition~\ref{stc} in Section~\ref{TCSPPs}
 implies that any assignment which is an optimal solution of
 $P_{\omega}$ is a solution of $Q_{\omega}$.
 If STPU $Q$ is WC then for every projection $Q_{\omega}$ 
 there exists a control sequence $\delta$ such that schedule
 $T_{\delta,\omega}$ is a solution of $Q_{\omega}$. Again by
 Definition~\ref{stc} in Section~\ref{TCSPPs} we can conclude that the
 corresponding STPP $P_{\omega}$ at least a solution and thus it must
 have at least an optimal solution, that is a solution such that no other
 solution has a higher preference. $\Box$}

\begin{theorem2}
Given an STPPU $P$, consider any preference level $\alpha$ such that
STPU $Q^{\alpha}$, obtained cutting $P$ at $\alpha$, is
consistent. 
If STPU \pca($Q^{\alpha}$) is not DC 
then $P$ is not ODC and it is not $\beta$-DC, $\forall \beta \geq \alpha$.
\label{thnecc12}
\end{theorem2}
\proof{Assume that there is a preference level $\alpha$ such that 
\pca($Q^{\alpha}$) is not DC. 
This means that there is no viable execution strategy 
$S^{\alpha}: Proj$(\pca($Q^{\alpha}$))$ \longrightarrow 
Sol$(\pca($Q^{\alpha}$)) such that 
$\forall P_1, P_2$ in $Proj(Q^{\alpha})$ and for any executable $x$,
if  $[S(P_1)]_{<x} = [S(P_2)]_{<x}$ then $[S(P_1)]_{x} = [S(P_2)]_{x}$.

Let us recall that, due to the semi-convexity of the preference
functions, cutting the STPPU at any given preference level can only
return smaller intervals on the constraints.  
Thus, every projection in $Proj(Q^{\alpha})$ (which is an STP)
corresponds to a projection in $Proj(P)$ which is the STPP obtained 
from the STP by restoring the preference functions as in P.  

Let us now assume, on the contrary, that $P$ is ODC and, thus, that
there exists a viable strategy $S': Proj(P) \longrightarrow Sol(P)$
such that $\forall P_1,P_2 \in Proj(P)$, if  $[S'(P_1)]_{<x} =
[S'(P_2)]_{<x}$ then $[S'(P_1)]_{x} = [S'(P_2)]_{x}$, and 
$pref(S'(P_i))=opt(P_i)$, $i=1,2$. 
Consider, now the restriction of $S'$ to the projections 
in $Proj$(\pca($Q^{\alpha}$)). Since
$pref(S'(P_{\omega})=opt(P_{\omega})$ for every $P_{\omega}$, 
it must be that $\forall P_{\omega} \in Proj$((\pca($Q^{\alpha}$)),
$S'(P_{\omega}) \in Sol$((\pca($Q^{\alpha}$)). Thus 
the restriction of $S'$ satisfies the requirements of the strategy in
the definition of DC. This is in contradiction with the fact that  
\pca($Q^{\alpha}$) is not DC. Thus $P$ cannot be ODC.

By Theorem~\ref{lem12}, 
$\forall P_{\omega} \in Proj(P)$, $P_{\omega} \in
Proj$(\pca($Q^{\alpha}$)) iff $opt(P_{\omega}) \geq \alpha$.
This allows us to conclude that $P$ is not 
$\alpha$-DC. Finally, Theorem~\ref{lowerp2} allows to conclude that $P$
is not $\beta$-DC, $\forall \beta \geq \alpha$. $\Box$}



\begin{lemma2}[useful for the proof of Theorem~\ref{themerge2}]
\label{theostpu2}
Consider an STPU $Q$ on which \solver{DynamicallyCo} \solver{ntrollable} 
has reported success on $Q$.
Consider any constraint AB, where A and B are executables and 
the execution of A always precedes that of B, defined by interval
$[p,q]$ and wait $t_{max}$ \footnote{
Notice that $t_{max}$ is
the longest wait B must satisfy imposed by 
any contingent time-point C on constraint AB.}. 
Then, there exists a viable dynamic strategy $S$
such that $\forall Q_i \in Proj(Q)$, 
$[S(Q_i)]_{B}-[S(Q_i)]_{A} \leq t_{max}$. 
\end{lemma2}
\proof{Such a dynamic strategy is produced 
by algorithm \solver{DC-Execute} shown in  
Figure~\ref{fig:dc:execution}, Section~\ref{stppu-back}.
In fact, in line 5 it is stated that an executable B can be 
executed as soon as, at the current time, the three following
conditions are all satisfied: 
(1) B is $live$, \ie the current time must lie between its lower 
and upper bounds, (2) B is $enabled$, \ie all the variables which must precede
B have been executed, and (3) all waits on B have been satisfied.
Let us denote the current time as $T$, and assume B is $live$ and 
$enabled$ at $T$. 
Thus, $T-([S(Q_i)]_{A}) \in [p,q]$. 
The third requirement
is  satisfied at $T$ only in one of the two following scenarios:  
either the last contingent
time-point for which B had to wait has just occurred 
and thus B can be executed immediately, or the waits for the contingent
time-points, among those for which B had to wait, which have not yet
occurred have expired at $T$. 
In both cases it must be that $T \leq t_{max}+[S(Q_i)_{A}]$.
Thus, $([S(Q_i)]_{B}=T)-[S(Q_i)]_{A} \leq t_{max}$. $\Box$}

\begin{theorem2}
Consider STPPU $P$ and STPUs, $T^{\beta}$ and $T^{\beta+1}$, 
obtained by cutting $P$ respectively at level 
$\beta$ and $\beta +1$ and applying \pc , without finding
inconsistencies, and \DC with success. 
Consider STPU $P^{\beta+1}=$ \solverns{Merge}($T^{\beta}$ , $T^{\beta+1}$).

Then, 
\solverns{Merge}($T^{\beta}$ , $T^{\beta+1}$) does not fail
if and only if 
\begin{itemize}
\item $P^{\beta+1}$ is dynamically controllable and 
\item there is a viable dynamic strategy $S$ 
such that for every projection 
$P_i \in Proj(P^{\beta+1})$, 
\begin{itemize} 
\item if $opt(P_i)=\beta$ or $opt(P_i)=\beta+1$ in
$P$, $pref (S(P_i))=opt(P_i)$;
\item otherwise $pref(S(P_i)) \geq \beta+1$.
\end{itemize}
\end{itemize}
\label{themerge2}
\end{theorem2}

\proof{$\Rightarrow$ The following is a constructive proof in which,
assuming \solver{Merge} has not failed,  
a strategy $S$, satisfying the requirements of the theorem, is defined.

First notice that  $Proj(P^{\beta+1})=Proj(T^{\beta})$. 
In fact, in line 2 of \solver{Merge}, $P^{\beta+1}$ is initialized to 
$T^{\beta}$. 
and  \solver{Merge} changes only requirement intervals leaving
all contingent intervals unaltered. 

Furthermore, $Proj(T^{\beta+1}) \subseteq Proj(T^{\beta})$. This can be
seen using the first claim of Theorem~\ref{lem12} in Section~\ref{osc}.

Let $S'$ and $S''$ be the viable dynamic execution strategies obtained running 
\solver{DC-Execute} respectively on $T^{\beta}$ and $T^{\beta+1}$.
Now, since $Proj(T^{\beta+1}) \subseteq Proj(T^{\beta})$, 
the projections of $T^{\beta}$ will be mapped into two, possibly
different, schedules: one 
by $S'$ and one by $S''$. 
For every projection $P_i \in
Proj(P^{\beta+1})$ and for every executable B, 
notice that if $S''[P_i]_{<B}$ exists then it is equal to 
$S'[P_i]_{<B}$.
We can thus define 
the history of B (which we recall is the set of durations of all contingent
events which have finished prior to B) in the new strategy $S$ as   
$S[P_i]_{<B}=S'[P_i]_{<B}$ for every projection $P_i \in Proj(P^{\beta+1})$ . 
Notice that $S''[P_i]_{<B}$ is not defined if the history 
of B in $P_i$ contains a duration which is mapped into a preference
exactly equal to $\beta$ and thus $P_i$ cannot be a projection of
$T^{\beta+1}$.

We will now consider how to define $S$ depending on which case 
the AB constraint is in $T^{\beta}$ and in $T^{\beta+1}$.

\begin{itemize}

\item Constraint AB is a {\bf Follow or Unordered in $T^{\beta}$ and 
Follow in $T^{\beta+1}$}.
In both cases, \solver{Merge} does not change interval AB, leaving it as
it is in $T^{\beta}$.

Let us first analyze the scenario in which AB is in the {\em Follow} case
in both STPUs. In such a case,
the execution of B will always follow that of any contingent time
point C in both problems. Thus, for every projection 
$P_{\omega} \in Proj(P^{\beta+1})$, 
we have $S[P_{\omega}]_{<B}=\omega$.
Since both problems are dynamically controllable 
$[p^{\beta},q^{\beta}] \neq \emptyset$ and 
$[p^{\beta+1},q^{\beta+1}] \neq \emptyset$.
Furthermore, since path consistency 
has been enforced in both problems, the constraints are
in minimal form (see Section~\ref{stppu-back}),
that is, for every value $\delta_{AB}$ in $[p^{\beta},q^{\beta}]$ (resp. 
$[p^{\beta+1},q^{\beta+1}]$) there is a situation $\omega$ of $T^{\beta}$
(resp. $T^{\beta+1}$) such that $T_{\delta,\omega} \in Sol(P_{\omega})$
and $\delta _{\downarrow AB}=\delta_{AB}$. 
Finally, since $Proj(T^{\beta+1}) \subseteq Proj(T^{\beta})$, it must be that 
$[p^{\beta+1},q^{\beta+1}] \subseteq [p^{\beta},q^{\beta}]$.


Next we consider the scenario in which AB is in the {\em Unordered}
case in $T^{\beta+1}$.
Let us start by proving that, in such a case, it must be that 
$[p^{\beta+1}, q^{\beta+1}] \subseteq [p^{\beta}, t^{\beta}]$. 
First, we show that $p^{\beta+1} \geq p^{\beta}$. By definition, 
$p^{\beta+1}$ is such that there is a situation $\omega$ such that 
$P_{\omega} \in Proj(T^{\beta+1})$ and there is a schedule 
$T_{\delta,\omega} \in Sol(P_{\omega})$ such that 
$\delta_ {\downarrow AB}=p^{\beta+1}$. Since 
$Proj(T^{\beta+1}) \subseteq Proj(T^{\beta})$, then 
$p^{\beta+1} \in [p^{\beta},q^{\beta}]$.
Next let us prove that it must be $t^{\beta} > q^{\beta+1}$.
Notice that the wait $t^{\beta}$ induces a partition of the 
situations of $T^{\beta}$ into two sets: those such that, 
for every contingent point C, $\omega_{\downarrow AC}< t^{\beta}$, and 
those which for some contingent point C$'$,  
$\omega_{\downarrow AC'}\geq t^{\beta}$. In the first case, all the
contingent events will have occurred before the expiration of the 
wait and B will be executed before $t_A+t^{\beta}$ (where $t_A$ 
is the execution time of A).
In the second case it will be safe to execute B at $t_A+t^{\beta}$.
Given that $Proj(T^{\beta+1}) \subseteq Proj(T^{\beta})$, 
and that B is constrained to follow the execution of every 
contingent time-point in $T^{\beta+1}$, it must be that 
all the projections of $T^{\beta+1}$ 
belong to the first set of the partition
and thus $q^{\beta+1} < t^{\beta}$.

In both cases it is, hence, 
sufficient to define the new strategy $S$ as follows:
on all projections, $P_i, P_j \in Proj(P^{\beta+1})$ such that
$[S(P_i)]_{<B} =[S(P_j)]_{<B}$ then 
$[S(P_i)]_{B} =[S(P_j)]_{B}=[S''(P_i)]_{B}$ 
if 
$[S''(P_i)]_{B}$ exists, otherwise 
$[S(P_i)]_{B} =[S(P_j)]_{B}=[S'(P_i)]_{B}$.
This assignment guarantees to identify projections on constraints
mapped into preferences $\geq \beta+1$ if $[S''(P_i)]_{B}$ exists and
thus $P_i \in Proj(T^{\beta+1})$, otherwise $\geq \beta$ for those 
projections in $Proj(T^{\beta})$ but not in $Proj(T^{\beta+1})$. 

\item  Constraint AB is a {\bf Precede case in $T^{\beta}$ 
and  in $T^{\beta+1}$}.
B must precede any contingent time-point C. 
This means that 
any assignment to A and B corresponding to a value 
in $[p^{\beta},q^{\beta}]$ (resp. $[p^{\beta+1},q^{\beta+1}]$)
can be extended to a complete solution of any projection in 
$Proj(T^{\beta})$ (resp. $Proj(T^{\beta+1})$).
Interval $[p',q']$ is, in fact, obtained by \solver{Merge}, by 
intersecting the two intervals. 
Since
we are assuming that \solver{Merge} has not failed, 
such intersection cannot be empty (line 6 of Figure~\ref{merge}). 
We can, thus, for example, define S as follows: on any pair of projections 
$P_i, P_j \in Proj(P^{\beta+1})$
if $[S(P_i)]_{<B} =[S(P_j)]_{<B}$
then 
$[S(P_i)]_{B} (=[S(P_j)]_{B})=p'$.

\item Constraint AB is 
{\bf Unordered in $T^{\beta}$ and Unordered or Precede in $T^{\beta+1}$}.
First let us recall that
the result of applying \solver{Merge} is interval $[p',q']$, where 
$p'=p^{\beta}$, $q'=min(q^{\beta},q^{\beta+1})$ 
and wait $t'=max(t^{\beta},t^{\beta+1})$. Since, by hypothesis,
\solver{Merge} has not failed, it must be that $t' \leq q'$ 
(line 9, Figure~\ref{merge}.

Notice that, due to the semi-convexity of the preference 
functions, $p^{\beta} \leq p^{\beta+1}$. In fact, B will be executed 
at $t_{A} +p^{\beta}$ (where $t_{A}$ is the time at which A has been
executed) only if all the contingent 
time-points for which B has too wait for have occurred. 
Let us indicate with  
$x_{mlb}^{\beta}$ (resp. $x_{mlb}^{\beta+1}$) the maximum lower bound 
on any AC constraint in 
$T^{\beta}$ (resp. in $T^{\beta+1}$),
where B has to wait for C. 
Then it must be that 
$p^{\beta}\geq x_{mlb}^{\beta}$ (resp. 
$p^{\beta+1}\geq x_{mlb}^{\beta+1}$).
However due to the semi-convexity of the preference functions
$x_{mlb}^{\beta} \leq x_{mlb}^{\beta+1}$.

In this case we will define strategy $S$ as follows.
For any pair of projections $P_i, P_j \in Proj(P^{\beta+1})$,
if $[S(P_i)]_{<B}$ =$ [S(P_j)]_{<B}$ then 
$[S(P_i)]_{B} = [S(P_j)]_{B}=$ $max ([S''(P_i)]_{B},$ $[S'(P_i)]_{B})$
whenever $[S''(P_i)]_{B}$ is defined. 
Otherwise
$[S(P_i)]_{B}$ =$[S(P_j)]_{B}=[S'(P_i)]_{B}$.
From Lemma~\ref{theostpu2} we have that 
$max ([S''(P_i)]_{B}, [S'(P_i)]_{B}) \leq t'$, hence 
$[S(P_i)]_{B}=( [S(P_j)]_{B}) \in [p',q']$.

Let us now consider the preferences induced on the constraints by this 
assignment. First let us consider the case when 
$max ([S''(P_i)]_{B}, [S'(P_i)]_{B})=[S''(P_i)]_{B}$. Since 
$S''$ is the dynamic strategy in $T^{\beta+1}$ all its assignment
identify projections with preference $\geq \beta+1$. 
If instead $max ([S''(P_i)]_{B},$ $ [S'(P_i)]_{B})$ $=[S'(P_i)]_{B}$, then 
it must be that $[S'(P_i)]_{B}>[S''(P_i)]_{B}$. 
However we know, from 
Lemma~\ref{theostpu2} that $[S''(P_i)]_{B} \leq t^{\beta+1} \leq t'$
and that $[S'(P_i)]_{B} \leq t'$.
This implies that $[S'(P_i)]_{B} \in [p^{\beta+1}, t']$ and thus 
it is an assignment with preference $\geq \beta+1$.  
Finally, if $[S''(P_i)]_{B}$ is not defined, as noted above, 
then $P_i \not \in Proj(T^{\beta+1})$ 
and thus $opt(P_i) = \beta$ (since by Theorem~\ref{lem12} in Section 
\ref{osc} we have that $P_i \in Proj(T^{\beta}) \Leftrightarrow
opt(P_i) \geq \beta$). Thus, 
$[S(P_i)]_{B}$ =$ [S(P_j)]_{B}=[S'(P_i)]_{B}$, which, being an
assignment in $T^{\beta}$, identifies preferences $\geq
\beta=opt(P_i)$.
\end{itemize} 

$\Leftarrow$
We have just shown that, if \solverns{Merge} does not fail, then there
is a dynamic strategy (with the required additional properties) which 
certifies that $P^{\beta+1}$ is dynamically controllable.

Assume, instead, that 
\solver{Merge} fails on some constraint.
There are two cases in which this can happen.
The first one is when AB is a {\em Precede} case in both 
$T^{\beta}$ and $T^{\beta+1}$ and     
$[p^{\beta},$ $ q^{\beta}]$ $ \cap [p^{\beta+1},$ 
$ q^{\beta+1}]$ $ = \emptyset$.
As proven in \cite{mor:stnu}, the projection on AB of 
any viable dynamic strategy for $T^{\beta}$ is in $[p^{\beta},
  q^{\beta}]$ and the projection on AB of 
any viable dynamic strategy for $T^{\beta+1}$ is in 
$[p^{\beta+1},q^{\beta+1}]$.
The dynamic viable strategies of $T^{\beta}$ give optimal 
solutions for projections with optimal preference equal to $\beta$.
The dynamic viable strategies of the $T^{\beta+1}$ give optimal 
solutions for projections with optimal preference equal to $\beta+1$.
Since the projections of $T^{\beta+1}$ are a subset of those in 
$T^{\beta}$, if $[p^{\beta}, q^{\beta}] \cap 
[p^{\beta+1},q^{\beta+1}] = \emptyset$ then a strategy either is
optimal for projection in $T^{\beta}$ but not for those in
$T^{\beta+1}$ or vice-versa.

The second case occurs when 
\solver{Merge} fails on some constraint AB which is either 
an {\em Unordered} case in both $T^{\beta}$ and $T^{\beta+1}$ or 
is an  {\em Unordered} case in $T^{\beta}$ and a $precede$ case in 
$T^{\beta+1}$.
In such cases the failure is due to the fact that
$[t^{\beta}, q^{\beta}]$ $\cap$ $ 
[t^{\beta+1},$ $q^{\beta+1}] = \emptyset$.
It must be that either $q^{\beta+1}< t^{\beta}$ or  
$q^{\beta}< t^{\beta+1}$.
If the upper bound of the interval on AB is $q^{\beta+1}$ 
the  there must be at at least a contingent time-point C such that 
executing B more than $q^{\beta+1}$ after A is either
inconsistent with some assignment of C or 
it gives a preference lower than $\beta+1$.
On the other side, if the wait on constraint AB in $T^{\beta}$ is 
$t^{\beta}$  there must be at least a contingent time-point C$'$ such
that executing B before $t^{\beta}$ is either inconsistent or not
optimal with some future occurrences of C'.
Again there is no way to define a viable dynamic strategy that is
simultaneously optimal for projections with optimal value equal to
$\beta$ and for those with optimal value $\beta+1$. $\Box$}  


\begin{lemma2}[Useful for the proof of Theorem~\ref{cor22}]
\label{cor12}
Consider strategies $S'$, $S''$ and $S$ as defined in Theorem~\ref{themerge2}.
Then
\begin{enumerate}
\item for any projection of $P^{\beta+1}$, $P_i$, 
$pref(S(P_i)) \geq pref(S'(P_i))$ and for every projection, $P_z$, 
of $T^{\beta+1}$, $pref(S(P_z)) \geq \beta +1$;
\item for any constraint AB, $[S(P_i)]_{B} \geq t'$. 
\end{enumerate}
\end{lemma2}
\proof{ 
\begin{enumerate}
\item Obvious, since in all cases 
either $[S(P_i)]_{B}=$ $[S'(P_i)]_{B}$
or $[S(P_i)]_{B}=$ $[S''(P_i)$ $]_{B}$ and 
$pref(S''(P_i)) \geq pref(S'(P_i))$ since for every executable B
$[S''(P_i)]_{B} $ $\in T^{\beta+1}$. Moreover, for every projection
$P_z$ of $T^{\beta+1}$, for every executable B, 
$[S(P_z)]_{B}=$ $[S''(P_z)]_{B}$.  
\item Derives directly from the fact that either 
$[S(P_i)]_{B}=[S'(P_i)]_{B}$ or $[S(P_i)]_{B}=$ $[S''(P_i)]_{B}$ 
and Lemma~\ref{theostpu2} $\Box$. 
\end{enumerate}
}

\begin{theorem2}
\label{cor22}

Consider STPPU $P$ and for every preference level, $\alpha$, define
$T^{\alpha}$ as the STPU obtained by cutting $P$ at $\alpha$, then
applying \pc and then \DC. Assume that $\forall \alpha \leq \beta$,
$T^{\alpha}$ is DC.  Consider STPU $P^{\beta}$:
$$
P^{\beta}=\solverns{Merge}( \solverns{Merge}(\dots \solverns{Merge}(
\solverns{Merge}(T^{\alpha_{min}},T^{\alpha_{min}+1}), T^{\alpha_{min}+2}),
\dots), T^{\beta})
$$
with $\alpha_{min}$ the minimum preference on any constraint in P.
Assume that, when applied, \solverns{Merge} always returned a consistent STPU.
Then, there is a viable dynamic strategy $S$, 
such that $\forall P_i \in Proj(P)$, if $opt(P_i) \leq \beta$ 
then $S(P_i)$ is an optimal solution of $P_i$, otherwise 
$pref(S(P_i)) \geq \beta+1$.
\end{theorem2}
\proof{
We will prove the theorem by induction.
First, notice that, by construction 
$Proj$ $(T^{\alpha_{min}})=Proj(P)$. This allows us to conclude
that $Proj(P^{\beta})=Proj(P)$, since, every time 
\solver{Merge} is applied, the new STPU has the same contingent 
constraints as the STPU given as first argument.

Now, since $T^{\alpha_{min}}$ is dynamically controllable 
any of its viable dynamic strategies, say $S^{\alpha_{min}}$ will be
such that $S^{\alpha_{min}}(P_i)$ is optimal if
$opt(P_i)=\alpha_{min}$ and, otherwise, $pref(S(P_i)) \geq \alpha_{min}$.
Consider now  $P^{\alpha_{min}+1}$=\solverns{Merge}
($T^{\alpha_{min}}$,$T^{\alpha_{min}+1}$). Then 
by Theorem~\ref{themerge2}, we know that there is a 
strategy, $S^{\alpha_{min}+1}$, 
such that $S^{\alpha_{min}+1}(P_i)$ is an optimal solution 
of $P_i$ if $opt(P_i) \leq \alpha_{min}+1$ and $pref(S(P_I))\geq
\alpha_{min} +1$ otherwise.

Let us assume that STPU $P^{\alpha_{min}+k}$, as defined in the
hypothesis, satisfies the thesis and
that $P^{\alpha_{min}+k+1}$, as defined in the hypothesis, 
where $\alpha_{min}+k+1 \leq \beta$, does
not.
Notice that this implies that there is a strategy, 
$S^{\alpha_{min}+k}$, 
such that $S^{\alpha_{min}+k}(P_i)$ is an optimal solution 
of $P_i$ if $opt(P_i) \leq \alpha_{min}+k$ and 
$pref(S(P_i)) \geq \alpha_{min} +k$ for all other projections.
Since $\alpha_{min}+k+1 \leq \beta$, then, by hypothesis we also have that 
$T^{\alpha_{min}+k+1}$ is DC.
Moreover, by construction, $P^{\alpha_{min}+k+1}$=\solverns{Merge}
($P^{\alpha_{min}+k}$,$T^{\alpha_{min}+k+1}$), since 
\solver{Merge} doesn't fail.
Thus,
using Theorem~\ref{themerge2} and 
using strategy $S^{\alpha_{min}+k}$ for $P^{\alpha_{min}+k}$ 
in the construction of 
Theorem~\ref{themerge2}, by Lemma~\ref{cor12}, 
we will obtain a dynamic strategy, 
$S^{\alpha_{min}+k+1}$, such
that for every projection $P_i$, 
$pref(S^{\alpha_{min}+k+1}(P_i)) \geq pref(S^{\alpha_{min}+k}(P_i))$
and such that $S^{\alpha_{min}+k+1}$ $(P_j)$ is an optimal solution 
for all projections $P_j$ such that $opt(P_j)= \alpha_{min}+k+1$
and $pref(S(P_j)) \geq \alpha_{min}+k+1$ on all other projections.
This allows us to conclude that $S^{\alpha_{min}+k+1}(P_h)$ is an
optimal solution for all projections $P_h$ such that $opt(P_h) 
\leq \alpha_{min}+k+1$. This is contradiction with the assumption that
$P^{\alpha_{min}+k+1}$ doesn't satisfy the thesis of the theorem. $\Box$}

\begin{theorem2}
\label{stop2}
Given an STPPU P, the execution of algorithm \bestdc on $P$ 
terminates.
\end{theorem2} 
\proof{We assume that the preference set is 
discretized and that there are a finite number of different
preferences.
\bestdc starts from the lowest preference and cuts at each level P.
If, at a given level, the STPU obtained is not consistent or
not dynamically controllable or the merging procedure fails, then \bestdc
stops at that level.
Assume, instead, that, 
as it moves up 
in the preference ordering, 
none of the events above occur.
However at a certain point the cutting level will be higher than 
the maximum on some preference function (or it will be outside of the
preference set) in which case cutting the problem will give an
inconsistent STP.$\Box$ }

\begin{theorem2}
\label{stop42}
Given an STPPU $P$ as input, \bestdc terminates in line 4 iff 
$\not \exists \alpha \geq 0$  such that 
P is $\alpha$-DC.
\end{theorem2}

\proof{ $\Rightarrow$. Assume \bestdc terminates in line 4. Then, 
the STPU obtained by cutting $P$ at the minimum preference,
$\alpha_{min}$, on any
  constraint is not DC. However cutting at the minimum preference on
  any constraint or at preference level 0 gives the same STPU. 
By Theorem~\ref{thnecc12} we can conclude that $P$ is not 
$\alpha$-DC $\forall \alpha \geq 0$ and, thus, not ODC.

$\Leftarrow$. Assume $P$ is not $\alpha$-DC for all preferences 
$\alpha \geq 0$. Then cutting $P$ at the minimum preference $\alpha_{min}$ 
cannot give a dynamically controllable problem, otherwise, $P$ would be 
$\alpha_{min}$-DC. Hence, \bestdc will exit in line 4. $\Box$}

\begin{theorem2}
\label{stop112}
Given an STPPU $P$ as input, \bestdc terminates in line 11 iff 
P is ODC.
\end{theorem2}
\proof{$\Rightarrow$.
Assume \bestdc terminates in line 11 when considering 
preference level $\beta$.
Then, STPU $Q^{\beta}$ obtained by cutting STPPU $P$ at level $\beta$ is
not path consistent. From this we can immediately conclude that there
is no projection $P_i \in Proj(P_i)$ such that $opt(P_i) \geq \beta$.

Since \bestdc did not terminate before, we must assume that up to
preference $\beta-1$, all the tests (path consistency, dynamic
controllability, and \solver{Merge}) were successful.

Now consider the STPU $P^{\beta-1}$
obtained at the end of the iteration corresponding to preference level
$\beta-1$.
It is easy to see that $P^{\beta-1}$ satisfies the hypothesis of
Theorem~\ref{cor22}. This allows us to conclude that there is 
a viable dynamic strategy $S$ such that for every projection $P_i$, such
that $opt(P_i) \leq \beta-1$, $S(P_i)$ is an optimal solution 
of $P_i$. However since we know that all projections of $P$ are such
that $opt(P_i) < \beta$, this allows us to conclude that $P$ is ODC.

$\Leftarrow$. If $P$ is ODC then 
there is a viable strategy S such that for every pair of projections,
$P_i, P_j \in Proj(P)$, and for very executable B, 
if $[S(P_i)]_{<B} =[S(P_j)]_{<B}$ then $[S(P_i)]_{B} =[S(P_j)]_{B}$
and $S(P_i)$ is an optimal solution of $P_i$ and $S(P_j)$ is an
optimal solution of $P_j$.

By Theorem~\ref{stop42} we know that \bestdc cannot stop in line 4.

Let us now consider line 13 and show that 
if \bestdc sets $\alpha$-DC to $true$ in that line 
then $P$ cannot be ODC.
In fact the condition of setting  $\alpha$-DC to $true$ in line 13 is
that the STPU obtained by cutting $P$ at preference level $\beta$ is path
consistent but not dynamically controllable. This means that there are 
projections, \eg $P_j$, of $P$ such that $opt(P_j)=\beta$. However, there is no
dynamic strategy for the set of those projections. Thus, $P$ cannot be ODC.

Let us now consider line 16, and show that, if $P$ is ODC \bestdc cannot
set $\alpha$-DC to $true$. 
If \bestdc sets $\alpha$-DC to $true$ 
then \solver{Merge} failed.
Using Theorem 
\ref{themerge2}, we can conclude  
that there is no dynamic viable strategy $S$ such that
for every projection of $P$, $P_i$, (remember that
$Proj(P^{\beta-1})=Proj(P)$) $S(P_i)$ is an optimal solution 
if $opt(P_i) \leq \beta$. However, we know there are projections of P
with optimal preference equal to $\beta$ (since we are assuming
\bestdc is stopping at line 16 and not 11). Thus, $P$ cannot be ODC.$\Box$}

\begin{theorem2}
Given STPPU $P$ in input, \bestdc stops at lines 13 or 16 
at preference level $\beta$ iff $P$ is $(\beta-1)$-DC and not ODC.
\end{theorem2} 
\proof{
$\Rightarrow$. Assume that \bestdc sets $\alpha$-DC to $true$ in line 13, when
considering preference level $\beta$. Thus, 
the STPU obtained by cutting $P$ at level $\beta$ is path consistent but not 
DC. However since $\beta$ must be the first preference level at which
this happens, otherwise the \bestdc would have stopped sooner, 
we can conclude that the iteration at preference level $\beta-1$ 
was successful. Considering $P^{\beta-1}$ and using Theorem~\ref{cor22} we
can conclude that there is a viable dynamic strategy $S$ such that,
for every projection of $P$, $P_i$, if $opt(P_i)\leq \beta-1$ then 
$S(P_i)$ is an optimal solution of $P_i$ and $pref(S(P_i)) \geq
\beta-1$ otherwise.
But this is the definition of $\beta-1$-dynamic controllability.

If \bestdc terminates in line 16, by Theorem~\ref{cor22} and
and Theorem~\ref{themerge2} we can conclude that, while there is 
a viable dynamic strategy $S$ such that
for every projection of $P$, $P_i$, if $opt(P_i)\leq \beta-1$ then 
$S(P_i)$ is an optimal solution of $P_i$ and $pref(S(P_i)) \geq
\beta-1$ otherwise, 
there is no such strategy
guaranteeing optimality also for projections with optimal preference 
$\beta$. Again, $P$ is $\beta-1$-DC.

$\Leftarrow$. If $P$ is $\alpha$-DC, for some $\alpha \geq 0$ 
then by Theorem~\ref{stop42}, \bestdc does not stop in line 4.
If $P$ is $\alpha$-DC, but not ODC, 
for some $\alpha \geq 0$  then by Theorem~\ref{stop112}, 
\bestdc does not stop in line 11. By Theorem~\ref{stop2}, \bestdc
always terminates, so it must stop at line 13 or 16.$\Box$}

\begin{theorem2}
The complexity of determining ODC or 
the highest preference level $\alpha$ of $\alpha$-DC of an STPPU 
with \defn{n} variables, 
a bounded number of preference levels \defn{l} 
is $O(n^5 \ell)$.
\end{theorem2}
\proof{
Consider the pseudocode of algorithm 
\bestdc in Figure~\ref{hadc}.

The complexity of 
$\alpha_{min}$-\solverns{Cut}($P$) 
in line 3 is $O(n^2)$, since every
constraint must be considered, an there are up to $O(n^2)$
constraints, and for each constraint the time for finding the interval 
of elements mapped into preference $\geq \alpha_{min}$ is constant.
The complexity of checking if the STPU obtained is DC is
$O(n^5)$.
Thus, lines 3 and 4, which are always performed, have an overall
complexity of $O(n^5)$.
Lines 7 and 8, clearly, take constant time.

Let us now consider a fixed preference level $\beta$ and compute the
cost of a complete $while$ iteration on $\beta$.
\begin{itemize}
\item (line 10) the complexity of $\beta$-\solverns{Cut}($P$) is $O(n^2)$;
\item (line 11) the complexity of applying \pc for testing path
  consistency is $O(n^3)$ (see Section~\ref{tcsp}, 
\cite{meiri});

\item (line 13) the complexity of testing DC using \DC is $O(n^5)$,
  (see Section~\ref{stppu-back} and  \citeauthor{dc-rev05}, 2005);
\item (line 15) constant time;
\item (line 16-18) the complexity of \solver{Merge} is $O(n^2)$, since 
at most $O(n^2)$ constraints must be considered and for each
constraint merging the two intervals has constant cost;
\item (line 19) constant time.
\end{itemize}
We can conclude that the complexity of a complete iteration at any
given preference level is $O(n^5)$.
In the worst case, the $while$ cycle is performed $\ell$ times.
We can, thus, conclude that the total complexity of \bestdc is $O(n^5 \ell)$ 
since
the complexity of the operations performed in lines 24-27
is constant. $\Box$}

\bibliographystyle{theapa}
\bibliography{biblio-stppujour}





\end{document}